\documentclass{article}
\PassOptionsToPackage{numbers,square,sort&compress}{natbib}
\usepackage[main, final]{neurips_2026}

\usepackage[utf8]{inputenc} 
\usepackage[T1]{fontenc}   
\usepackage{hyperref}       
\usepackage{url}            
\usepackage{booktabs}      

\usepackage{amsfonts}       
\usepackage{nicefrac}      
\usepackage{microtype}     
\usepackage{xcolor}        
\usepackage{kotex}
\usepackage{graphicx}
\usepackage{amsmath}
\usepackage{multirow}
\usepackage{float}
\usepackage{wrapfig}
\usepackage[table]{xcolor}
\usepackage{caption}
\usepackage{placeins} 
\usepackage{titletoc}

\title{Disentangled Sparse Representations for Concept-Separated Diffusion Unlearning}

\author{
  Hyeonjin Kim\thanks{Equal Contribution.} \\
  Yonsei University\\
  \texttt{hyeonjin@yonsei.ac.kr} \\
  \And
  Hangyeol Jung\footnotemark[1] \\
  KAIST\\
  \texttt{han.jung@kaist.ac.kr} \\
  \And
  Heechan Yun \\
  Yonsei University\\
  \texttt{jason0295@yonsei.ac.kr} \\
  \And
  Sungjun Yun \\
  Yonsei University\\
  \texttt{sungjun.yun@yonsei.ac.kr} \\
  \And
  Dong-Jun Han \\
  Yonsei University\\
  \texttt{djh@yonsei.ac.kr} \\
}

\begin{document}

\maketitle

\begingroup
\renewcommand{\thefootnote}{}
\footnotetext{Code available at \textcolor{magenta}{https://github.com/khyeonjin/SAEParate}}
\endgroup

\begin{abstract} 
Unlearning specific concepts in text-to-image diffusion models has become increasingly important for preventing undesirable content generation. Among prior approaches, sparse autoencoder (SAE)-based methods have attracted attention due to their ability to suppress target concepts through lightweight manipulation of latent features, without modifying model parameters. However, SAEs trained primarily for sparse reconstruction do not explicitly enforce concept-wise separation, allowing semantically related concepts to share latent features. To address this, we propose \textbf{SAEParate}, which explicitly separates semantically overlapping concepts in the SAE latent space through concept-aware contrastive learning, supported by cross-attention-guided spatial aggregation that emphasizes concept-relevant regions. In addition, we enhance the encoder with a nonlinear transformation to increase its expressive capacity under this separation objective, enabling more concept-discriminative latent representations. Experiments on UnlearnCanvas demonstrate state-of-the-art performance, with particularly strong gains in joint style-object unlearning, a challenging setting where existing methods suffer from severe interference between target and non-target concepts.
\end{abstract}

\section{Introduction} 
Diffusion-based text-to-image (T2I) models have recently achieved remarkable progress in image synthesis, producing highly realistic and diverse visual content \cite{rombach2022high, ramesh2022hierarchical}. However, these models can also generate undesirable content, including violent or sexually explicit imagery \cite{rando2022redteaming}, as well as content that raises copyright or privacy concerns. This has motivated growing interest in machine unlearning \cite{10.1109/SP.2015.35}, which aims to remove the influence of specific data or concepts from trained models.

T2I diffusion unlearning methods can be broadly categorized into two approaches, namely modifying model parameters \cite{fan2024salun, Wu_2025_CVPR, lyu2024one, gandikota2024unified} and suppressing the generation of target concepts without directly editing the model \cite{li2024get, cywinski2025saeuron}. Among the latter, SAeUron~\cite{cywinski2025saeuron} employs sparse autoencoders (SAEs)~\cite{pmlr-v15-coates11a} as auxiliary modules to extract interpretable latent features~\cite{huben2024sparse} and suppresses target concepts by ablating selected features. 
This provides a lightweight and modular alternative to parameter-modifying approaches, enabling concept removal while keeping the base diffusion model unchanged.

\begin{figure}
\vspace{-8mm}
    \centering
    \includegraphics[width=0.93\linewidth]
    {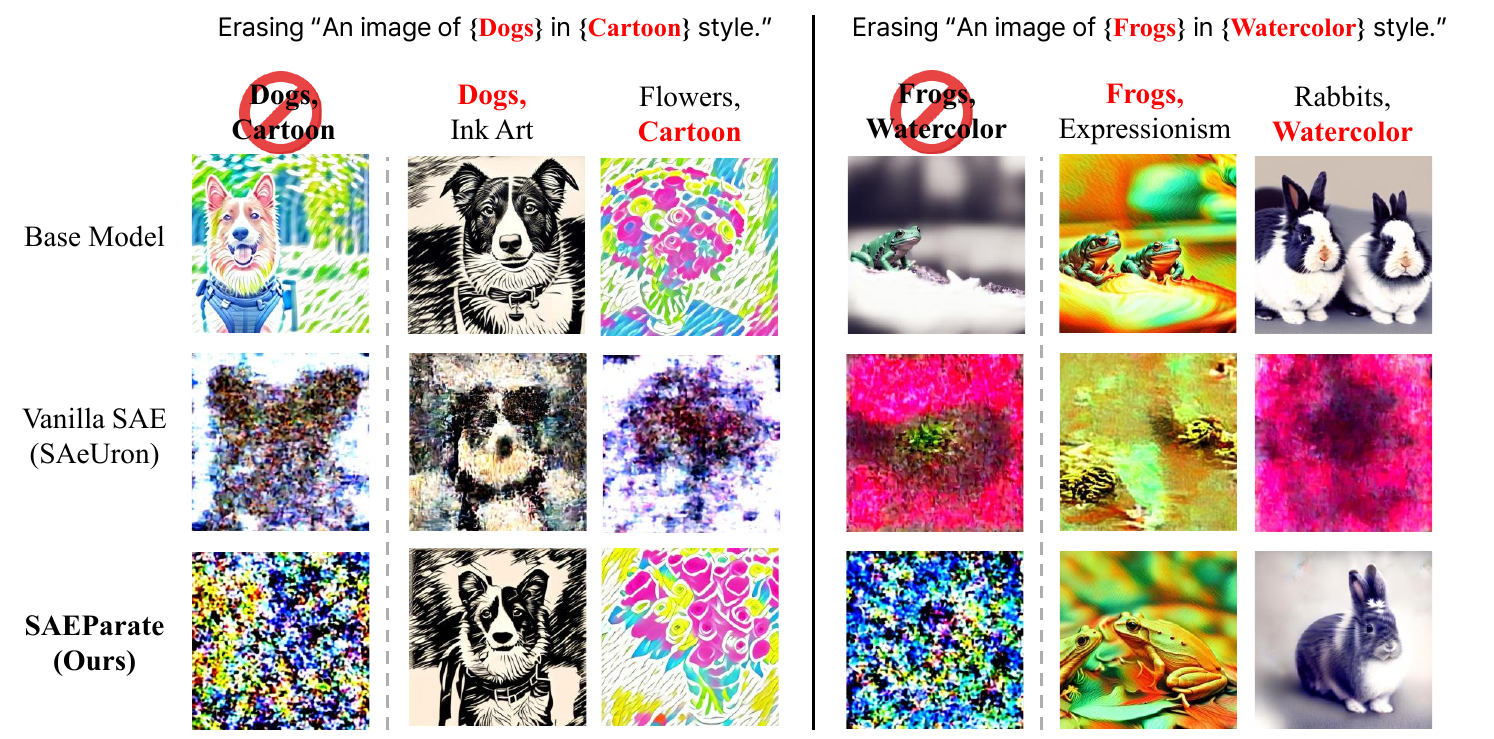}
        \caption{\textbf{Qualitative comparison of joint combination unlearning results.} 
       Unlearning results for the target style--object combinations \textit{Dogs in Cartoon style} (left) and \textit{Frogs in Watercolor style} (right). First column: target combinations to be removed. Other columns: non-target combinations sharing either the same object or style. SAEParate selectively erases the target while preserving non-target concepts, while SAeUron degrades non-target outputs due to latent space interference.}
    \label{fig:feature_overlap_viz}
    \vspace{-3mm}
\end{figure}

However, the sparse reconstruction objective of conventional SAEs is not necessarily aligned with fine-grained diffusion unlearning. Accurate sparse reconstruction does not guarantee concept-level separation, and semantically related concepts may therefore share latent features. Ablating target-associated features can consequently interfere with retained concepts that rely on the same features. This issue is particularly problematic in joint style--object unlearning, where concepts share partial semantics. As shown in Figure~\ref{fig:feature_overlap_viz}, removing \textit{``Dogs in Cartoon style''} with SAeUron also degrades related non-target concepts such as \textit{``Flowers in Cartoon style''}. This raises a key question:

\textit{Can we better separate semantically overlapping concepts in the SAE latent space so that target concepts can be removed with minimal impact on non-target ones?}

In this work, we propose \textbf{SAEParate}, which explicitly learns concept-separated sparse representations for diffusion model unlearning. The key idea is to organize latent activation patterns into concept-specific clusters via a latent-space contrastive objective computed over concept-relevant regions identified by cross-attention, enabling more precise suppression of target concepts with minimal interference to non-target ones. To further improve this separation, we enhance the encoder with a nonlinear transformation, increasing its expressive capacity to learn more discriminative and disentangled features. As shown in Figure~\ref{fig:feature_overlap_viz}, SAEParate effectively erases the target style--object combination \textit{``Dogs in Cartoon style''}, with minimal impact on non-target concepts that share partial attributes (e.g., \textit{``Dogs in Ink Art style''} or \textit{``Flowers in Cartoon style''}). 
This leads to more reliable concept removal in challenging settings where existing approaches suffer from significant interference, highlighting the importance of concept separation.

We evaluate our method on UnlearnCanvas \cite{zhang2024unlearncanvas}, a standard benchmark for diffusion model unlearning. SAEParate outperforms existing baselines on both single-attribute unlearning (style or object) and joint style-object unlearning while preserving the generation quality of non-target concepts. 
Our main contributions are summarized as follows:
\begin{enumerate}
    \item We identify concept-level sharing of SAE latent features as a key cause of unintended interference in SAE-based diffusion unlearning, and characterize this phenomenon through overlapping activation patterns in the latent space.

    \item We propose SAEParate, a unified concept-separation framework that explicitly separates semantically overlapping concepts in the SAE latent space, enabling selective feature intervention with reduced interference to retained concepts.

    \item We demonstrate that SAEParate consistently improves unlearning across benchmark, unseen-concept, and safety-critical settings, with the largest benefits arising when target and retained concepts are semantically similar and highly entangled in the latent space.
\end{enumerate}

\vspace{-2mm}

\section{Related Work}
\vspace{-1.5mm}

\textbf{Concept Unlearning in Diffusion Models.}
Machine unlearning~\cite{10.1109/SP.2015.35} aims to remove specific data or knowledge from a trained model. In text-to-image diffusion models, this is commonly formulated as concept erasure while preserving unrelated generation capabilities~\cite{zhang2024unlearncanvas}. Existing approaches intervene at different levels: ESD~\cite{gandikota2023erasing} and EDiff~\cite{Wu_2025_CVPR} modify model behavior through optimization-based objectives; SalUn~\cite{fan2024salun}, SHS~\cite{wu2024scissorhands}, and FMN~\cite{zhang2024forget} restrict updates to selected parameters or layers; and CA~\cite{kumari2023ablating} and SA~\cite{heng2023selective} reformulate the target distribution for selective forgetting. Parameter-efficient or training-free alternatives include SPM~\cite{lyu2024one}, UCE~\cite{gandikota2024unified}, and SEOT~\cite{li2024get}.

SAE-based approaches take a different route by intervening on intermediate activations through an auxiliary sparse representation rather than encoding the forgetting behavior into the diffusion model parameters. This modular formulation leaves the pretrained generator itself untouched and allows the intervention to be removed when no longer needed. Moreover, sparsity provides a direct view of the latent components being modified during unlearning. SAeUron~\cite{cywinski2025saeuron} suppresses target-associated SAE features. SAEmnesia~\cite{cassano2026saemnesia} instead assigns each labeled concept to a designated latent, enabling explicit concept localization, but this one-to-one mapping can be restrictive when fine-grained or compositional semantics are distributed across multiple features. PER~\cite{cassano2026look} addresses an orthogonal intervention-level issue by replacing target patches to mitigate out-of-distribution activations induced by steering. In contrast, SAEParate retains distributed feature usage and explicitly organizes sparse activation patterns according to concept identity, promoting concept-level separation across the full latent space. This design is particularly suited to fine-grained settings where target and retained concepts share substantial semantic attributes, such as closely related objects or style--object combinations.

\textbf{Sparse Autoencoders (SAEs).} 
SAEs are feature learning methods that expand input representations into a higher-dimensional latent space, trained to reconstruct inputs using only a sparse subset of activated features~\cite{pmlr-v15-coates11a}. They have become a widely used tool for mechanistic interpretability, decomposing dense representations into more interpretable feature dictionaries~\cite{bricken2023monosemanticity, huben2024sparse}. Sparsity can be enforced through different mechanisms: ReLU-based SAEs apply a threshold but are prone to dead neurons, TopK SAEs~\cite{Makhzani2013kSparseA} fix the number of active features per forward pass, and BatchTopK SAEs~\cite{bussmann2024batchtopk} relax this constraint at the batch level for more flexible per-sample activation. Beyond standard sparsification, recent SAE variants explore alternative objectives for improving feature organization, including Matryoshka SAE~\cite{bussmann2025learning} and OrtSAE~\cite{korznikov2026ortsae}. Complementary to these approaches, SAEParate introduces a concept-level training signal that structures the latent space by concept identity.

\textbf{Contrastive Learning.}
The core idea of contrastive learning is to shape the representation space based on relative similarity, drawing semantically related samples closer while pushing distinct samples apart. This objective has proven effective in self-supervised visual representation learning~\cite{chen2020simple,he2020momentum} and vision--language alignment~\cite{radford2021learning}. With label supervision, supervised contrastive learning~\cite{khosla2020supervised} treats samples from the same class as positives, yielding more compact intra-class representations and clearer inter-class separation. Recent work has also extended contrastive objectives beyond instance-level representation learning. For example, T-SAE~\cite{bhalla2026temporal} leverages temporal relationships between neighboring language tokens to encourage semantically consistent latent activations. Our work introduces a contrastive objective in the SAE latent space to explicitly enforce concept-level separation, extending contrastive learning to structured concept separation for diffusion unlearning.

\vspace{-1mm}
\section{Problem Background}
\label{sec:problem_background}

\vspace{-1mm}

\subsection{Problem Setup}
\label{sec:problem_setup}
\vspace{-1mm}

We study concept unlearning for a pre-trained text-to-image diffusion model, where the goal is to suppress a specified target concept while preserving the model's ability to generate retained concepts. We evaluate this objective across style, object, and joint style--object unlearning, with the joint setting targeting a specific style--object combination~\cite{zhang2024unlearncanvas}. For instance, given the joint target \textit{``Dogs in Cartoon style''}, the model should suppress this specific combination while still generating related non-target concepts such as \textit{``Flowers in Cartoon style''}. We focus on SAE-based unlearning, in which an auxiliary SAE is trained on intermediate diffusion activations and target-associated sparse features are subsequently identified and suppressed at inference time. This setting requires the learned sparse representation to support fine-grained interventions even when target and retained concepts share semantic attributes.

\vspace{-1mm}

\subsection{Preliminaries: SAE-based Diffusion Unlearning}
\label{sec:sae_based_diffusion_unlearning}
\vspace{-1mm}

SAeUron~\cite{cywinski2025saeuron} instantiates SAE-based unlearning by training an SAE to reconstruct intermediate diffusion activations. Given an input activation $x \in \mathbb{R}^d$, a ReLU SAE with $\mathrm{TopK}$~\cite{Makhzani2013kSparseA, bricken2023monosemanticity, gao2025scaling} encodes it into a latent space as
\begin{equation}
\label{eq:vanilla_sae_encoder}
z = \mathrm{ReLU}\!\left(W_{\mathrm{enc}}(x - b_{\mathrm{pre}}) + b_{\mathrm{enc}}\right),
\end{equation}
where $W_{\mathrm{enc}} \in \mathbb{R}^{n \times d}$ and $W_{\mathrm{dec}} \in \mathbb{R}^{d \times n}$ denote the encoder and decoder weight matrices, respectively. From $z$, we then obtain the sparsified latent 
\begin{equation}
\label{eq:sae_topk}
\hat{z} = \mathrm{TopK}(z, k)
\end{equation}
by retaining only the top-$k$ activated features, and reconstruct the input as $\hat{x} = W_{\mathrm{dec}} \hat{z} + b_{\mathrm{pre}}$. Here, $b_{\mathrm{enc}} \in \mathbb{R}^n$ and $b_{\mathrm{pre}} \in \mathbb{R}^d$ are learnable bias terms, where $b_{\mathrm{pre}}$ centers the input before encoding and is added back after decoding.

The SAE is trained by minimizing the following objective:
\begin{align}
\label{eq:sae_loss}
\mathcal{L}_{\mathrm{SAE}}(x) = \lVert x - \hat{x} \rVert_2^2 + \alpha \mathcal{L}_{\mathrm{aux}},
\end{align}
where the first term is the reconstruction loss, $\mathcal{L}_{\mathrm{aux}}$ is an auxiliary term that mitigates inactive latent units, and $\alpha$ controls its contribution.

Given a trained SAE, a set of concept-relevant SAE features, $\mathcal{F}_c$, is identified for each target concept $c$ based on feature-importance scores after training. During inference, activations from the corresponding diffusion layer are encoded into the SAE latent space, where the selected features in $\mathcal{F}_c$ are suppressed with negative multipliers $\gamma_c$. The modified features are then reconstructed through the SAE decoder. Further details on SAE-based diffusion unlearning are provided in Section~\ref{sec:unlearning_procedure}.

\begin{figure}[t]
\vspace{-6mm}
    \centering
    \includegraphics[width=1.0\linewidth]{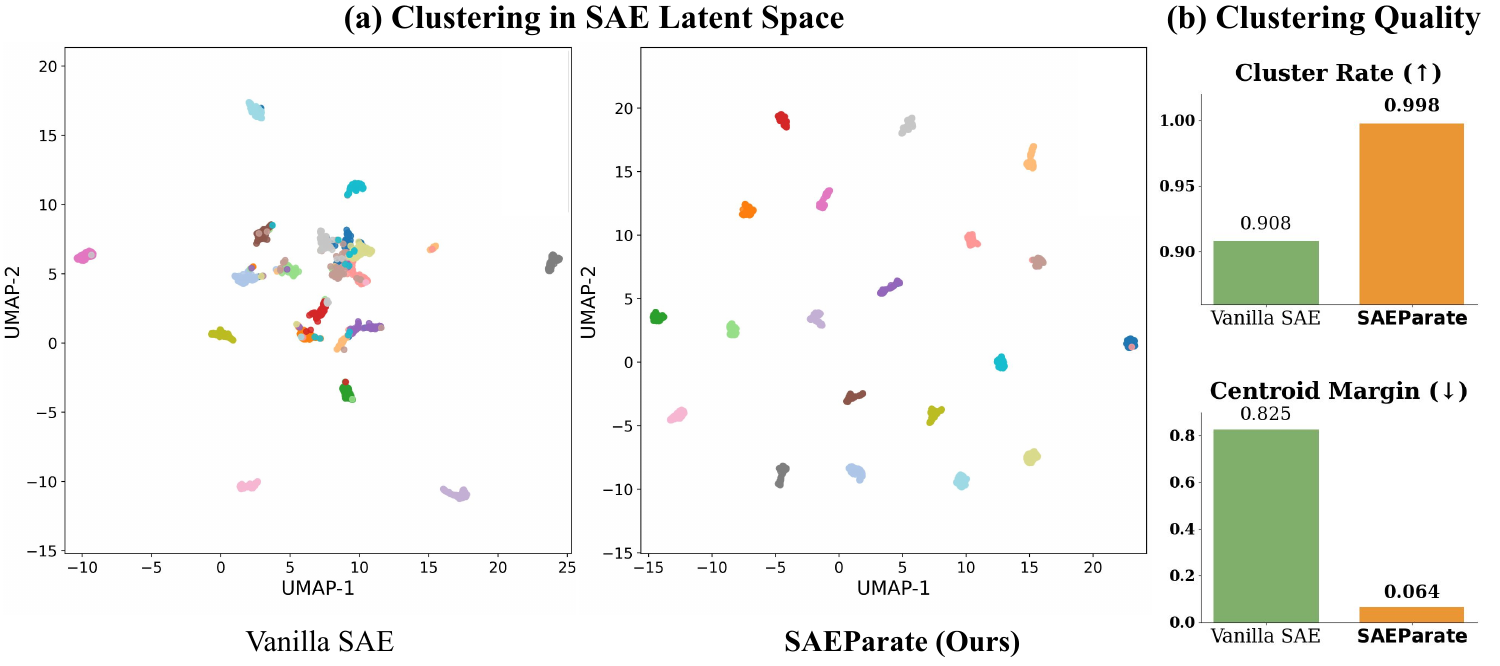}
    \caption{Visualization of SAE latent representations. (a) UMAP~\cite{mcinnes2018umap} projections of Vanilla SAE (SAeUron) and SAEParate on UnlearnCanvas~\cite{zhang2024unlearncanvas}, with each color representing a different concept. (b) Quantitative comparison of cluster rate and centroid margin, which measure centroid-based class assignment and inter-centroid similarity, respectively. Both results demonstrate that SAEParate produces clearer concept-wise clusters with reduced overlap, indicating improved concept separation.}
    \label{fig:sae_latent_clusters}
\vspace{-3mm}
\end{figure}

\subsection{Motivation: Latent Entanglement in SAE-based Unlearning}
\label{sec:latent_entanglement}

However, SAE-based unlearning inherits a key limitation of SAE training, whose objective primarily encourages sparse reconstruction but does not explicitly enforce concept-wise separation. As a result, recent studies suggest that sparse autoencoders often learn shared or compositional features, where correlated attributes are encoded by shared latents to preserve reconstruction quality with fewer active features~\cite{leask2025sparseautoencoderscanonicalunits,bussmann2025learning}. Figure~\ref{fig:sae_latent_clusters}(a) illustrates this behavior in our setting, where representations of different concepts substantially overlap in the latent space. Suppressing concept-relevant features $\mathcal{F}_c$ can therefore also affect retained concepts that rely on overlapping features, degrading the unlearning--retention trade-off (Section~\ref{sec:experiments}). This motivates explicitly structuring SAE representations according to concept identity.
\vspace{-1mm}

\section{Method}
\label{sec:method}
\vspace{-1mm}

\begin{figure}[t]
    \vspace{-4mm}
    \centering
    \includegraphics[width=0.85\linewidth]{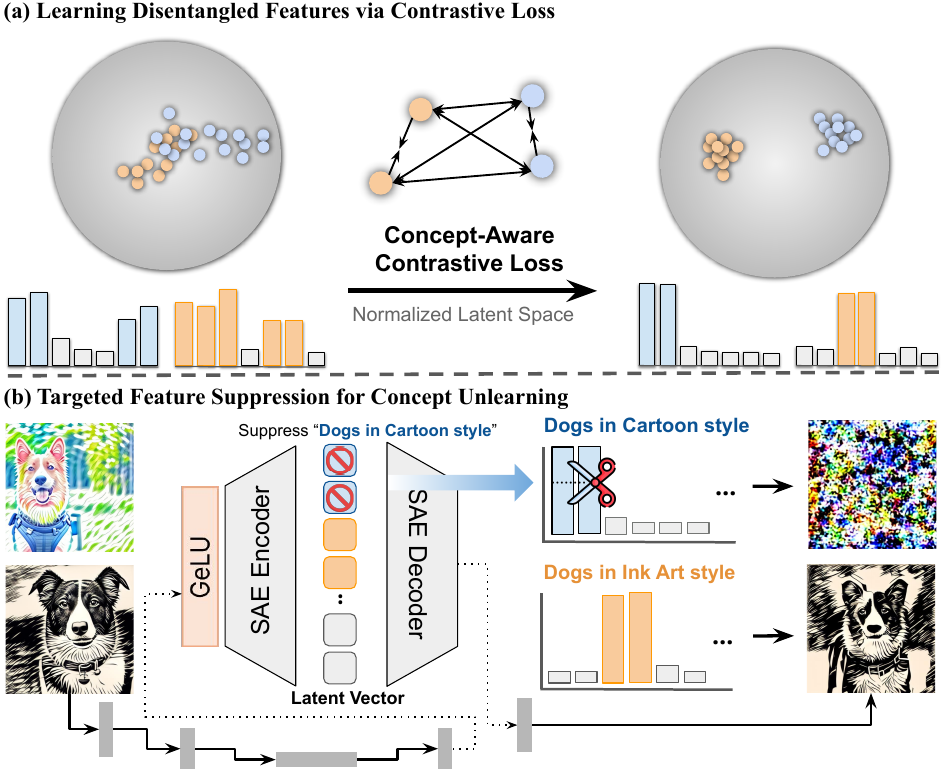}
    \caption{\textbf{Overview of SAEParate.} We perform concept unlearning in text-to-image diffusion models by suppressing concept-specific features in a Sparse Autoencoder (SAE). \textbf{(a) Training.} A concept-aware contrastive loss is applied to normalized SAE latents, pulling together representations of samples with the same concept label and pushing apart those with different labels, yielding well-separated concept-wise activation patterns. \textbf{(b) Inference-time unlearning.} Given a target concept (e.g., \textit{Dogs in Cartoon style}), we identify and suppress the corresponding SAE features in the latent vector before decoding, erasing the target combination while leaving non-target concepts (e.g., \textit{Dogs in Ink Art style}) intact.}
    \label{fig:method_overview}
    \vspace{-3mm}
\end{figure}

We propose \textbf{SAEParate}, which trains the SAE so that concept identity is explicitly encoded in its latent space, allowing semantically overlapping concepts to be separated at the representation level. Section~\ref{sec:supcon} introduces concept-aware spatial aggregation, which focuses supervision on concept-relevant positions, and a latent-space contrastive objective that structures the aggregated representations by concept. Section~\ref{sec:nonlinear_encoder} enhances the SAE encoder with an additional nonlinear transformation, providing the expressivity needed to reorganize entangled activations under this objective. Section~\ref{sec:unlearning_procedure} describes how the resulting representation is used for inference-time unlearning. An overview of SAEParate is shown in Figure~\ref{fig:method_overview}.


\subsection{Concept-Discriminative Sparse Representation Learning}
\label{sec:supcon}
    \vspace{-2mm}

We introduce a \textit{concept-aware contrastive objective in the SAE latent space}, designed to explicitly separate concept-level activation patterns. While inspired by supervised contrastive (SupCon) learning~\cite{khosla2020supervised}, our formulation departs from the standard setting by operating on concept-conditioned representations aggregated from SAE latent activations and using diffusion timesteps to construct multiple views.

\textbf{Multiviewed Batch.} 
Let $x(t)^{c} \in \mathbb{R}^{S \times d}$ denote the diffusion activation of concept $c$ at timestep $t$, where $S = h \times w$ is the number of spatial positions in the feature map and $d$ is the dimensionality of each spatial activation. Unlike the standard SupCon framework, which constructs positive pairs using random augmentations of the same sample, our setting naturally provides multiple views through the temporal structure of the diffusion process. Specifically, for each concept $c$, we sort activations by timestep and construct adjacent-timestep pairs $\{(x(t_0)^c, x(t_1)^c), (x(t_2)^c, x(t_3)^c), \dots\}$, where each pair is treated as a positive pair because both activations share the same concept label and arise from neighboring denoising steps. Following this construction, each activation in a training batch is accompanied by its adjacent-timestep counterpart, resulting in a multiviewed batch \cite{khosla2020supervised}. As a result, a batch of $N$ samples forms $M$ positive pairs with $N = 2M$. This adjacent-timestep construction ensures that each sample has at least one same-concept positive view, enabling stable and effective concept-aware contrastive learning.

\textbf{Concept-Aware Spatial Aggregation.}
Given the resulting multi-view batch, each diffusion activation is denoted by $x_i \in \mathbb{R}^{S \times d}$ for $i \in \mathcal{I} \equiv \{1, \dots, N\}$. For each spatial position $p \in \{1, \dots, S\}$, we map $x_{i,p}$ into an $n$-dimensional latent representation using the SAE encoder $f_{\theta}$:
\begin{align}
\label{eq:supcon_z}
z_{i,p}=f_{\theta}(x_{i,p}),
\end{align}
where $f_{\theta}$ denotes the nonlinearity-enhanced encoder introduced in Section~\ref{sec:nonlinear_encoder}, and $z_{i,p}$ is taken before the $\mathrm{TopK}$ sparsification.
To obtain a sample-level representation, we aggregate the spatial latents $\{z_{i,p}\}_{p=1}^{S}$. Since many spatial positions correspond to concept-irrelevant regions (e.g., background or unrelated objects), uniformly pooling all positions would introduce noise. We therefore weight each position by its cross-attention relevance to the concept:
\begin{align}
\label{eq:cross_attention_weight}
\alpha_{i,p}
=
\frac{\sum_{k \in \mathcal{C}_i}\left(\frac{1}{H}\sum_{h=1}^{H} A_{i,h,p,k}\right)}
{\sum_{p'=1}^{S}\sum_{k \in \mathcal{C}_i}\left(\frac{1}{H}\sum_{h=1}^{H} A_{i,h,p',k}\right)},
\end{align}
where $A_{i,h,p,k}$ denotes the cross-attention probability from spatial position $p$ to text token $k$ at head $h$, $H$ is the number of attention heads, and $\mathcal{C}_i$ is the set of prompt tokens corresponding to the concept of sample $i$. The weights satisfy $\sum_{p=1}^{S}\alpha_{i,p}=1$, and the sample-level representation is
\begin{align}
\label{eq:supcon_r}
r_i
=
\sum_{p=1}^{S}\alpha_{i,p} z_{i,p}.
\end{align}
Concept-wise visualizations of Eq.~\ref{eq:cross_attention_weight} and an ablation on cross-attention weighted pooling are provided in Appendix~\ref{app:cross_attention_weight}.

\textbf{Latent-Space Contrastive Objective.}
\label{sec:lsc}
Let $y_i$ denote the label of sample $x_i$, and define the positive set for anchor $i$ as $\mathcal{P}(i)=\{j \neq i \mid y_j = y_i\}$. For example, in joint style--object unlearning, two activations generated from the same style--object concept (e.g., \textit{Dogs in Cartoon style}) form a positive pair, whereas activations from different concepts (e.g., \textit{Dogs in Ink Art style}, \textit{Flowers in Cartoon style}) form negative pairs, even when they share partial attributes. We $\ell_2$-normalize each sample-level representation as $u_i = r_i / \|r_i\|_2$, so that $u_i^\top u_j$ is the cosine similarity between two samples. The latent-space contrastive loss $\mathcal{L}_{\mathrm{LSC}}$ is then defined as
\vspace{-1mm}
\begin{align}
\mathcal{L}_{\mathrm{LSC}}
=
-\frac{1}{|\mathcal{I}|}
\sum_{i\in\mathcal{I}}
\frac{1}{|\mathcal{P}(i)|}
\sum_{j\in\mathcal{P}(i)}
\log
\frac{\exp(u_i^\top u_j/\tau)}
{\sum_{k\neq i}\exp(u_i^\top u_k/\tau)},
\label{eq:supcon}
\end{align}
where $\tau > 0$ denotes the temperature parameter. This objective encourages activation patterns from the same concept to form compact clusters while remaining well separated from those of other concepts. We make minor extensions to Eq.~\ref{eq:supcon} for joint style--object unlearning in Section~\ref{par:joint_unlearning}.

\textbf{Training Objective.}
The final training objective combines the SAE objective from Eq.~\ref{eq:sae_loss} with the proposed latent-space contrastive regularization of Eq.~\ref{eq:supcon}, where $\lambda$ controls the contribution of the contrastive term:
\begin{align}
    \mathcal{L}_{\mathrm{total}}
    =
    \mathcal{L}_{\mathrm{SAE}}
    +
    \lambda \mathcal{L}_{\mathrm{LSC}}.
\end{align}

\subsection{Nonlinearity-Enhanced SAE Encoder}
\label{sec:nonlinear_encoder}

The latent representations of the SAE encoder in Eq.~\ref{eq:vanilla_sae_encoder} are obtained by applying a ReLU to a single affine transformation. Although this formulation is simple and effective for reconstruction, its representational capacity remains limited by its shallow structure, making it insufficient to fully accommodate the discriminative pressure imposed by the latent-space contrastive loss. 

To increase the expressive power of the encoder, we enhance it with an additional hidden layer followed by a nonlinear activation $\sigma$, introducing a deeper nonlinear transformation before sparse feature selection. The modified version of Eq.~\ref{eq:vanilla_sae_encoder} becomes
\begin{align}
\label{eq:gelu_encoder}
z &= \mathrm{ReLU}\!\left(
W_{\mathrm{enc2}} \, \sigma\!\left(
W_{\mathrm{enc1}} (x - b_{\mathrm{pre}}) + b_{\mathrm{enc1}}
\right) + b_{\mathrm{enc2}}
\right),
\end{align}
where $W_{\mathrm{enc1}} \in \mathbb{R}^{d \times d}$ and $W_{\mathrm{enc2}} \in \mathbb{R}^{n \times d}$ denote the weights of the additional hidden layer and the final encoder projection, and $b_{\mathrm{enc1}} \in \mathbb{R}^{d}$ and $b_{\mathrm{enc2}} \in \mathbb{R}^{n}$ are the corresponding biases. We then obtain the sparsified latent $\hat{z}$ by applying the $\mathrm{TopK}$ operation in Eq.~\ref{eq:sae_topk}, and use this encoder consistently during both training and inference. The decoder remains linear, adding expressivity only to the encoder while preserving sparsification and inference-time ablation.

For $\sigma$, we adopt GeLU~\cite{hendrycks2016gaussian}, defined as $\mathrm{GeLU}(x)=x\Phi(x)$ with $\Phi$ the standard Gaussian CDF. Appendix~\ref{app:encoder_architecture_ablation} compares linear, ReLU, and GeLU hidden layers, showing that GeLU further improves target-concept removal while maintaining comparable retention.

\vspace{-1.5mm}

\subsection{Unlearning Procedure}
\label{sec:unlearning_procedure}
\vspace{-1.5mm}

After completing the training procedure of SAEParate, we perform concept unlearning via an inference-time intervention in the concept-separated latent space learned by the SAE. For each target concept $c$, we compute feature importance scores using feature activations from target and non-target samples, and select the top-$\tau_c$ concept-relevant features as $\mathcal{F}_c$. For example, for the concept \textit{Dogs in Cartoon style}, this step identifies features that are strongly activated by this concept compared to non-target concepts. During inference, we encode the diffusion activation from the layer on which the SAE was trained into the SAEParate latent space, and suppress the selected features in $\mathcal{F}_c$ by multiplying their activations with a negative multiplier $\gamma_c$. The main cost of SAEParate therefore comes from offline SAE training and feature selection, while unlearning itself only requires an inference-time intervention. Further details of the unlearning procedure are provided in Appendix~\ref{app:saeuron_alg_details}.

\vspace{-2mm}
\section{Experiments}\label{sec:experiments}
\vspace{-2mm}

We evaluate our method under the 50-style, 20-object evaluation
setting of UnlearnCanvas~\cite{zhang2024unlearncanvas}, using its
fine-tuned Vision Transformer-based~\cite{dosovitskiy2021an}
classifiers for evaluation. The benchmark also provides a
Stable Diffusion v1.5~\cite{rombach2022high} checkpoint fine-tuned
on UnlearnCanvas, which we use as the basis for unlearning evaluation.

\textbf{Baselines.} We compare SAEParate against nine existing T2I diffusion unlearning methods from the UnlearnCanvas benchmark~\cite{zhang2024unlearncanvas}: ESD~\cite{gandikota2023erasing}, FMN~\cite{zhang2024forget}, UCE~\cite{gandikota2024unified}, CA~\cite{kumari2023ablating}, SalUn~\cite{fan2024salun}, SEOT~\cite{li2024get}, SPM~\cite{lyu2024one}, EDiff~\cite{Wu_2025_CVPR}, and SHS~\cite{wu2024scissorhands}. We additionally include two SAE-based methods closely related to ours, SAeUron~\cite{cywinski2025saeuron} and SAEmnesia~\cite{cassano2026saemnesia}. 

\textbf{Performance Metrics.} Following the UnlearnCanvas~\cite{zhang2024unlearncanvas} protocol, we report Unlearning Accuracy (\textbf{UA}), In-domain Retain Accuracy (\textbf{IRA}), Cross-domain Retain Accuracy (\textbf{CRA}), and FID~\cite{heusel2017gans} for single-attribute unlearning. For joint style--object unlearning, in addition to UA, we report Style Consistency (\textbf{SC}), Object Consistency (\textbf{OC}), and Unrelated Prompting (\textbf{UP}), where UP measures generation quality on non-target (style, object) combinations.

\textbf{Implementation Details.} For fair comparison with SAeUron \citep{cywinski2025saeuron}, we keep the SAE application location, training data construction, and feature-calibration pipeline unchanged, and modify only the SAE training objective and encoder architecture. Following SAeUron, we apply the SAE to \texttt{up.1.2} for style unlearning and \texttt{up.1.1} for object unlearning. For joint unlearning, we uniformly use \texttt{up.1.1} for all SAEs. We also use the same anchor-prompt-based training and calibration setup, while keeping the training prompts disjoint from the UnlearnCanvas evaluation prompts. Specifically, we use 80 one-sentence anchor prompts for each of the 20 object concepts, appending \textit{``in \{style\} style.''} so that the SAE is exposed to benchmark styles during training. For calibration, we use 20 prompts per style for style unlearning, 80 prompts per object for object unlearning, and 80 prompts per (object, style) combination for joint unlearning. The contrastive objective for joint unlearning and the feature scoring/suppression pipeline are detailed in Appendices~\ref{app:joint_unlearning_formulation} and~\ref{app:saeuron_alg_details}, respectively.

\vspace{-1.5mm}

\subsection{Main Experimental Results}
\vspace{-1.5mm}

\textbf{Single-Attribute Concept Unlearning.} 
We first perform single-attribute concept unlearning by removing all samples from either a target class or a target style (i.e., one attribute at a time). 
We use two hyperparameters for each target concept \(c\): \(\tau_c\), the number of selected features in the ablation set \(\mathcal{F}_c\), and \(\gamma_c\), a negative multiplier controlling the suppression strength of each selected feature. Following the SAeUron setup, we fix \(\tau_c = 1\) and \(\gamma_c = -0.1\) for all style concepts, while tuning them on a validation set for object unlearning. The selected values are reported in Appendix~\ref{app:hyperparams}.

\begin{table*}[t]
\vspace{-4mm}
\centering
\caption{Comparison of diffusion unlearning methods on style, object, and joint unlearning benchmarks. The best and second-best results in each column are highlighted in green and underlined, respectively. Notably, SAEParate demonstrates strong unlearning performance across all unlearning settings, with particularly pronounced gains on object and joint unlearning.}
\label{tab:main_results_grouped}
\vspace{2pt}
\small
\setlength{\tabcolsep}{4pt}
\resizebox{\textwidth}{!}{%
\begin{tabular}{l|ccc|ccc|c|c|ccccc|c}
\toprule
\multirow{2}{*}{Method}
& \multicolumn{3}{c|}{Style Unlearning}
& \multicolumn{3}{c|}{Object Unlearning}
& \multirow{2}{*}{Avg. ($\uparrow$)}
& \multirow{2}{*}{FID ($\downarrow$)}
& \multicolumn{5}{c|}{Joint Unlearning}
& \multirow{2}{*}{\textbf{Total Avg.} ($\uparrow$)} \\
\cmidrule(lr){2-4} \cmidrule(lr){5-7} \cmidrule(lr){10-14}
& UA ($\uparrow$) & IRA ($\uparrow$) & CRA ($\uparrow$)
& UA ($\uparrow$) & IRA ($\uparrow$) & CRA ($\uparrow$)
&
&
& UA ($\uparrow$) & SC ($\uparrow$) & OC ($\uparrow$) & UP ($\uparrow$) & Avg. ($\uparrow$)
& \\
\midrule
ESD
& \underline{98.58\%} & 80.97\% & 93.96\%
& 92.15\% & 55.78\% & 44.23\%
& 77.61\%
& 65.55
& \underline{91.42\%} & 4.88\% & 14.72\% & 84.38\% & 48.85\%
& 66.11\% \\
FMN
& 88.48\% & 56.77\% & 46.60\%
& 45.64\% & 90.63\% & 73.46\%
& 66.93\%
& 131.37
& 45.37\% & \underline{68.73\%} & 62.74\% & 83.25\% & 65.02\%
& 66.17\% \\
UCE
& 98.40\% & 60.22\% & 47.71\%
& 94.31\% & 39.35\% & 34.67\%
& 62.44\%
& 182.01
& 75.97\% & 4.53\% & 5.72\% & 35.42\% & 30.41\%
& 49.63\% \\
CA
& 60.82\% & 96.01\% & 92.70\%
& 46.67\% & 90.11\% & 81.97\%
& 78.05\%
& \cellcolor{green!30}\textbf{54.21}
& 47.92\% & 10.08\% & 56.35\% & 81.54\% & 48.97\%
& 66.42\% \\
SalUn
& 86.26\% & 90.39\% & 95.08\%
& 86.91\% & \cellcolor{green!30}\textbf{96.35\%} & \cellcolor{green!30}\textbf{99.59\%}
& 92.43\%
& 61.05
& 42.21\% & 62.45\% & 70.93\% & \underline{87.28\%} & \underline{65.72\%}
& \underline{81.75\%} \\
SEOT
& 56.90\% & 94.68\% & 84.31\%
& 23.25\% & 95.57\% & 82.71\%
& 72.90\%
& 62.38
& 29.32\% & 45.31\% & 53.64\% & 85.45\% & 53.43\%
& 65.11\% \\
SPM
& 60.94\% & 92.39\% & 84.33\%
& 71.25\% & 90.79\% & 81.65\%
& 80.23\%
& 59.79
& 45.72\% & 41.34\% & 36.32\% & 67.82\% & 47.80\%
& 67.26\% \\
EDiff
& 92.42\% & 73.91\% & 98.93\%
& 86.67\% & 94.03\% & 48.48\%
& 82.41\%
& 81.42
& 71.33\% & 35.23\% & 26.32\% & 51.52\% & 46.10\%
& 67.89\% \\
SHS
& 95.84\% & 80.42\% & 43.27\%
& 80.73\% & 81.15\% & 67.99\%
& 74.90\%
& 119.34
& 55.32\% & 14.34\% & 24.32\% & 83.95\% & 44.48\%
& 62.73\% \\
SAeUron
& 95.80\% & \underline{99.10\%} & \cellcolor{green!30}\textbf{99.40\%}
& 78.82\% & 95.47\% & \underline{95.58\%}
& 94.03\%
& 62.15
& 63.00\% & 16.98\% & \underline{75.84\%} & 71.90\% & 56.93\%
& 79.19\% \\
SAEmnesia$^{\dagger}$
& 96.60\% & 98.67\% & \underline{99.30\%}
& \underline{94.65\%} & 91.39\% & 88.48\%
& \underline{94.85\%}
& \underline{56.15}
& 81.00\% & 59.58\% & 11.18\% & 54.60\% & 51.59\%
& 77.55\% \\
\textbf{SAEParate}
& \cellcolor{green!30}\textbf{99.60\%} & \cellcolor{green!30}\textbf{99.24\%} & 98.30\%
& \cellcolor{green!30}\textbf{95.20\%} & \underline{95.91\%} & 94.26\%
& \cellcolor{green!30}\textbf{97.09\%}
& 63.98
& \cellcolor{green!30}\textbf{94.11\%} & \cellcolor{green!30}\textbf{93.49\%} & \cellcolor{green!30}\textbf{91.66\%} & \cellcolor{green!30}\textbf{98.82\%} & \cellcolor{green!30}\textbf{94.52\%}
& \cellcolor{green!30}\textbf{96.06\%} \\
\bottomrule
\end{tabular}%
}
\begin{flushleft}
\footnotesize
$^{\dagger}$ SAEmnesia does not originally consider joint style--object unlearning; we adapt it to this setting as detailed in Appendix~\ref{app:saemnesia_details}.
\end{flushleft}
\vspace{-6mm}
\end{table*}


As shown in Table~\ref{tab:main_results_grouped}, compared with the baseline results reported in the UnlearnCanvas benchmark, SAEParate achieves the best average performance and outperforms prior methods on most metrics for both style and object unlearning. Notably, it preserves non-target concepts well, as reflected in IRA and CRA, while improving UA. The competitive FID further indicates that retain-set image quality is preserved at the distributional level. These results suggest that SAEParate learns a concept-discriminative latent space that mitigates redundant feature selection and enables more precise suppression of target-relevant features. SAEParate also improves efficiency over other baselines while remaining competitive with SAeUron, as shown in Table~\ref{tab:efficiency_comparison_small}. Additional details are in Appendix~\ref{app:training_memory_and_storage}.

\begin{wraptable}{r}{0.3\textwidth}
\vspace{-5mm}
\centering
\scriptsize
\caption{Efficiency comparison. Memory is measured with a batch size of 4096 following SAeUron. Storage denotes additional method-specific files.}
\setlength{\tabcolsep}{3pt}
\begin{tabular}{lcc}
\toprule
Method & Memory & Storage \\
       & (GB)$\downarrow$ & (GB)$\downarrow$ \\
\midrule
ESD     & 17.8 & 4.3 \\
FMN     & 17.9 & 4.2 \\
UCE     & 5.1  & 1.7 \\
CA      & 10.1 & 4.2 \\
SalUn   & 30.8 & 4.0 \\
SEOT    & 7.34 & 0.0 \\
SPM     & 6.9  & 0.0 \\
EDiff   & 27.8 & 4.0 \\
SHS     & 31.2 & 4.0 \\
SAeUron & 2.8  & 0.2 \\
\textbf{SAEParate} & 3.4 & 0.2 \\
\bottomrule
\end{tabular}
\label{tab:efficiency_comparison_small}
\vspace{-3mm}
\end{wraptable}

\textbf{Joint Style-Object Concept Unlearning.} 
\label{par:joint_unlearning}
We further evaluate on the joint style-object unlearning setting from UnlearnCanvas, where the target is a specific style-object pair (e.g., \textit{Dogs in Cartoon style}), while partially matched prompts (e.g., the same object in different styles or different objects in the same style) must be preserved. This requires finer-grained selectivity than single-attribute removal, providing a stricter test of whether the learned SAE representations are sufficiently concept-specific. For this setting, we retain the SAEParate framework and  adapt Eq.~\ref{eq:supcon} to the style-object level: positives are defined at the style-object combination level, while partially matched samples are treated as hard negatives which get more separation pressure. Further details are provided in Appendix~\ref{app:joint_unlearning_formulation}.

As shown in Table~\ref{tab:main_results_grouped}, SAEParate achieves the best performance across all four
joint unlearning metrics, improving the average score from 65.72\% to 94.52\% over the strongest baseline. SAEParate achieves effective UA while maintaining OC and SC, demonstrating that our method performs surgical rather than destructive intervention.

Figure~\ref{fig:feature_overlap_viz} further provides qualitative comparisons with SAeUron, showing cleaner removal of the target combination while better preserving partially matched prompts. These results suggest that the effectiveness of SAEParate extends beyond single-concept removal to finer-grained unlearning settings that require more selective and concept-specific representations. We provide more qualitative results in Appendix~\ref{app:quality_eval_results}. 
\vspace{-2.5mm}

\subsection{Further Analyses}
\label{sec:further_analyses}
\vspace{-1.5mm}

\begin{figure}[h]
\vspace{-2mm}
    \centering
    \begin{minipage}[c]{0.49\linewidth}
        \centering
        \includegraphics[width=\linewidth]{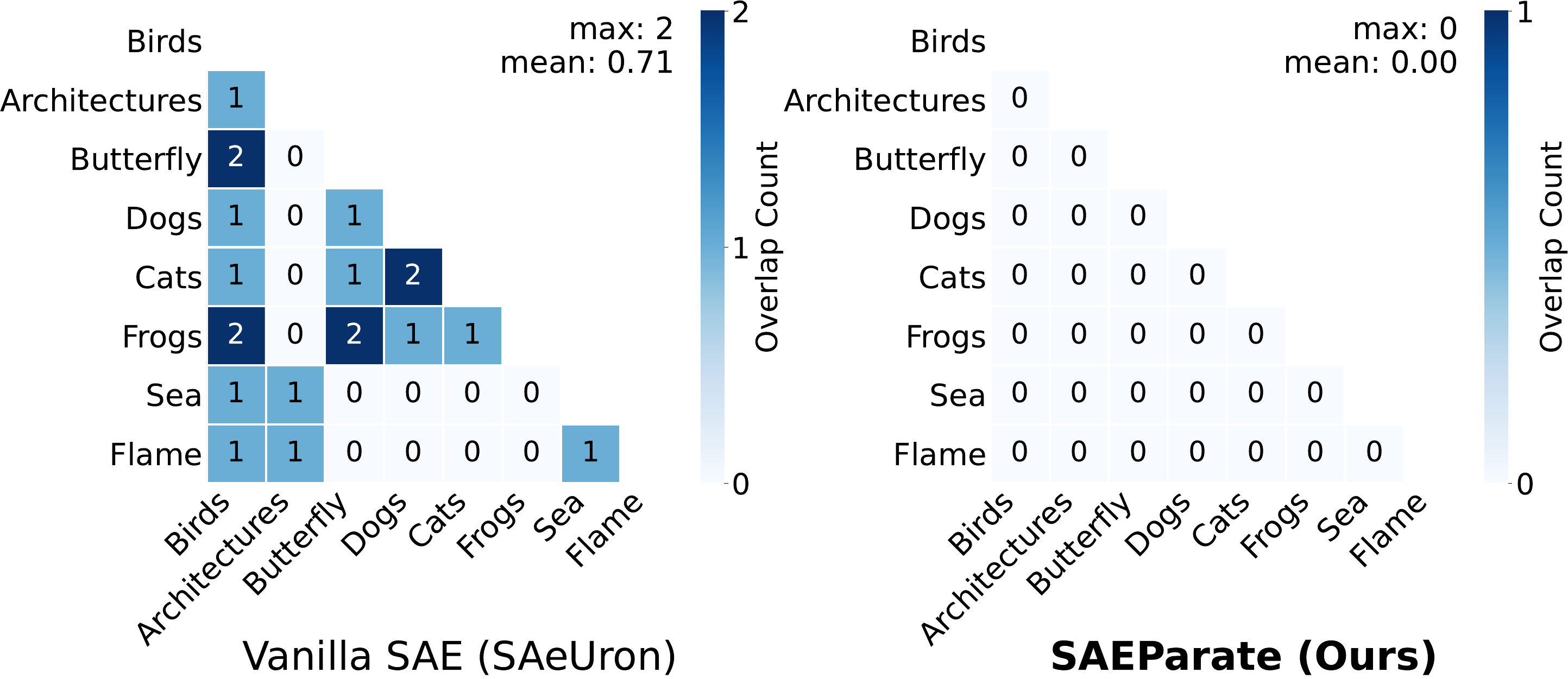}
        \caption*{(a) Pairwise overlap counts between selected class-specific feature sets.}
    \end{minipage}
    \hfill
    \begin{minipage}[c]{0.49\linewidth}
        \centering
        \includegraphics[width=\linewidth]{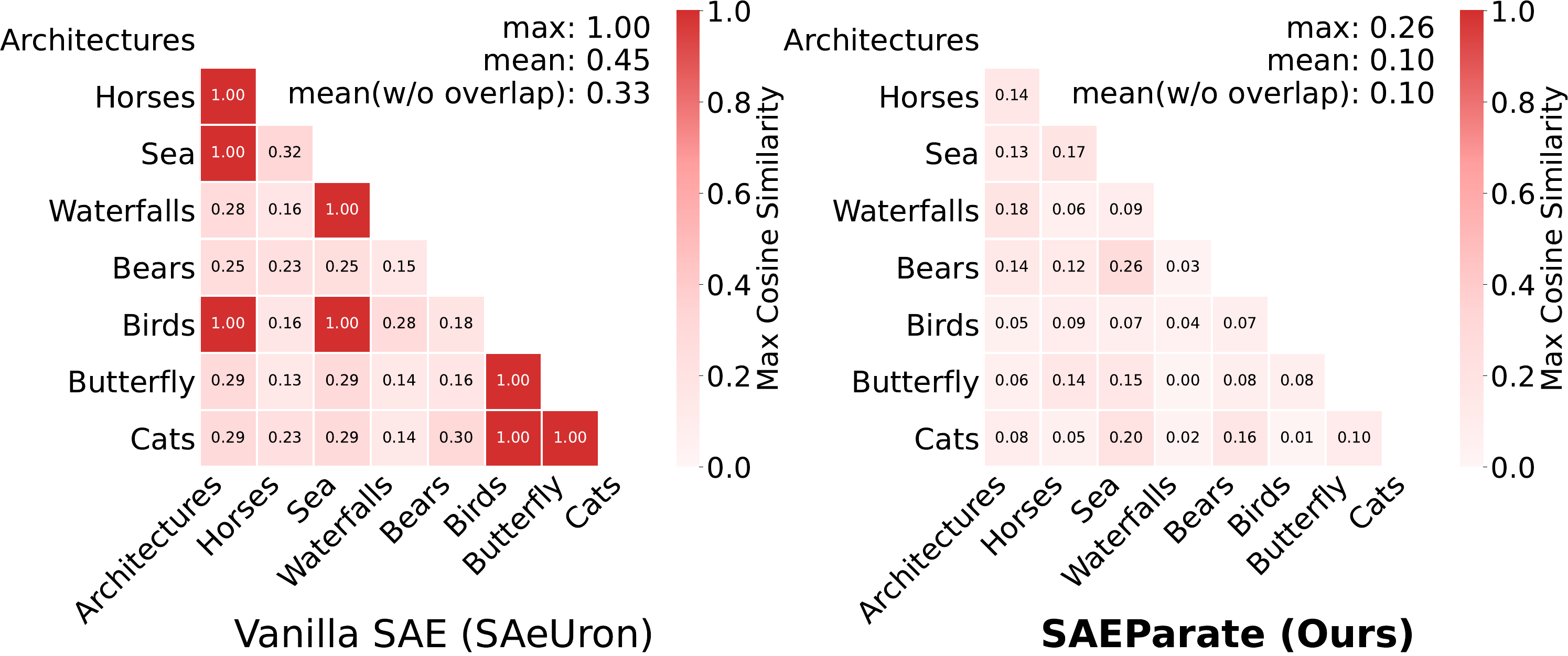}
        \caption*{(b) Pairwise maximum cosine similarity between selected class-specific feature groups.}
    \end{minipage}
    \vspace{-2mm}
    \caption{Class-wise overlap analysis comparing Vanilla SAE (left) and SAEParate (right).}
\label{fig:overlap_details_heatmap}
\vspace{-3mm}
\end{figure}

\textbf{Overlap Analysis.} To directly assess whether the proposed method mitigates redundant feature selection, we perform a pairwise analysis of class-specific selected feature groups. We measure both the number of overlapping selected features and the maximum cosine similarity between feature groups, capturing overlap in feature identity as well as similarity in their decoder directions. SAEParate reduces feature overlap to zero and substantially lowers pairwise similarity (Figure~\ref{fig:overlap_details_heatmap}), indicating more distinct intervention features and a reduced risk of affecting retained concepts during target suppression. We provide the full class-wise overlap analysis, including the more challenging joint style--object setting, in Appendix~\ref{app:full_overlap_analysis}.

\textbf{Concept-wise $k$-sparse Probing.}
To assess how compactly concept-discriminative information is encoded in the SAE latent space, we perform concept-wise $k$-sparse probing~\cite{gurnee2023finding}. For each concept, we select the top-$k$ latents according to the feature-importance score used for unlearning, and train a linear probe with cross-entropy loss to predict concept labels using only the union of these selected latents. 

\begin{figure}[t]
\vspace{-3mm}
    \centering
    \includegraphics[width=1\linewidth]{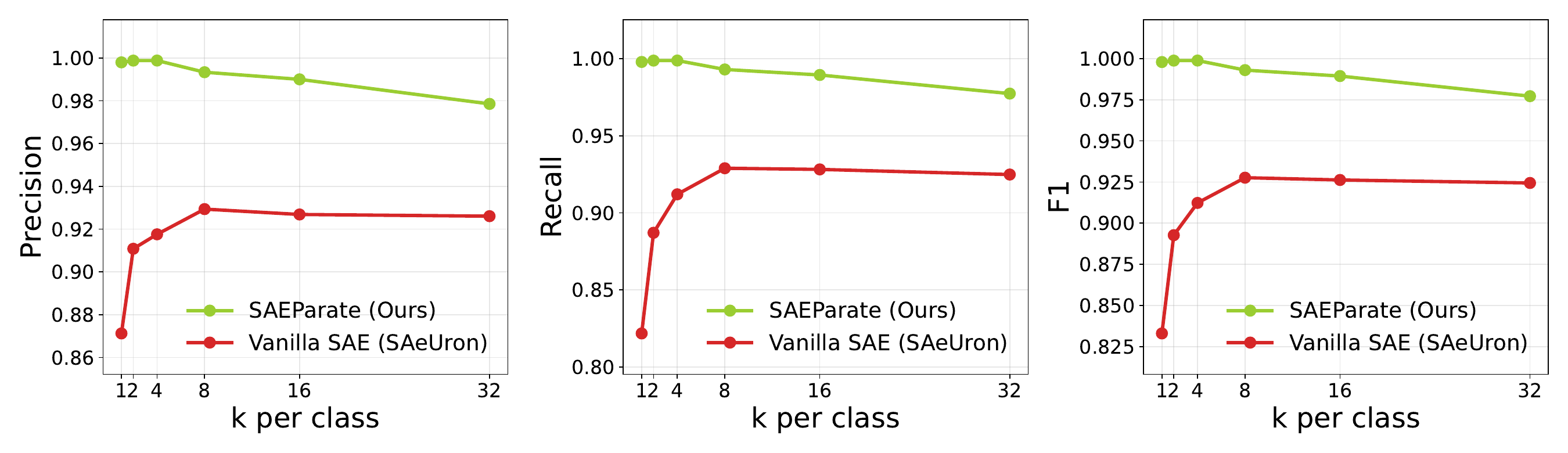}
    \vspace{-5mm}
    \caption{Concept-wise $k$-sparse probing comparison on UnlearnCanvas object concepts.}
    \label{fig:k-sparse_probing}
\vspace{-4mm}
\end{figure}

As shown in Figure~\ref{fig:k-sparse_probing}, SAEParate achieves strong classification scores even with $k=1$ or $k=2$, whereas SAeUron that uses Vanilla SAE requires larger $k$ and still underperforms. These results indicate that SAEParate concentrates concept-discriminative information into a small number of latent features, reducing redundancy and improving controllability.

\begin{wraptable}{r}{0.6\linewidth}
   \vspace{-5mm}
    \centering
    \caption{Quantitative ablation results on object unlearning.}
    \label{tab:saeparate_ablation}
    \vspace{-4pt}
    \small
    \setlength{\tabcolsep}{3pt}
    \renewcommand{\arraystretch}{1.05}
    \resizebox{\linewidth}{!}{%
    \begin{tabular}{l|ccc|c|cc}
        \toprule
        \multirow{2}{*}{Variants}
        & \multicolumn{3}{c|}{Unlearning Performance}
        & \multirow{2}{*}{Avg. ($\uparrow$)}
        & \multicolumn{2}{c}{Cluster Quality} \\
        \cmidrule(lr){2-4} \cmidrule(lr){6-7}
        & UA ($\uparrow$)
        & IRA ($\uparrow$)
        & CRA ($\uparrow$)
        &
        & Cluster Rate ($\uparrow$)
        & Centroid Margin ($\downarrow$) \\
        \midrule
        Vanilla SAE
        & 58.14\% & 96.74\% & 97.45\% & 84.11\%
        & 0.93 & 0.73 \\
        \textit{only}-$\mathrm{LSC}$
        & 83.82\% & 95.23\% & 91.55\% & 90.20\%
        & 0.98 & 0.71 \\
        \textit{only}-GeLU
        & 47.55\% & 96.69\% & 97.53\% & 80.59\%
        & 0.94 & 0.64 \\
        \textbf{SAEParate}
        & 95.59\% & 96.48\% & 93.46\% & 95.18\%
        & 1.00 & 0.06 \\
        \bottomrule
    \end{tabular}
    }
    \vspace{-3mm}
\end{wraptable}

\textbf{Ablation Study.} To analyze the contributions of the contrastive objective and nonlinear encoder, we conduct an ablation study on contrastive learning (Section~\ref{sec:lsc}) and the nonlinearity-enhanced encoder (Section~\ref{sec:nonlinear_encoder}) under the subset setting (Appendix~\ref{app:subset_setting}). We evaluate two ablated variants, 
\textit{only}-$\mathrm{LSC}$ and \textit{only}-GeLU, together with the full SAEParate model. The quantitative results in Table~\ref{tab:saeparate_ablation} show that the full model achieves the best overall performance on object unlearning. Moreover, the cluster quality indicates that the clusters remain only partially separated when using either \textit{only}-$\mathrm{LSC}$ or \textit{only}-GeLU, whereas combining both components yields substantially clearer concept-wise clusters. 

\textbf{Nudity Unlearning on I2P Benchmark.}
We further evaluate SAEParate on safety-critical nudity unlearning using the I2P benchmark~\cite{schramowski2023safe},
targeting \textit{naked woman} and \textit{naked man} as distinct concepts.
SAEParate achieves strong gender-specific unlearning accuracy while retaining the opposite-gender detections,
demonstrating its ability to separate fine-grained concepts.
Full results are in Appendix~\ref{app:nudity_unlearning}.

\textbf{Generalization to Unseen Concepts and Sequential Unlearning.}
SAEParate also generalizes beyond the standard setup. When the SAE is trained on only half of the styles, it improves unlearning on unseen styles and reduces cross-domain interference. In a sequential setting where five objects are removed consecutively, it avoids the retention collapse observed in SAeUron. Full results are provided in Appendices~\ref{app:unseen_evaluations} and~\ref{app:object_sequential_unlearning}.

\textbf{Other Results.} Additional appendix experiments demonstrate robustness to concept-restoration attacks and prompt variations (Appendix~\ref{app:robustness}), compatibility with Matryoshka SAEs (Appendix~\ref{app:matryoshka_compatibility}), and additional qualitative generation results (Appendix~\ref{app:quality_eval_results}). SAEParate also generalizes beyond the standard benchmark setting, enabling selective \textit{Golden Retriever} removal under coarse binary forget/retain supervision (Appendix~\ref{app:golden-retriever}) and transferring to the MM-DiT-based SD3.5 Medium Turbo under a preliminary adaptation, where it outperforms SAeUron in UA and overall average (Appendix~\ref{app:mmdit_extension}).

\vspace{-1.5mm}

\section{Conclusion}
\label{sec:conclusion}
\vspace{-1.5mm}

We proposed \textbf{SAEParate}, which learns concept-discriminative representations for diffusion unlearning through concept-aware spatial aggregation, contrastive representation structuring, and a nonlinearity-enhanced encoder. SAEParate achieves state-of-the-art performance on UnlearnCanvas, with particularly strong gains in fine-grained joint unlearning, and demonstrates effective gender-distinct nudity unlearning on I2P. Further analyses show substantially reduced overlap among concept-relevant latent features, suggesting that concept-discriminative sparse representations provide an effective direction for reliable diffusion unlearning.

\textbf{Limitations and Future Directions.} Our method requires at least group-level supervision, inherits SAE-wide issues such as feature splitting and absorption that may incompletely capture target concepts, and is currently evaluated only on image generation. Our compatibility experiments with Matryoshka SAEs (Appendix~\ref{app:matryoshka_compatibility}) suggest that advances in SAE design could help alleviate limitations arising from feature splitting and absorption. Exploring such combinations, unsupervised training, and extension to video generation are promising directions for future work.

\subsubsection*{Acknowledgments}
This work was supported in part by the “Advanced GPU Utilization Support Program” funded by the Government of the Republic of Korea (Ministry of Science and ICT), the National Research Foundation of Korea (NRF) grants funded by the Korea government (MSIT) (RS-2026-25470983, RS-2026-25612350), the Institute of Information \& Communications Technology Planning \& Evaluation (IITP) grant funded by the Korea government (MSIT) (IITP-2026-RS-2026-25544188, Leading Generative AI Human Resources Development). Dong-Jun Han is the corresponding author.

\bibliography{references}
\bibliographystyle{plainnat}
\newpage

\appendix

\begin{center}
    {\bf\Large Appendix}
\end{center}

\startcontents[sections]
\printcontents[sections]{l}{1}{\setcounter{tocdepth}{3}}

\newpage

\section{Implementation Details}
\label{app:implementation_details}

\subsection{SAE Training Details}
\label{app:sae_training_details}

We follow the SAE training setup of SAeUron~\cite{cywinski2025saeuron} for our object and style blocks, training BatchTopK sparse autoencoders with $k=32$ and an expansion factor of $16$. For the joint block, we use a larger $k=64$ to accommodate the increased granularity of joint (style--object) features. Optimization uses Adam~\cite{kingma2014adam} with a learning rate of $4\times 10^{-4}$ and a linear scheduler without warmup. We use an effective batch size of $131{,}072$ and unit-normalize decoder weights after each training step. We set $k_{\text{aux}}$ to a power of two close to $n/2$ and $\alpha = 1/32$. A latent is considered dead if it has not activated over the last 10M training samples.

We jointly optimize the standard reconstruction objective with a latent-space contrastive loss from scratch. The contrastive loss is applied to unit-normalized SAE latents and uses temperature $\tau = 0.07$ across all settings. The contrastive loss weight $\lambda$ is set to $0.1$ for the object SAE, $0.01$ for the style SAE, and $1.0$ for the joint SAE. For the joint SAE, we additionally apply a hard negative weighting factor of $4$ to up-weight semantically close negatives that share either the object or the style with the anchor.

We train SAEs for the object, style, and joint settings. For object and style unlearning, we follow the empirically selected cross-attention blocks identified by SAeUron, using \texttt{up.1.1} for object unlearning and \texttt{up.1.2} for style unlearning. For joint unlearning, we apply SAEParate to the \texttt{up.1.1} block. Specifically, we train the object SAE for 40 epochs, the style SAE for 20 epochs, and the joint SAE for 40 epochs, using Stable Diffusion v1.5~\cite{rombach2022high} as the base model. Table~\ref{tab:sae_hparams} summarizes the key training hyperparameters and metrics. Training takes approximately 46, 20, and 69 hours on a single NVIDIA B200 GPU for the object, style, and joint SAEs, respectively.

\begin{table}[h]
\centering
\caption{Summary of SAE training for SAEParate.}
\label{tab:sae_hparams}
\resizebox{\textwidth}{!}{%
\begin{tabular}{lcccccccccc}
\toprule
\textbf{Block} & \textbf{\# Latents $n$} & $k$ & $\alpha$ & \textbf{$\lambda$} & $\tau$ & \textbf{Hard Neg.} & \textbf{LR} & \textbf{Batch Size} & \textbf{Epochs} & \textbf{GPU-hours} \\
\midrule
\texttt{up.1.1} (object) & 20480 & 32 & $\frac{1}{32}$ & 0.1  & 0.07 & --   & 0.0004 & 131072 & 40 & 46h 21min \\
\texttt{up.1.2} (style)  & 20480 & 32 & $\frac{1}{32}$ & 0.01 & 0.07 & --   & 0.0004 & 131072 & 20 & 20h 18min \\
\texttt{up.1.1} (joint)  & 20480 & 64 & $\frac{1}{32}$ & 1.0  & 0.07 & 4    & 0.0004 & 131072 & 40 & 68h 35min \\
\bottomrule
\end{tabular}%
}
\end{table}

\paragraph{SAEmnesia Adaptation for Joint Unlearning.}
\label{app:saemnesia_details}
Because SAEmnesia~\cite{cassano2026saemnesia} does not originally support joint style--object unlearning, we treat each style--object pair as a single atomic concept and assign it to a distinct latent, preserving its one-to-one concept--latent design. The model is trained from scratch on the same \texttt{up.1.1} activation data as our joint SAE, with all other training settings following the official implementation except that the object--style decorrelation term is disabled. At inference, the assigned latent is suppressed and $\gamma_c$ is tuned per target pair using the same validation and evaluation protocol as SAEParate.

\subsection{Subset Training Setting}
\label{app:subset_setting}

For ablations and additional analyses, we train SAEs on one-eighth of the
full UnlearnCanvas training prompt set, keeping all other settings identical
to Table~\ref{tab:sae_hparams}. Within each experiment, all compared variants,
including Vanilla SAE (SAeUron), are trained under this same setting, so
comparisons remain controlled. However, absolute numbers are not directly
comparable to the main results in Table~\ref{tab:main_results_grouped}; for
example, Vanilla SAE attains lower UA on object unlearning in this setting
(58.14\%) than its reported full-data result (78.82\%). This setting is used
for the component ablation (Table~\ref{tab:saeparate_ablation}), the
cross-attention weighted pooling ablation (Table~\ref{tab:cawp_ablation}),
the reconstruction--separation trade-off
(Table~\ref{tab:reconstruction_tradeoff}), compatibility with Matryoshka SAE
(Table~\ref{tab:matryoshka_compatibility}), and prompt variations in joint
unlearning (Table~\ref{tab:prompt_variations}).
\clearpage
\subsection{Sensitivity Analysis on the Contrastive Loss Weight}
\label{app:lambda_ablation}

We analyze the sensitivity of SAEParate to the contrastive loss weight
$\lambda$ in the overall training objective,
\begin{equation}
  \mathcal{L}_{\mathrm{total}}
  =
  \mathcal{L}_{\mathrm{SAE}}
  +
  \lambda \mathcal{L}_{\mathrm{LSC}}.
\end{equation}
The hyperparameter $\lambda$ controls the trade-off between preserving
the original diffusion activations through sparse reconstruction and
encouraging concept-wise separation in the SAE latent space.

We conduct the ablation in the object-unlearning setting by varying
$\lambda \in \{0.1, 0.2, 0.3, 0.4, 0.5\}$ while keeping all other
training and inference hyperparameters fixed. We report Unlearning
Accuracy (UA), In-domain Retain Accuracy (IRA), Cross-domain Retain
Accuracy (CRA), and their arithmetic mean.

\begin{table}[h]
  \centering
  \caption{
  Sensitivity analysis of the contrastive loss weight $\lambda$ on
  object unlearning. All other training and inference hyperparameters
  are held fixed. The best result in each column is highlighted in
  bold.
  }
  \label{tab:lambda_ablation}
  \small
  \setlength{\tabcolsep}{6pt}
  \renewcommand{\arraystretch}{1.05}
  \begin{tabular}{c|ccc|c}
      \toprule
      $\lambda$
      & UA ($\uparrow$)
      & IRA ($\uparrow$)
      & CRA ($\uparrow$)
      & Avg. ($\uparrow$) \\
      \midrule
      0.1
      & \textbf{95.20\%}
      & 95.91\%
      & 94.26\%
      & 95.12\% \\
      0.2
      & 94.61\%
      & \textbf{97.09\%}
      & 94.79\%
      & \textbf{95.50\%} \\

      0.3
      & 93.33\%
      & 96.47\%
      & 94.53\%
      & 94.78\% \\

      0.4
      & 92.65\%
      & 96.39\%
      & \textbf{95.12\%}
      & 94.72\% \\

      0.5
      & 94.61\%
      & 96.97\%
      & 94.37\%
      & 95.31\% \\
      \bottomrule
  \end{tabular}
\end{table}

As shown in Table~\ref{tab:lambda_ablation}, SAEParate remains
relatively stable across the evaluated range of $\lambda$, with the
average performance varying by approximately one percentage point.
Nevertheless, the effect of increasing $\lambda$ is not monotonic.
While a stronger contrastive objective can encourage concept-wise
separation, placing excessive emphasis on it does not consistently
improve downstream unlearning performance.

\subsection{Encoder Nonlinearity Ablation}
\label{app:encoder_architecture_ablation}

Section~\ref{sec:nonlinear_encoder} introduces a two-layer nonlinear encoder to provide additional flexibility for reshaping overlapping concept representations before sparsification. In particular, we adopt GeLU~\cite{hendrycks2016gaussian}, whose smooth and mildly nonmonotonic response differs from the coordinate-wise monotonic ReLU. We hypothesize that this property enables more flexible local transformations under the concept-separation objective. Relatedly, \citet{ren2026lowdimensional} show that, under certain assumptions, nonmonotonic activations can induce local folding of representations that is unavailable to coordinate-wise monotone activations. We regard this as a possible geometric explanation rather than a formal characterization of our encoder.

To examine the effect of this design, we compare otherwise identical two-layer encoders using Linear, ReLU, or GeLU as the intermediate activation $\sigma$ in Eq.~\ref{eq:gelu_encoder}. The SAE width, hidden dimensionality, sparsity level, training objective, and unlearning hyperparameters are fixed across all variants. The Linear variant contains no intermediate nonlinearity and is therefore expressively equivalent to a single affine transformation before the final ReLU, allowing us to isolate the contribution of nonlinear transformation.

\begin{table}[h]
  \centering
  \caption{
  Ablation of the intermediate encoder activation on object unlearning.
  All variants use the same two-layer architecture and differ only in
  the intermediate activation. The best result in each column is bolded.
  }
  \label{tab:encoder_architecture_ablation}
  \small
  \setlength{\tabcolsep}{6pt}
  \renewcommand{\arraystretch}{1.05}
  \begin{tabular}{l|ccc|c}
      \toprule
      Activation
      & UA ($\uparrow$)
      & IRA ($\uparrow$)
      & CRA ($\uparrow$)
      & Avg. ($\uparrow$) \\
      \midrule
      Linear
      & 74.02\%
      & \textbf{97.34\%}
      & \textbf{96.49\%}
      & 89.28\% \\

      ReLU
      & 90.88\%
      & 96.77\%
      & 95.31\%
      & 94.32\% \\

      GeLU
      & \textbf{95.20\%}
      & 95.91\%
      & 94.26\%
      & \textbf{95.12\%} \\
      \bottomrule
  \end{tabular}
\end{table}

Introducing an intermediate nonlinearity yields most of the improvement: ReLU increases UA from 74.02\% to 90.88\% and the average from 89.28\% to 94.32\%. GeLU further raises UA to 95.20\% and achieves the best average of 95.12\%, while maintaining retention performance comparable to ReLU. These results suggest that under the LSC objective, introducing the nonlinear hidden transformation accounts for most of the improvement over the linear encoder, with GeLU providing a further benefit for target-concept removal.

\subsection{Effect of SAE Layer Selection}
\label{app:layer_selection}

SAeUron~\cite{cywinski2025saeuron} uses \texttt{up.1.1} for object unlearning and \texttt{up.1.2} for style unlearning, and we follow the same layer choices in the corresponding settings for a fair comparison. For joint style--object unlearning, which requires representing both types of concepts within a single SAE, we use \texttt{up.1.1} as the default configuration in our main experiments. To examine the sensitivity of SAEParate to this layer choice, we additionally evaluate both SAEParate and SAeUron using \texttt{up.1.2}. Table~\ref{tab:layer_selection_joint} compares the two candidate layers under the same joint unlearning protocol.

\begin{table}[h]
\centering
\caption{Effect of SAE insertion layer on joint style--object unlearning. Results are reported for both \texttt{up.1.1} and \texttt{up.1.2}.}
\label{tab:layer_selection_joint}
\small
\setlength{\tabcolsep}{4.5pt}
\begin{tabular}{lcccccc}
\toprule
Method & Layer & UA ($\uparrow$) & SC ($\uparrow$) & OC ($\uparrow$) & UP ($\uparrow$) & Avg. ($\uparrow$) \\
\midrule
Vanilla SAE (SAeUron) & \texttt{up.1.1} & 63.00\% & 16.98\% & 75.84\% & 71.90\% & 56.93\% \\
SAEParate & \texttt{up.1.1} & \textbf{94.11\%} & \textbf{93.49\%} & \textbf{91.66\%} & \textbf{98.82\%} & \textbf{94.52\%} \\
\midrule
Vanilla SAE (SAeUron) & \texttt{up.1.2} & 93.90\% & 0.89\% & 55.73\% & 54.63\% & 51.29\% \\
SAEParate & \texttt{up.1.2} & \textbf{94.00\%} & \textbf{93.88\%} & \textbf{91.60\%} & \textbf{96.90\%} & \textbf{94.10\%} \\
\bottomrule
\end{tabular}
\end{table}

SAEParate consistently maintains strong joint unlearning performance across the two candidate layers, indicating that its effectiveness is not tied to a specific SAE insertion point. The comparison also suggests that the improvement over the reconstruction-oriented SAE baseline persists across layer choices, supporting the robustness of the proposed representation-learning approach.

\subsection{Computational Efficiency and Storage Overheads}
\label{app:training_memory_and_storage}

We also conduct an efficiency comparison with existing methods, including SAeUron. For a fair comparison, we follow the SAE training batch size of 4096 reported in Appendix~E of SAeUron~\cite{cywinski2025saeuron}. As shown in Table~\ref{tab:efficiency_comparison}, since SAEParate introduces supervised contrastive regularization only as an additional training objective, it does not substantially increase the training memory budget under the same batch size. Moreover, although SAEParate employs a lightweight enhanced SAE encoder, the resulting SAE checkpoint size remains nearly unchanged compared to SAeUron.

\begin{table}[h]
\centering
\small
\caption{Efficiency comparison with existing diffusion unlearning methods. Memory is measured under the same SAE training batch size of 4096 following SAeUron Appendix~E. Storage denotes additional method-specific files required for unlearning or inference.}
\begin{tabular}{lcc}
\toprule
\multirow{2}{*}{Method}
& \multicolumn{2}{c}{Efficiency} \\
\cmidrule(lr){2-3}
& Memory (GB) ($\downarrow$)
& Storage (GB) ($\downarrow$) \\
\midrule
ESD~\cite{gandikota2023erasing}     & 17.8 & 4.3 \\
FMN~\cite{zhang2024forget}          & 17.9 & 4.2 \\
UCE~\cite{gandikota2024unified}     & 5.1  & 1.7 \\
CA~\cite{kumari2023ablating}        & 10.1 & 4.2 \\
SalUn~\cite{fan2024salun}           & 30.8 & 4.0 \\
SEOT~\cite{li2024get}               & 7.3 & 0.0 \\
SPM~\cite{lyu2024one}               & 6.9  & 0.0 \\
EDiff~\cite{Wu_2025_CVPR}           & 27.8 & 4.0 \\
SHS~\cite{wu2024scissorhands}       & 31.2 & 4.0 \\
SAeUron~\cite{cywinski2025saeuron}  & 2.8  & 0.2 \\
\textbf{SAEParate}                  & 3.4  & 0.2 \\
\bottomrule
\end{tabular}
\label{tab:efficiency_comparison}
\end{table}

\paragraph{Inference-Time Overhead.}
We additionally measure end-to-end image-generation wall-clock time on a
single NVIDIA RTX 4090 under identical generation settings. As shown in
Table~\ref{tab:inference_efficiency}, both SAE-based methods introduce only
a small inference overhead relative to the original Stable Diffusion
backbone.

\begin{table}[h]
\centering
\small
\caption{
Inference-time efficiency measured on a single NVIDIA RTX 4090.
Relative inference time is normalized by the Stable Diffusion backbone.
}
\label{tab:inference_efficiency}
\begin{tabular}{lcc}
\toprule
Variant
& Relative Inference Time
& Inference Time (s) \\
\midrule
SD Backbone       & 1.00$\times$ & 2.9252 \\
Vanilla SAE (SAeUron)            & 1.02$\times$ & 2.9776 \\
\textbf{SAEParate} & 1.02$\times$ & 2.9907 \\
\bottomrule
\end{tabular}
\end{table}

SAEParate processes the spatial activations through a single batched SAE
forward pass at one U-Net hook point per denoising step, resulting in
approximately 2\% inference-time overhead relative to the backbone.
The additional storage requirement is also modest: we train one SAE per
unlearning setting, and each FP32 SAE checkpoint occupies approximately
200\,MB.

\clearpage

\subsection{Unlearning Procedure Details}
\label{app:saeuron_alg_details}
We perform concept unlearning by suppressing a small set of concept-specific SAE features at inference time. Both the feature importance score function and the suppression mechanism are taken directly from Sections 4.1 and 4.2 of SAeUron~\cite{cywinski2025saeuron} without modification; we describe them here for completeness and specify the implementation used in our experiments. 

\subsubsection{Feature Importance Scoring}
\label{app:feature_importance_scoring}
Following SAeUron~\cite{cywinski2025saeuron}, we identify SAE features that strongly and exclusively correspond to a target concept $c$ at each denoising timestep $t$. Given a dataset of activations $\mathcal{D} = \mathcal{D}_c \cup \mathcal{D}_{\neg c}$ partitioned by concept presence, the importance of the $i$-th feature for concept $c$ at timestep $t$ is defined as
\begin{equation}
\text{score}(i, t, c, \mathcal{D}) = \frac{\mu(i, t, \mathcal{D}_c)}{\sum_{j=1}^{n} \mu(j, t, \mathcal{D}_c) + \delta} - \frac{\mu(i, t, \mathcal{D}_{\neg c})}{\sum_{j=1}^{n} \mu(j, t, \mathcal{D}_{\neg c}) + \delta},
\label{eq:saeuron_score}
\end{equation}
where $\mu(i, t, \mathcal{D}) = \frac{1}{|\mathcal{D}|} \sum_{x \in \mathcal{D}} f_i(x_t)$ is the average activation of feature $i$ at timestep $t$, $n$ is the SAE width, and $\delta$ is a small constant for numerical stability. Normalization by the per-subset feature mass prevents broadly active features from dominating the score, so high-scoring features are those activating strongly on $\mathcal{D}_c$ while remaining weak on $\mathcal{D}_{\neg c}$.

\subsubsection{Inference-Time Suppression}
\label{app:saeuron_selection_suppression}
For each concept $c$, we form the suppression set $\mathcal{F}_c$ by selecting the top-$\tau_c$ features by score, after filtering out dead features and those whose activation frequency exceeds the 99th percentile of the feature density distribution. For features in $\mathcal{F}_c$, the activation is modified as
\begin{equation}
\tilde{f}_i(x) = 
\begin{cases} 
\gamma_c \, \mu(i, t, \mathcal{D}_c) \, f_i(x), & \text{if } i \in \mathcal{F}_c \text{ and } f_i(x) > \mu(i, t, \mathcal{D}), \\
f_i(x), & \text{otherwise,}
\end{cases}
\end{equation}
where the condition $f_i(x) > \mu(i, t, \mathcal{D})$ prevents random ablation when activations are uniformly low. The modified activations are decoded back through the SAE decoder while preserving the reconstruction error. The percentile is selected via grid search over the score distribution; the chosen values for each experimental setup are reported in Appendix~\ref{app:hyperparams}.

\subsection{Joint Unlearning Formulation}
\label{app:joint_unlearning_formulation}
To enable precise unlearning in the style--object combination setup, we extend the SupCon objective in Eq.~\ref{eq:supcon} along three axes that exploit the compositional structure of the target: (i) the positive set $\mathcal{P}(i)$ is restricted to samples matching on \emph{both} attributes, (ii) the negatives in the partition function of Eq.~\ref{eq:supcon} are reweighted to emphasize partial-match neighbors, and (iii) the cross-attention weights $\alpha_{i,p}$ in Eq.~\ref{eq:cross_attention_weight} are computed from prompts specifying the joint concept. Throughout this section we reuse the per-sample representation $u_i = r_i/\|r_i\|_2$ defined in Section~\ref{sec:supcon}. 

\paragraph{Joint positive set.}
Each target combination consists of two attributes, an object and a style. Let $y_i^{cls}$ and $y_i^{thm}$ denote the object and style labels of sample $i$, and let $\mathcal{V}$ be the set of target (object, style) combinations. We replace the single-label positive set $\mathcal{P}(i) = \{j \neq i \mid y_j = y_i\}$ from Eq.~\ref{eq:supcon} with one that requires agreement on both attributes within $\mathcal{V}$:
\begin{equation}
    \mathcal{P}^+(i) = \{j \neq i \mid y_j^{cls} = y_i^{cls},\ y_j^{thm} = y_i^{thm},\ (y_i^{cls}, y_i^{thm}) \in \mathcal{V}\}.
\end{equation}
Reducing to a single-attribute (i.e., $y_i \equiv y_i^{cls}$ or $y_i \equiv y_i^{thm}$) recovers Eq.~\ref{eq:supcon} exactly.

\paragraph{Hard negative reweighting.}
A central difficulty specific to the joint setting is that samples sharing exactly one attribute sit in an ambiguous region of representation space. For instance, if the unlearning target is \textit{Dogs in Cartoon style}, then \textit{Flowers in Cartoon style} (matching only the style) and \textit{Dogs in Ink Art style} (matching only the object) are partial-match samples that remain close to the anchor under $u_i^\top u_j$ yet belong to distinct concepts. The standard SupCon partition function in Eq.~\ref{eq:supcon} treats these as ordinary negatives, providing an insufficiently strong separation signal. We therefore define the hard negative set
\begin{equation}
    \mathcal{H}(i) = \{j \neq i \mid (y_j^{cls} = y_i^{cls}) \oplus (y_j^{thm} = y_i^{thm})\},
\end{equation}
where $\oplus$ denotes exclusive-or, i.e., the condition holds iff exactly one of the two equalities is satisfied. Rather than altering the cosine similarity itself, we modify the per-pair logit $u_i^\top u_j / \tau$ used inside Eq.~\ref{eq:supcon} by an additive bias:
\begin{equation}
    \tilde{s}_{ij}
    =
    \frac{u_i^\top u_j}{\tau}
    + \log w_{ij},
    \qquad
    w_{ij}
    =
    \begin{cases}
        \omega_{\mathrm{HN}} & \text{if } j \in \mathcal{H}(i), \\
        1 & \text{otherwise.}
    \end{cases}
\end{equation}
Inflating the logit of a hard negative by $\log \omega_{\mathrm{HN}}$ increases its mass in the partition function, yielding a stronger push away from partial-match neighbors while leaving non-hard negatives at their original SupCon weighting. The joint SupCon loss is then a direct generalization of Eq.~\ref{eq:supcon}:
\begin{equation}
    \mathcal{L}_{\text{Joint}}
    =
    -\frac{1}{|\mathcal{I}^+|}
    \sum_{i \in \mathcal{I}^+}
    \frac{1}{|\mathcal{P}^+(i)|}
    \sum_{j \in \mathcal{P}^+(i)}
    \left(
        \tilde{s}_{ij}
        -
        \log \sum_{k \neq i} \exp(\tilde{s}_{ik})
    \right),
\end{equation}
where $\mathcal{I}^+ = \{i \in \mathcal{I} \mid |\mathcal{P}^+(i)| > 0\}$. Setting $\omega_{\mathrm{HN}} = 1$ and replacing $\mathcal{P}^+$ with $\mathcal{P}$ recovers Eq.~\ref{eq:supcon}, making explicit that $\mathcal{L}_{\text{Joint}}$ adds two compositional ingredients on top of the base objective.

\paragraph{Joint cross-attention weighting and target-biased sampling.}
The spatial weighting in Eq.~\ref{eq:cross_attention_weight} is concept-conditioned through the prompt token set $\mathcal{C}_i$. For joint unlearning we set $\mathcal{C}_i$ to include both the target object and style tokens, so that $\alpha_{i,p}$ reflects spatial relevance to both attributes of the joint concept; the aggregated representation
\begin{equation}
    r_i = \sum_{p=1}^{S} \alpha_{i,p}\, z_{i,p}
\end{equation}
from Eq.~\ref{eq:supcon_r} is otherwise unchanged. Finally, to ensure that target combinations receive sufficient gradient signal, we enforce that at least one fifth of each mini-batch consists of samples drawn from $\mathcal{V}$. Together with $\mathcal{P}^+$ and $\mathcal{H}$, this prevents $\mathcal{L}_{\text{Joint}}$ from being dominated by non-target pairs and provides consistent exposure to the fine-grained distinctions the SAE must learn, encouraging tight, well-separated clusters for each (object, style) combination.

\clearpage

\subsection{Unlearning Hyperparameters}
\label{app:hyperparams}

For each concept $c$, $\tau_c$ denotes the number of selected features used to suppress the target concept, and $\gamma_c$ denotes the negative multiplier controlling the suppression strength. Following SAeUron, which selects these hyperparameters through validation-based search, we tune $\tau_c$ and $\gamma_c$ on validation prompts for object and joint unlearning, while fixing $\tau_c=1$ and $\gamma_c=-0.1$ for style unlearning. This yields one hyperparameter pair per target concept, resulting in 20 pairs for object unlearning and 50 pairs for joint unlearning. The selected hyperparameters are reported in Tables~\ref{tab:hyperparams_object} and~\ref{tab:hyperparams_joint}, respectively.

\begin{table}[h]
\centering
\caption{Hyperparameters of our method for object unlearning.}
\label{tab:hyperparams_object}
\begin{tabular}{lcc}
\toprule
\textbf{Object} & \textbf{Selected features $\tau_c$} & \textbf{Multiplier $\gamma_c$} \\
\midrule
Architectures & 3 & $-0.1$ \\
Bears         & 2 & $-0.1$ \\
Birds         & 3 & $-0.25$ \\
Butterfly     & 2 & $-0.1$ \\
Cats          & 2 & $-0.1$ \\
Dogs          & 2 & $-0.1$ \\
Fishes        & 1 & $-0.1$ \\
Flame         & 3 & $-5.0$ \\
Flowers       & 3 & $-1.0$ \\
Frogs         & 2 & $-0.1$ \\
Horses        & 3 & $-0.75$ \\
Human         & 4 & $-0.25$ \\
Jellyfish     & 1 & $-10.0$ \\
Rabbits       & 2 & $-1.0$ \\
Sandwiches    & 3 & $-0.5$ \\
Sea           & 3 & $-1.0$ \\
Statues       & 1 & $-0.1$ \\
Towers        & 2 & $-0.1$ \\
Trees         & 3 & $-0.25$ \\
Waterfalls    & 3 & $-0.25$ \\
\bottomrule
\end{tabular}
\end{table}

\begin{table}[h]
\centering
\small
\caption{Hyperparameters of our method for joint (object, style) unlearning.}
\label{tab:hyperparams_joint}
\begin{tabular}{llcc@{\hspace{1em}}llcc}
\toprule
\textbf{Object} & \textbf{Style} & $\tau_{c}$ & $\gamma_{c}$ & \textbf{Object} & \textbf{Style} & $\tau_{c}$ & $\gamma_{c}$ \\
\midrule
Architectures & Abstractionism   & 2  & $-$1.0  & Dogs          & Magic\_Cube     & 2  & $-$9.0  \\
Bears         & Artist\_Sketch   & 3  & $-$9.0  & Fishes        & Meta\_Physics   & 2  & $-$7.0  \\
Birds         & Blossom\_Season  & 4  & $-$11.0 & Flame         & Meteor\_Shower  & 2  & $-$5.0  \\
Butterfly     & Bricks           & 4  & $-$7.0  & Flowers       & Monet           & 4  & $-$11.0 \\
Cats          & Byzantine        & 3  & $-$1.0  & Frogs         & Mosaic          & 3  & $-$9.0  \\
Dogs          & Cartoon          & 2  & $-$3.0  & Horses        & Neon\_Lines     & 3  & $-$3.0  \\
Fishes        & Cold\_Warm       & 3  & $-$11.0 & Human         & On\_Fire        & 4  & $-$7.0  \\
Flame         & Color\_Fantasy   & 4  & $-$9.0  & Jellyfish     & Pastel          & 2  & $-$5.0  \\
Flowers       & Comic\_Etch      & 2  & $-$1.0  & Rabbits       & Pencil\_Drawing & 5  & $-$5.0  \\
Frogs         & Crayon           & 13 & $-$11.0 & Sandwiches    & Picasso         & 2  & $-$5.0  \\
Horses        & Cubism           & 2  & $-$7.0  & Sea           & Pop\_Art        & 3  & $-$11.0 \\
Human         & Dadaism          & 3  & $-$5.0  & Statues       & Red\_Blue\_Ink  & 3  & $-$9.0  \\
Jellyfish     & Dapple           & 2  & $-$5.0  & Towers        & Rust            & 2  & $-$1.0  \\
Rabbits       & Defoliation      & 11 & $-$11.0 & Waterfalls    & Sketch          & 2  & $-$5.0  \\
Sandwiches    & Early\_Autumn    & 2  & $-$11.0 & Architectures & Sponge\_Dabbed  & 3  & $-$9.0  \\
Sea           & Expressionism    & 3  & $-$9.0  & Bears         & Structuralism   & 3  & $-$9.0  \\
Statues       & Fauvism          & 2  & $-$7.0  & Birds         & Superstring     & 4  & $-$11.0 \\
Towers        & French           & 2  & $-$1.0  & Butterfly     & Surrealism      & 3  & $-$11.0 \\
Trees         & Glowing\_Sunset  & 3  & $-$9.0  & Cats          & Ukiyoe          & 12 & $-$7.0  \\
Waterfalls    & Gorgeous\_Love   & 2  & $-$7.0  & Dogs          & Van\_Gogh       & 3  & $-$1.0  \\
Architectures & Greenfield       & 2  & $-$9.0  & Fishes        & Vibrant\_Flow   & 3  & $-$7.0  \\
Bears         & Impressionism    & 4  & $-$5.0  & Flame         & Warm\_Love      & 3  & $-$7.0  \\
Birds         & Ink\_Art         & 4  & $-$5.0  & Flowers       & Warm\_Smear     & 4  & $-$5.0  \\
Butterfly     & Joy              & 2  & $-$11.0 & Frogs         & Watercolor      & 2  & $-$11.0 \\
Cats          & Liquid\_Dreams   & 2  & $-$9.0  & Horses        & Winter          & 2  & $-$3.0  \\
\bottomrule
\end{tabular}
\end{table}

\clearpage

\subsubsection{Sensitivity to Suppression Hyperparameters}
\label{app:suppression_sensitivity}

We further analyze the sensitivity of SAEParate to the two hyperparameters governing inference-time suppression: the multiplier $\gamma_c$, which controls the strength of suppression, and the percentile used to determine $\tau_c$. We vary $\gamma_c \in \{-1,-3,-5,-7,-9,-11\}$ and the percentile for $\tau_c \in \{99.95, 99.99, 99.995\}$ while keeping all other training and inference settings fixed.

We conduct this analysis in the joint style--object unlearning setting and report results averaged across the evaluated target concepts. We measure Unlearning Accuracy (UA), Style Consistency (SC), Object Consistency (OC), and their arithmetic mean (Avg.).

\begin{table}[ht]
\centering
\caption{
Sensitivity to inference-time suppression hyperparameters in joint unlearning.
The percentile is used to determine $\tau_c$. Results are averaged across
the evaluated joint concepts.
}
\label{tab:suppression_sensitivity}
\small
\setlength{\tabcolsep}{5pt}
\renewcommand{\arraystretch}{1.05}
\begin{tabular}{lc|cccc}
\toprule
\textbf{$\gamma_c$}
& \textbf{Percentile for $\tau_c$}
& \textbf{UA ($\uparrow$)}
& \textbf{SC ($\uparrow$)}
& \textbf{OC ($\uparrow$)}
& \textbf{Avg. ($\uparrow$)} \\
\midrule

\multirow{3}{*}{$-1$}
& 99.95  & 56.4 & 87.9 & 88.0 & 77.4 \\
& 99.99  & 57.0 & 88.0 & 88.0 & 77.7 \\
& 99.995 & 61.4 & 88.0 & 88.0 & 79.1 \\

\midrule
\multirow{3}{*}{$-3$}
& 99.95  & 81.0 & 86.0 & 86.4 & 84.5 \\
& 99.99  & 86.8 & 86.2 & 86.4 & 86.5 \\
& 99.995 & 86.6 & 86.4 & 86.4 & 86.5 \\

\midrule
\multirow{3}{*}{$-5$}
& 99.95  & 87.6 & 84.9 & 85.3 & 85.9 \\
& 99.99  & 91.6 & 85.1 & 85.3 & 87.3 \\
& 99.995 & 92.8 & 85.3 & 85.3 & 87.8 \\

\midrule
\multirow{3}{*}{$-7$}
& 99.95  & 92.4 & 84.0 & 84.6 & 87.0 \\
& 99.99  & 94.2 & 84.5 & 84.6 & 87.8 \\
& 99.995 & 94.4 & 84.6 & 84.6 & 87.9 \\

\midrule
\multirow{3}{*}{$-9$}
& 99.95  & 94.2 & 83.1 & 84.0 & 87.1 \\
& 99.99  & 94.4 & 83.6 & 84.0 & 87.3 \\
& 99.995 & 95.6 & 84.0 & 84.0 & 87.9 \\

\midrule
\multirow{3}{*}{$-11$}
& 99.95  & 96.2 & 82.5 & 83.5 & 87.4 \\
& 99.99  & 96.4 & 83.0 & 83.5 & 87.6 \\
& 99.995 & 96.8 & 83.5 & 83.5 & 87.9 \\

\bottomrule
\end{tabular}
\end{table}

We perform the sensitivity analysis under the joint unlearning setting.
As shown in Table~\ref{tab:suppression_sensitivity}, increasing the
magnitude of $\gamma_c$ consistently improves UA while gradually reducing
retention. Consequently, the overall average remains stable at approximately
87--88\% across a broad range of moderate-to-strong suppression settings.
The percentile used to determine $\tau_c$ has a comparatively smaller
effect, indicating that SAEParate does not rely on a narrowly tuned
suppression configuration.

\clearpage

\subsection{Prompt Details}
\label{app:prompts_details}

For training the SAE in concept unlearning tasks, we collect internal diffusion activations across 50 denoising steps using the prompts provided in UnlearnCanvas~\cite{zhang2024unlearncanvas}. Specifically, we use 80 simple one-sentence \textit{anchor prompts} for each object class, originally employed in CA~\cite{kumari2023ablating}. To train style-related representations, we append the postfix \texttt{``in \{style\} style.''} to each anchor prompt. UnlearnCanvas provides 20 object classes and 50 style themes.

The object classes are:
\begin{itemize}
    \item \textit{Architectures, Bears, Birds, Butterfly, Cats, Dogs, Fishes, Flame,
    Flowers, Frogs, Horses, Human, Jellyfish, Rabbits, Sandwiches,
    Sea, Statues, Towers, Trees, Waterfalls}.
\end{itemize}

The style themes are:
\begin{itemize}
    \item \textit{Abstractionism, Artist Sketch, Blossom Season, Bricks, Byzantine, Cartoon,
    Cold Warm, Color Fantasy, Comic Etch, Crayon, Cubism, Dadaism, Dapple,
    Defoliation, Early Autumn, Expressionism, Fauvism, French, Glowing Sunset,
    Gorgeous Love, Greenfield, Impressionism, Ink Art, Joy, Liquid Dreams,
    Magic Cube, Meta Physics, Meteor Shower, Monet, Mosaic, Neon Lines,
    On Fire, Pastel, Pencil Drawing, Picasso, Pop Art, Red Blue Ink, Rust,
    Sketch, Sponge Dabbed, Structuralism, Superstring, Surrealism, Ukiyoe,
    Van Gogh, Vibrant Flow, Warm Love, Warm Smear, Watercolor, Winter}.
\end{itemize}

The joint combinations follow the style--object Combination Unlearning setup in Appendix B.5 of UnlearnCanvas~\cite{zhang2024unlearncanvas}.
\begin{itemize}
    \item \textit{(Architectures, Abstractionism), (Bears, Artist Sketch), (Birds, Blossom Season), (Butterfly, Bricks), (Cats, Byzantine), (Dogs, Cartoon), (Fishes, Cold Warm), (Flame, Color Fantasy), (Flowers, Comic Etch), (Frogs, Crayon), (Horses, Cubism), (Human, Dadaism), (Jellyfish, Dapple), (Rabbits, Defoliation), (Sandwiches, Early Autumn), (Sea, Expressionism), (Statues, Fauvism), (Towers, French), (Trees, Glowing Sunset), (Waterfalls, Gorgeous Love), (Architectures, Greenfield), (Bears, Impressionism), (Birds, Ink Art), (Butterfly, Joy), (Cats, Liquid Dreams), (Dogs, Magic Cube), (Fishes, Meta Physics), (Flame, Meteor Shower), (Flowers, Monet), (Frogs, Mosaic), (Horses, Neon Lines), (Human, On Fire), (Jellyfish, Pastel), (Rabbits, Pencil Drawing), (Sandwiches, Picasso), (Sea, Pop Art), (Statues, Red Blue Ink), (Towers, Rust), (Waterfalls, Sketch), (Architectures, Sponge Dabbed), (Bears, Structuralism), (Birds, Superstring), (Butterfly, Surrealism), (Cats, Ukiyoe), (Dogs, Van Gogh), (Fishes, Vibrant Flow), (Flame, Warm Love), (Flowers, Warm Smear), (Frogs, Watercolor), (Horses, Winter)}.
\end{itemize}

For feature score calculation, which is used to identify concept-specific SAE features, we use a validation set based on the \textit{anchor prompts} same as SAeUron~\cite{cywinski2025saeuron}. For style unlearning, we use 20 selected \textit{anchor prompts}, each appended with the postfix \texttt{``in \{style\} style.''} for every style theme. The 20 prompts are:

\begin{enumerate}
    \item ``Gothic cathedral with flying buttresses and stained glass windows.''
    \item ``Bear dressed as a medieval knight in armor.''
    \item ``Bird with iridescent, oil-slick-like feathers.''
    \item ``Butterfly emerging from a jeweled cocoon.''
    \item ``Cat wearing a superhero cape, leaping between buildings.''
    \item ``Dog wearing aviator goggles, piloting an airplane.''
    \item ``Goldfish swimming in a crystal-clear bowl.''
    \item ``Candle flame flickering in an old mysterious library.''
    \item ``Flower blooming in a snow-covered landscape.''
    \item ``Frog whose croak sounds like a jazz trumpet.''
    \item ``Wild horse galloping across the prairie at sunrise.''
    \item ``Man hiking through a dense forest.''
    \item ``Jellyfish floating in deep blue water.''
    \item ``Rabbit peering out from a burrow.''
    \item ``Classic BLT sandwich on toasted bread.''
    \item ``Sea waves crashing over ancient coastal ruins.''
    \item ``Forgotten hero statue covered in ivy.''
    \item ``Tower soaring above the clouds.''
    \item ``Majestic oak tree in a serene forest.''
    \item ``Moonlit waterfall in a serene forest.''
\end{enumerate}

For object unlearning, we use the full set of 80 \textit{anchor prompts} for each object class without the style postfix. The joint unlearning setting uses the same 80 \textit{anchor prompts} for each (object, style) combination, with the corresponding style postfix appended. 

At evaluation time, we follow the UnlearnCanvas prompt format and generate style--object prompts using the template
\[
\texttt{``An image of \{object\} in \{style\} style.''}
\]
For example, \texttt{``An image of Architectures in Abstractionism style.''}.

\clearpage

\subsection{Cross-attention Weighted Pooling Details}
\label{app:cross_attention_weight}

\begin{figure}[h]
    \centering
    \includegraphics[width=1.0\linewidth]{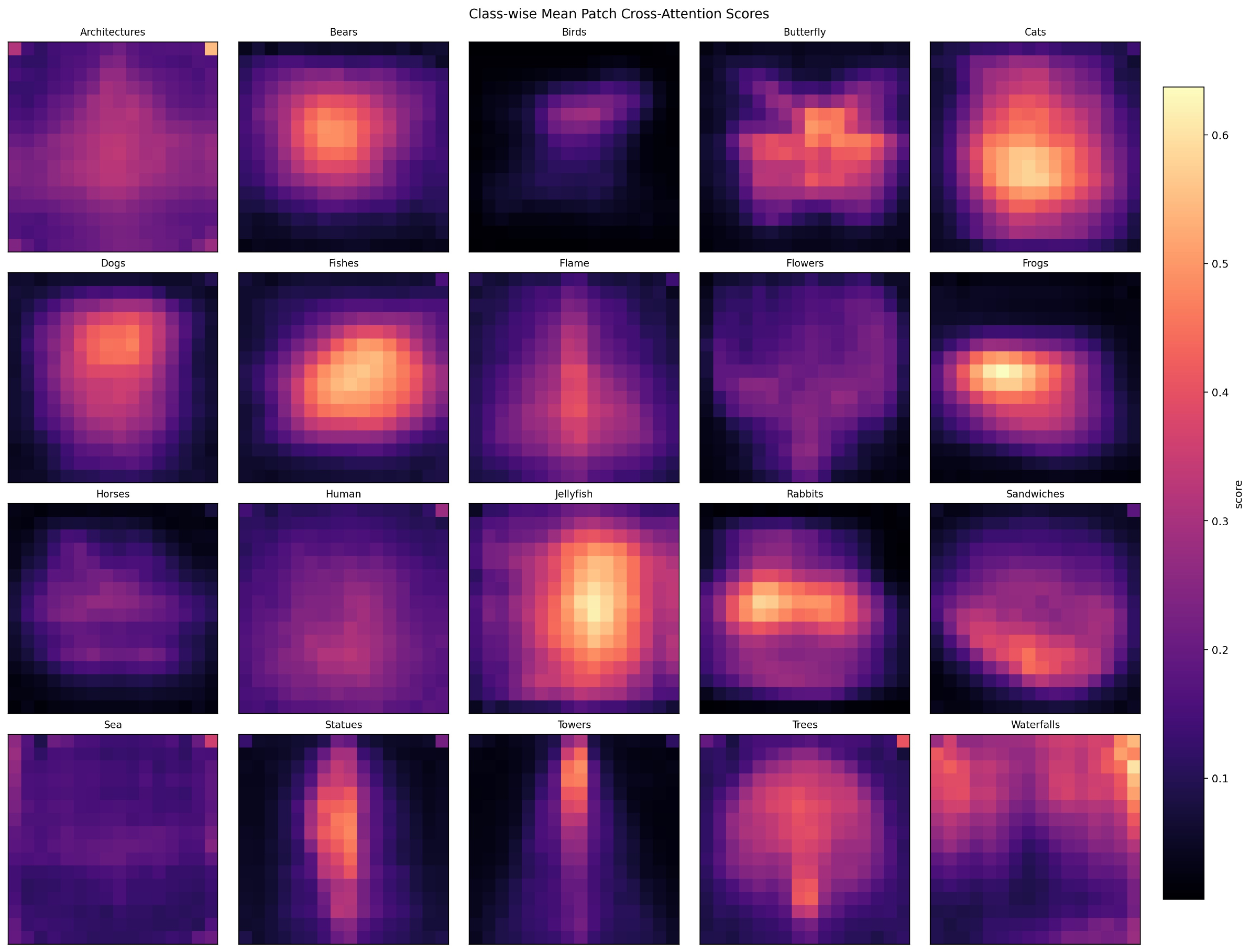}
    \caption{Cross-attention score visualization of object classes.}
    \label{fig:cross_attention_score_viz}
\end{figure}

We compute the cross-attention-guided weights in Eq.~\ref{eq:cross_attention_weight} by reusing the spatial grounding signal already present in the diffusion model's cross-attention maps. For each sample, we identify the prompt tokens corresponding to the target concept, denoted by $\mathcal{C}_i$. This set contains all tokenizer units matched to the concept phrase, since object names or style phrases may be split into multiple tokens.

We extract attention probabilities from the cross-attention layer in the same UNet block where the SAE is trained. We attach a forward hook to this layer and access the image hidden states and text encoder hidden states passed through it during the diffusion process. Since this layer performs text-conditioned cross-attention, image hidden states are used as queries, while text encoder hidden states provide the keys and values. The resulting attention probabilities therefore indicate how each spatial position in the image feature map attends to each prompt token.

Within the hook, we reproduce the layer's internal attention computation using its own projection weights to compute the query--key attention probabilities. Hence, the attention maps are extracted directly from the diffusion UNet's cross-attention operation at the SAE training block, rather than from an external localization or segmentation module.

Given the resulting attention probabilities $A_{i,h,p,k}$ over attention head $h$, spatial position $p$, and text token $k$, we compute a concept-level relevance score for each spatial position by averaging over heads and aggregating the attention mass assigned to tokens in $\mathcal{C}_i$. The scores are then normalized across spatial positions to obtain $\alpha_{i,p}$, where $\sum_p \alpha_{i,p}=1$. These weights are soft pooling coefficients rather than binary masks.

The weights are computed independently for each denoising timestep and used to construct the sample-level representation for supervised contrastive learning:
\begin{align}
r_i = \sum_{p=1}^{S} \alpha_{i,p} z_{i,p}.
\end{align}
Thus, concept-relevant spatial regions contribute more strongly to the contrastive representation, while background or concept-irrelevant regions are down-weighted. Figure~\ref{fig:cross_attention_score_viz} visualizes these stored spatial weights averaged over samples of the same concept.

\begin{table}[h]
    \centering
    \caption{Cross-attention weighted pooling ablation results on object unlearning.}
    \label{tab:cawp_ablation}
    \small
    \setlength{\tabcolsep}{6pt}
    \renewcommand{\arraystretch}{1.05}
    \begin{tabular}{l|ccc|c}
        \toprule
        \multirow{2}{*}{Variants}
        & \multicolumn{3}{c|}{Unlearning Performance}
        & \multirow{2}{*}{Avg. ($\uparrow$)} \\
        \cmidrule(lr){2-4}
        & UA ($\uparrow$)
        & IRA ($\uparrow$)
        & CRA ($\uparrow$)
        & \\
        \midrule
        \textit{w/o}-CAWP
        & 79.41\% & 97.28\% & 95.56\% & 90.75\% \\
        \textbf{SAEParate}
        & 95.59\% & 96.48\% & 93.46\% & 95.18\% \\
        \bottomrule
    \end{tabular}
\end{table}

To assess the contribution of cross-attention weighted pooling, we conduct an ablation study under the subset setting (Appendix~\ref{app:subset_setting}). The \textit{w/o}-CAWP variant constructs the representation for the contrastive loss by mean-pooling over all spatial positions, i.e., replacing $\alpha_{i,p}$ in Eq.~\ref{eq:supcon_r} with uniform weights $1/S$. We compare this variant with our full method, which uses cross-attention weighted pooling to emphasize concept-relevant spatial positions. As shown in Table~\ref{tab:cawp_ablation}, \textit{w/o}-CAWP exhibits a weaker trade-off between unlearning and retention, yielding lower unlearning accuracy.

These results indicate that treating all spatial positions equally can introduce concept-irrelevant evidence into the contrastive representation, such as background regions or co-occurring non-target objects. This weakens the concept-specific contrastive signal, leading to less discriminative feature separation and a weaker unlearning--retention trade-off.

\clearpage

\section{Further Experiments and Analyses}
\label{app:further_exp_and_ana}

\subsection{SAeUron with Matched Training Epochs}
\label{app:saeuron_matched_training_epoch}

For the main comparison, we report the published SAeUron results obtained under its original training configuration. To examine whether the performance gap could be attributed to differences in SAE training duration, we additionally retrain the official SAeUron implementation using the same number of epochs as SAEParate. As shown in Table~\ref{tab:matched_epoch}, extending SAeUron training to 40 epochs for object unlearning and 20 epochs for style unlearning yields comparable or slightly worse performance than its original 5- and 10-epoch schedules, respectively. This controlled comparison indicates that the gains of SAEParate are not simply due to longer SAE training.

\begin{table}[h]
\centering
\caption{Comparison with SAeUron under matched SAE training epochs. We retrain SAeUron with the same number of epochs used by our method (40 for object, 20 for style) and compare against its original training schedule (5 / 10 epochs) and our method. Best results in \textbf{bold}.}
\label{tab:matched_epoch}
\small
\setlength{\tabcolsep}{5pt} 
\resizebox{0.9\textwidth}{!}{
\begin{tabular}{llccccl}
\toprule
Setting & Method & UA ($\uparrow$) & IRA ($\uparrow$) & CRA ($\uparrow$) & Avg. ($\uparrow$) & \\
\midrule
\multirow{3}{*}{Object} 
  & SAeUron (5 epochs, original)   & 77.35\% & 95.68\% & \textbf{95.20\%} & 89.41\% & \\
  & SAeUron (40 epochs, matched)   & 75.98\% & 95.49\% & 94.45\% & 88.64\% & \\
  & \textbf{SAEParate (40 epochs)} & \textbf{95.20\%} & \textbf{95.91\%} & 94.26\% & \textbf{95.12\%} & \\
\midrule
\multirow{3}{*}{Style}  
  & SAeUron (10 epochs, original)  & 96.30\% & 98.86\% & 97.40\% & 97.52\% & \\
  & SAeUron (20 epochs, matched)   & 96.90\% & 98.76\% & 97.80\% & 97.82\% & \\
  & \textbf{SAEParate (20 epochs)} & \textbf{99.60\%} & \textbf{99.24\%} & \textbf{98.30\%} & \textbf{99.05\%} & \\
\bottomrule
\end{tabular}
}
\end{table}

\clearpage

\subsection{Reconstruction--Separation Trade-off}
\label{app:reconstruction_tradeoff}

Since SAEParate introduces an explicit concept-separation objective in
addition to sparse reconstruction, we examine whether the improved
concept discrimination comes at the cost of reconstruction fidelity.
We compare SAEParate with a Vanilla SAE (SAeUron), both trained under the subset setting (Appendix~\ref{app:subset_setting}) with a matched budget of 20K optimization steps.

We measure reconstruction quality using the fraction of variance
unexplained (FVU). For each microbatch $b$, we compute

\begin{equation}
\mathrm{FVU}_b =
\frac{
    \sum_{i \in b}
    \left\lVert x_i - \hat{x}_i \right\rVert_2^2
}{
    \sum_{i \in b}
    \left\lVert x_i - \bar{x}_b \right\rVert_2^2
},
\label{eq:fvu_microbatch}
\end{equation}

where $x_i$ and $\hat{x}_i$ denote the original and reconstructed
activations, respectively, and $\bar{x}_b$ is the mean activation
within microbatch $b$. We report the average of the microbatch-level
ratios over the corresponding logging interval:

\begin{equation}
\mathrm{FVU}
=
\frac{1}{B}
\sum_{b=1}^{B}
\mathrm{FVU}_b,
\label{eq:fvu_average}
\end{equation}

where $B$ denotes the number of microbatches in the interval. Lower
FVU indicates better reconstruction fidelity.

As shown in Table~\ref{tab:reconstruction_tradeoff}, SAEParate exhibits
a moderate increase in reconstruction error compared with the Vanilla
SAE, with FVU increasing from 0.172 to 0.217. This trade-off is expected
because SAEParate is not optimized solely for reconstruction: the
additional concept-aware contrastive objective explicitly reorganizes
the latent space to promote concept discrimination.

Importantly, the reduction in reconstruction fidelity is accompanied
by a substantial improvement in latent-space separation. SAEParate
increases the cluster rate from 0.93 to 1.00 and reduces the centroid
margin from 0.73 to 0.06, indicating markedly clearer concept-wise
separation. At the same time, CLIPScore remains comparable and is
slightly higher for SAEParate (29.323 vs.\ 29.158), suggesting that the
moderate reconstruction trade-off does not translate into a clear
degradation in generated-image quality. These results support our
design objective: for selective diffusion unlearning, preserving
reconstruction fidelity alone is not sufficient, and allowing a modest
reconstruction trade-off can yield substantially more
concept-discriminative sparse representations.

\begin{table}[h]
    \centering
    \caption{
    Reconstruction--separation trade-off between Vanilla SAE (SAeUron) and
    SAEParate under the subset setting (Appendix~\ref{app:subset_setting}), with a matched budget of 20K steps. Lower FVU and centroid margin are better, whereas
    higher cluster rate and CLIPScore are better.
    }
    \label{tab:reconstruction_tradeoff}
    \small
    \setlength{\tabcolsep}{6pt}
    \renewcommand{\arraystretch}{1.05}
    \begin{tabular}{lcccc}
        \toprule
        Method
        & FVU ($\downarrow$)
        & Cluster Rate ($\uparrow$)
        & Centroid Margin ($\downarrow$)
        & CLIPScore ($\uparrow$) \\
        \midrule
        Vanilla SAE (SAeUron)
        & \textbf{0.172}
        & 0.93
        & 0.73
        & 29.158 \\
        \textbf{SAEParate}
        & 0.217
        & \textbf{1.00}
        & \textbf{0.06}
        & \textbf{29.323} \\
        \bottomrule
    \end{tabular}
\end{table}

\clearpage

\subsection{Extended Overlap Analysis}
\label{app:full_overlap_analysis}

We provide the full class-wise overlap analysis for both object and joint style--object unlearning, extending the subset visualization presented in Figure~\ref{fig:overlap_details_heatmap}. Specifically, we measure (i) the pairwise overlap counts between concept-specific selected feature groups and (ii) the pairwise maximum cosine similarity between these groups.

\begin{figure}[h]
\centering
\begin{minipage}[c]{0.5\linewidth}
\centering
\includegraphics[width=\linewidth]{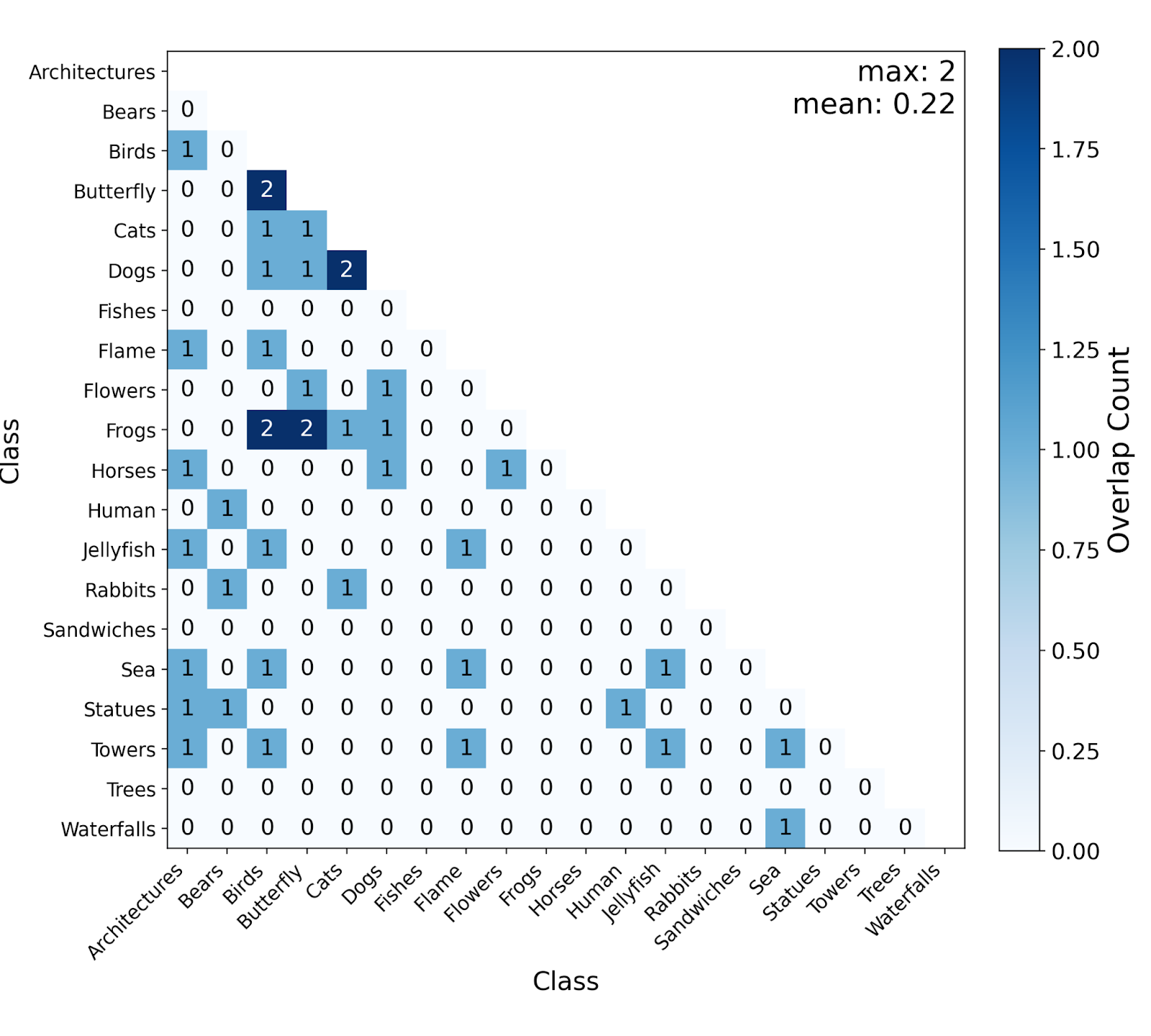}

    \vspace{0pt}
    {\small Vanilla SAE (SAeUron)}
\end{minipage}%
\begin{minipage}[c]{0.5\linewidth}
    \centering
    \includegraphics[width=\linewidth]{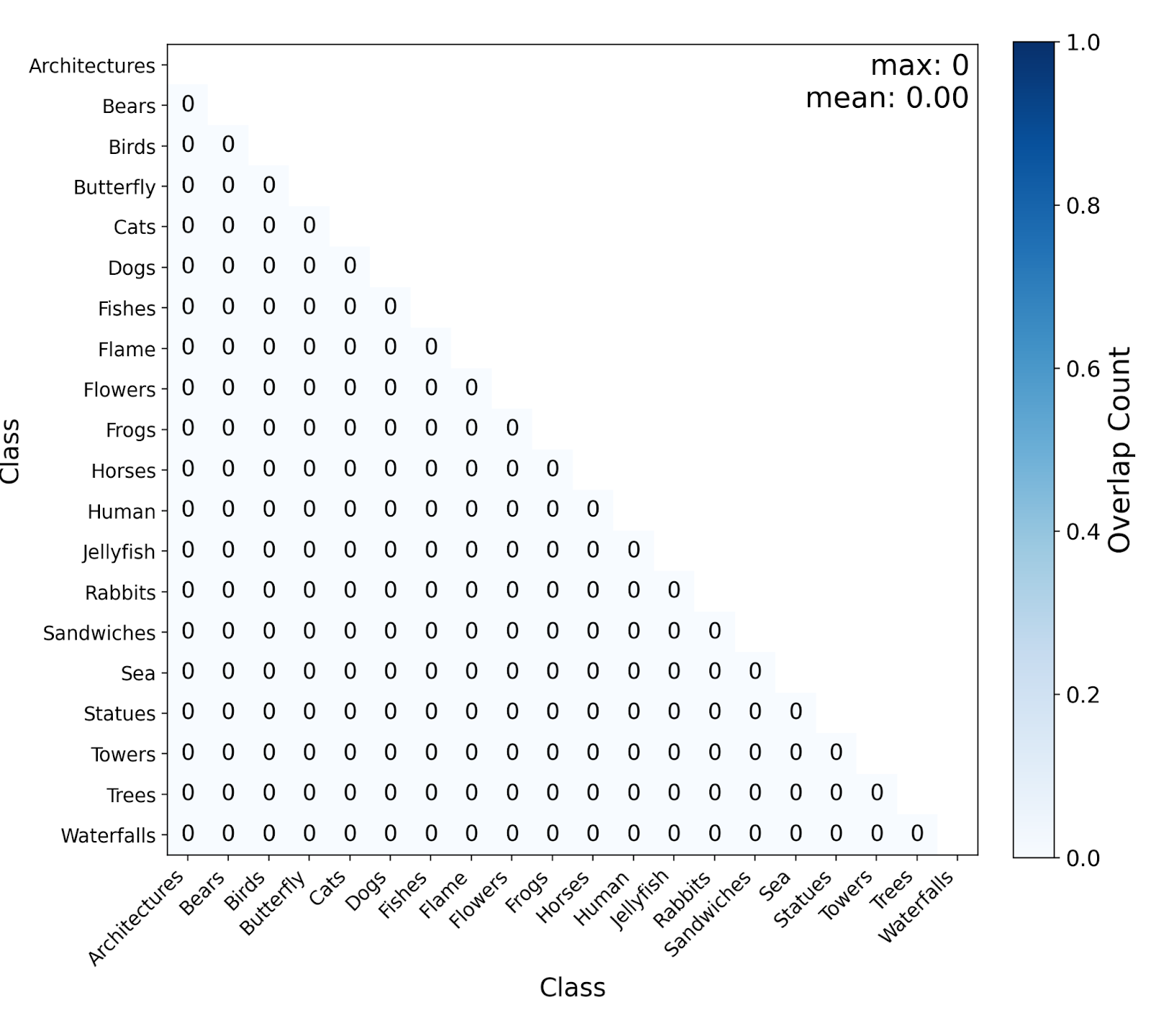}
    
    \vspace{0pt}
    {\small \textbf{SAEParate (Ours)}}
\end{minipage}
\caption{Full class-wise overlap counts of selected feature groups for object unlearning.}
\label{fig:overlap_count_heatmap_full_object}

\end{figure}

\begin{figure}[h]
\centering
\begin{minipage}[c]{0.5\linewidth}
\centering
\includegraphics[width=\linewidth]{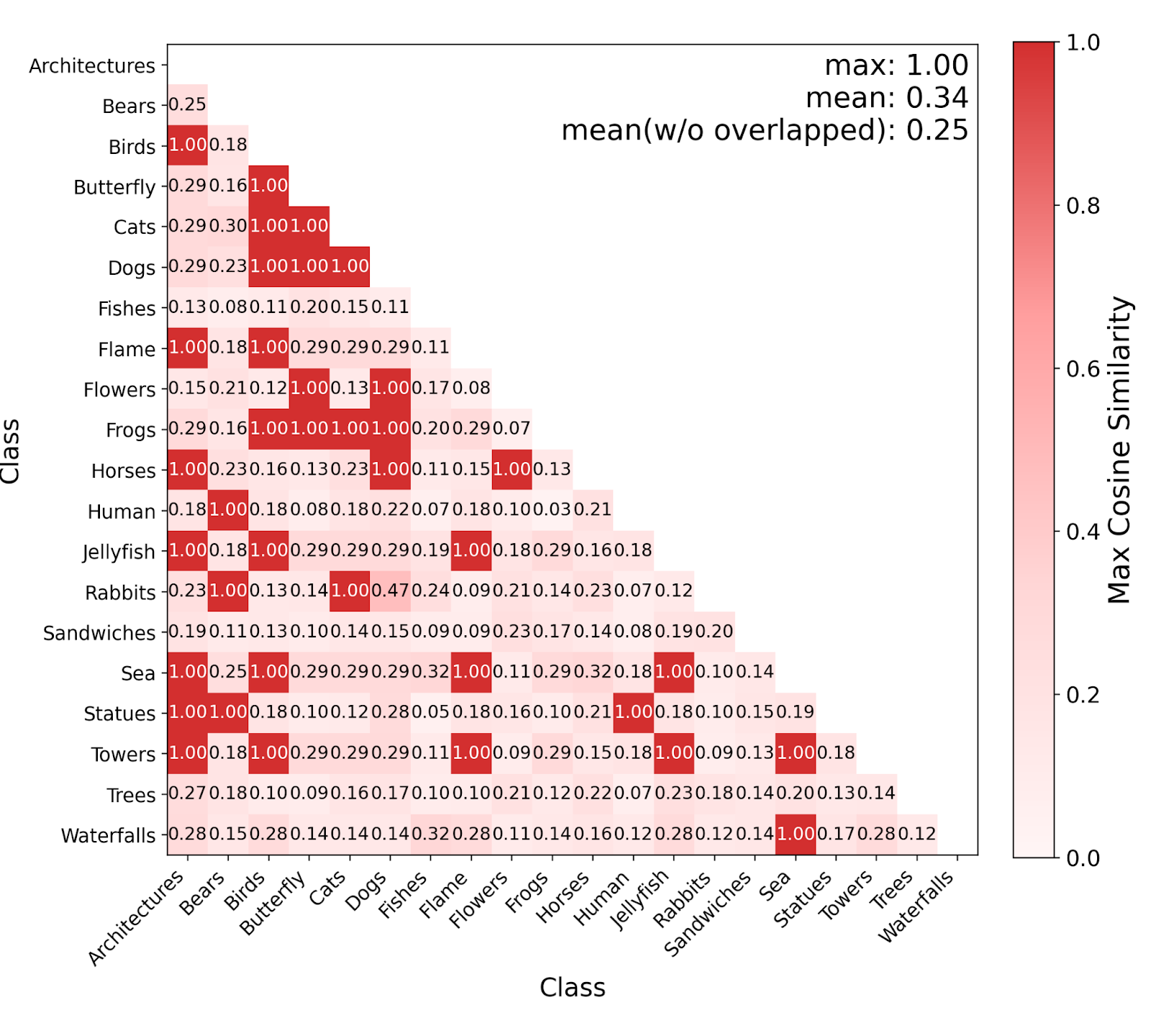}

    \vspace{0pt}
    {\small Vanilla SAE (SAeUron)}
\end{minipage}%
\begin{minipage}[c]{0.5\linewidth}
    \centering
    \includegraphics[width=\linewidth]{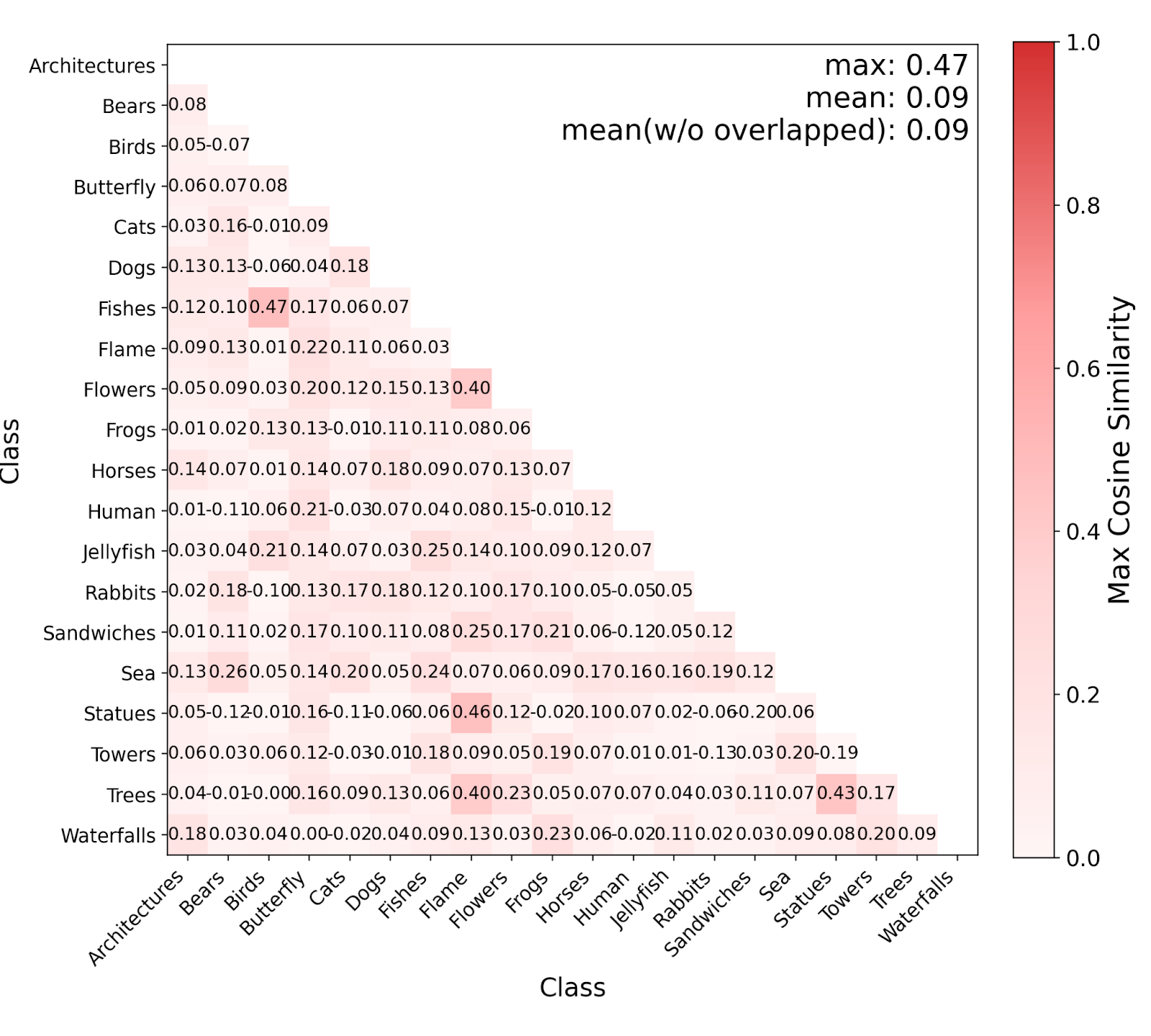}

    \vspace{0pt}
    {\small \textbf{SAEParate (Ours)}}
\end{minipage}
\caption{Full class-wise maximum cosine similarity between selected feature groups for object unlearning.}
\label{fig:overlap_cosine_heatmap_full_object}

\end{figure}

For object unlearning, Figure~\ref{fig:overlap_count_heatmap_full_object} shows that the Vanilla SAE (SAeUron) exhibits substantial redundancy in selected feature groups across different concepts, whereas SAEParate considerably reduces such overlap. Similarly, Figure~\ref{fig:overlap_cosine_heatmap_full_object} shows lower pairwise maximum cosine similarity under SAEParate, indicating that concept-specific feature groups become more distinct in the learned latent space.

\clearpage

\begin{figure}[h]
\centering
\includegraphics[width=\linewidth]{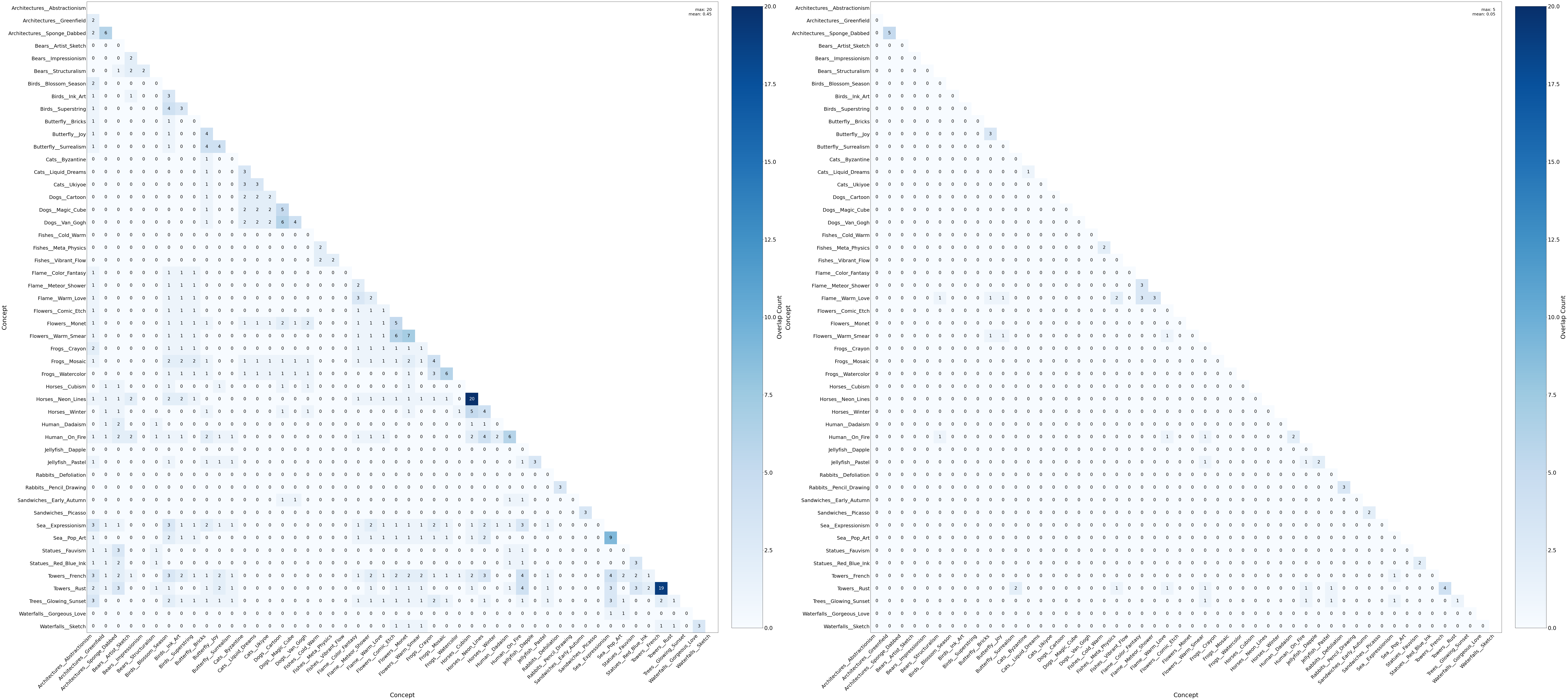}

\vspace{1mm}
\begin{minipage}[t]{0.49\linewidth}
    \centering
    {\small Vanilla SAE (SAeUron)}
\end{minipage}%
\begin{minipage}[t]{0.49\linewidth}
    \centering
    {\small \textbf{SAEParate (Ours)}}
\end{minipage}

\caption{Full pairwise overlap counts of selected feature groups for joint style--object unlearning. SAEParate exhibits substantially less feature sharing across different style--object concepts.}
\label{fig:overlap_count_heatmap_full_joint}
\end{figure}

\begin{figure}[h]
\centering
\includegraphics[width=\linewidth]{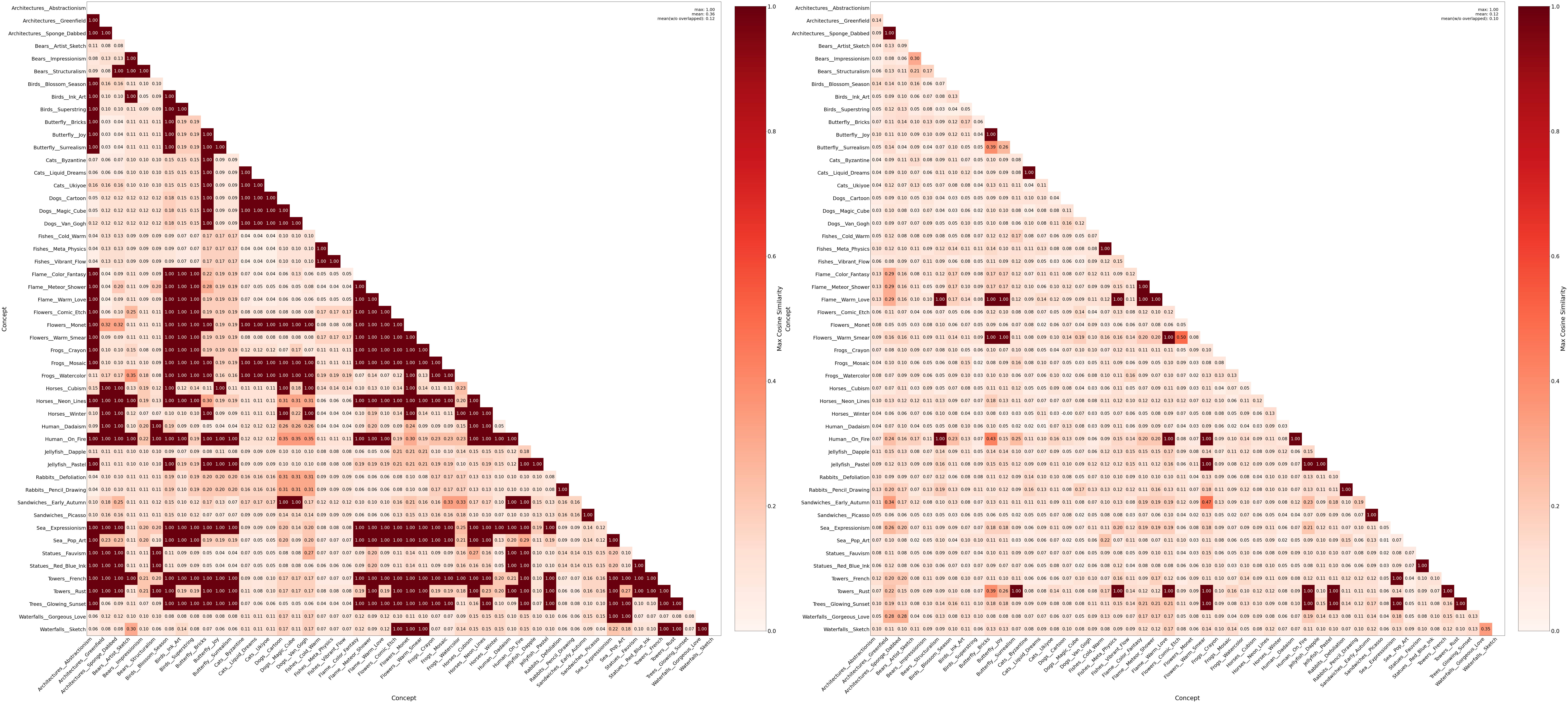}

\vspace{1mm}
\begin{minipage}[t]{0.49\linewidth}
    \centering
    {\small Vanilla SAE (SAeUron)}
\end{minipage}%
\begin{minipage}[t]{0.49\linewidth}
    \centering
    {\small \textbf{SAEParate (Ours)}}
\end{minipage}

\caption{Full pairwise maximum cosine similarity between selected feature groups for joint style--object unlearning. Lower off-diagonal similarities indicate improved separation between concept-specific feature groups.}
\label{fig:overlap_cosine_heatmap_full_joint}
\end{figure}

We further extend the analysis to joint style--object unlearning. Figures~\ref{fig:overlap_count_heatmap_full_joint} and~\ref{fig:overlap_cosine_heatmap_full_joint} show that SAEParate reduces both selected-feature overlap and pairwise maximum cosine similarity, indicating improved separation even among concepts sharing an object or style.

\clearpage
\subsection{Robustness to Concept Restoration and Prompt Variations}
\label{app:robustness}

\paragraph{Concept-Restoration Attacks.}
We first examine whether removed concepts can be restored through adversarial prompting. Specifically, we consider Ring-A-Bell~\cite{tsai2024ring} and UnlearnDiffAtk~\cite{zhang2024generate} under object unlearning, and compare SAEParate with SAeUron. For Ring-A-Bell, we extend the attack to object concepts in UnlearnCanvas by constructing concept-present/concept-absent prompt pairs and estimating the corresponding empirical concept vectors. We also evaluate UnlearnDiffAtk, which iteratively optimizes the input prompt using gradients from the target diffusion model.

\begin{table}[ht]
\centering
\caption{Robustness against Ring-A-Bell and UnlearnDiffAtk under object unlearning. We report unlearning accuracy (UA) before and after each attack.}
\label{tab:attack_robustness}
\small
\setlength{\tabcolsep}{6pt}
\begin{tabular}{lcccc}
\toprule
& \multicolumn{2}{c}{UnlearnDiffAtk} & \multicolumn{2}{c}{Ring-A-Bell} \\
\cmidrule(lr){2-3} \cmidrule(lr){4-5}
Method & Before ($\uparrow$) & After ($\uparrow$) & Before ($\uparrow$) & After ($\uparrow$) \\
\midrule
Vanilla SAE (SAeUron)
& 83.7 & 34.2
& 82.0 & 79.9 \\
\addlinespace[2pt]
\textbf{SAEParate}
& \textbf{94.9} & \textbf{49.7}
& \textbf{93.6} & \textbf{92.8} \\
\bottomrule
\end{tabular}
\end{table}

As shown in Table~\ref{tab:attack_robustness}, SAEParate achieves higher post-attack UA than SAeUron under both attacks, indicating that its unlearning effect is better preserved under concept-restoration attempts.

\paragraph{Prompt Variations.}
We next examine whether SAEParate generalizes beyond the fixed prompt template used during training and evaluation. For joint style--object unlearning, we consider three types of prompt variations: template paraphrasing, object synonyms, and natural-language reformulations of style descriptions.

\begin{table}[ht]
\centering
\caption{Robustness of SAEParate to prompt variations under joint style--object unlearning.}
\label{tab:prompt_variations}
\small
\setlength{\tabcolsep}{7pt}
\begin{tabular}{lcccc}
\toprule
Setting & UA ($\uparrow$) & SC ($\uparrow$) & OC ($\uparrow$) & Avg. ($\uparrow$) \\
\midrule
Original prompts & \textbf{86.2\%} & 81.8\% & \textbf{91.4\%} & \textbf{86.5\%} \\
Reformulated prompts & 81.3\% & \textbf{82.0\%} & 91.3\% & 84.9\% \\
\bottomrule
\end{tabular}
\end{table}

As shown in Table~\ref{tab:prompt_variations}, obtained under the subset setting (Appendix~\ref{app:subset_setting}), reformulating the prompts lowers UA from 86.2\% to 81.3\%, whereas SC and OC remain essentially unchanged (81.8\% to 82.0\% and 91.4\% to 91.3\%). The average therefore decreases only modestly, from 86.5\% to 84.9\%, and the drop is confined to target removal rather than retention, suggesting that the learned representations are not tied to the exact prompt template used during training.

\clearpage

\subsection{Gender-Distinct Nudity Unlearning}
\label{app:nudity_unlearning}

For real-world adaptation of our method, we consider gender-specific nudity unlearning scenarios. We collect internal activations from a pre-trained Stable Diffusion v1.4 model~\cite{rombach2022high} using randomly selected 5K captions from COCO~\cite{lin2014microsoft} train2014, supplemented with a small number of prompts containing \textit{``naked man''} and \textit{``naked woman''}. We train and apply the SAE on the \texttt{up.1.1} cross-attention block following the object unlearning setup.

While prior works, including SAeUron, mainly treat \textit{``nudity''} as a single target concept, we decompose it into two more fine-grained concepts, \textit{``naked woman''} and \textit{``naked man''}, to evaluate whether SAEParate can separate closely related real-world concepts. 

After training the SAE, we collect SAE latent representations and identify target-specific feature groups using the feature importance score described in Appendix~\ref{app:feature_importance_scoring}. We use the simple prompts \textit{``naked woman''} and \textit{``naked man''} during this feature selection stage. We then perform a concept-wise hyperparameter search for each target concept using 20 validation prompts related to \textit{``naked woman''} and \textit{``naked man''}. For evaluation, we select sexually related prompts from the I2P benchmark~\cite{schramowski2023safe} and use the NudeNet detector to classify generated images, which provides gender-specific exposed nudity detection results. We additionally report CLIPScore~\cite{hessel2021clipscore} and FID to measure the general image generation quality, using images generated from 1K COCO 2014 validation captions.

\begin{table}[h]
\centering
\caption{Gender-specific nudity evaluation on I2P prompts. Unlearning and Retention Accuracy are computed relative to the base model and clipped to $[0,100]\%$.}
\label{tab:nudenet_gender_class_counts}
\small
\setlength{\tabcolsep}{4.5pt}
\resizebox{\linewidth}{!}{%
\begin{tabular}{lc|cc|cc|cc|cc}
\toprule
\textbf{Method}
& \textbf{Concept}
& \textbf{Breasts (F)}
& \textbf{Genitalia (F)}
& \textbf{Breasts (M)}
& \textbf{Genitalia (M)}
& \textbf{UA. ($\uparrow$)}
& \textbf{RA. ($\uparrow$)}
& \textbf{CLIPScore ($\uparrow$)}
& \textbf{FID ($\downarrow$)} \\
\midrule
\multirow{2}{*}{Vanilla SAE (SAeUron)}
& Naked Woman
& 242 & 11
& 17 & 2
& 25.6\%
& 100.0\%
& 30.58 & 72.28 \\
& Naked Man
& 234 & 8
& 14 & 1
& 0.0\%
& 71.2\%
& 30.53 & 72.18 \\
\midrule
\multirow{2}{*}{\textbf{SAEParate (Ours)}}
& Naked Woman
& 85 & 4
& 13 & 1
& 73.8\%
& 100.0\%
& 30.70 & 70.12 \\
& Naked Man
& 250 & 16
& 0 & 0
& 100.0\%
& 78.2\%
& 30.70 & 69.98 \\
\midrule
SD v1.4 (Base)
& --
& 316 & 24
& 11 & 3
& --
& --
& 31.06 & 71.63 \\
\bottomrule
\end{tabular}%
}
\end{table}

\begin{figure}[h]
    \centering
    \begin{minipage}[c]{0.49\linewidth}
        \centering
        \includegraphics[width=\linewidth]{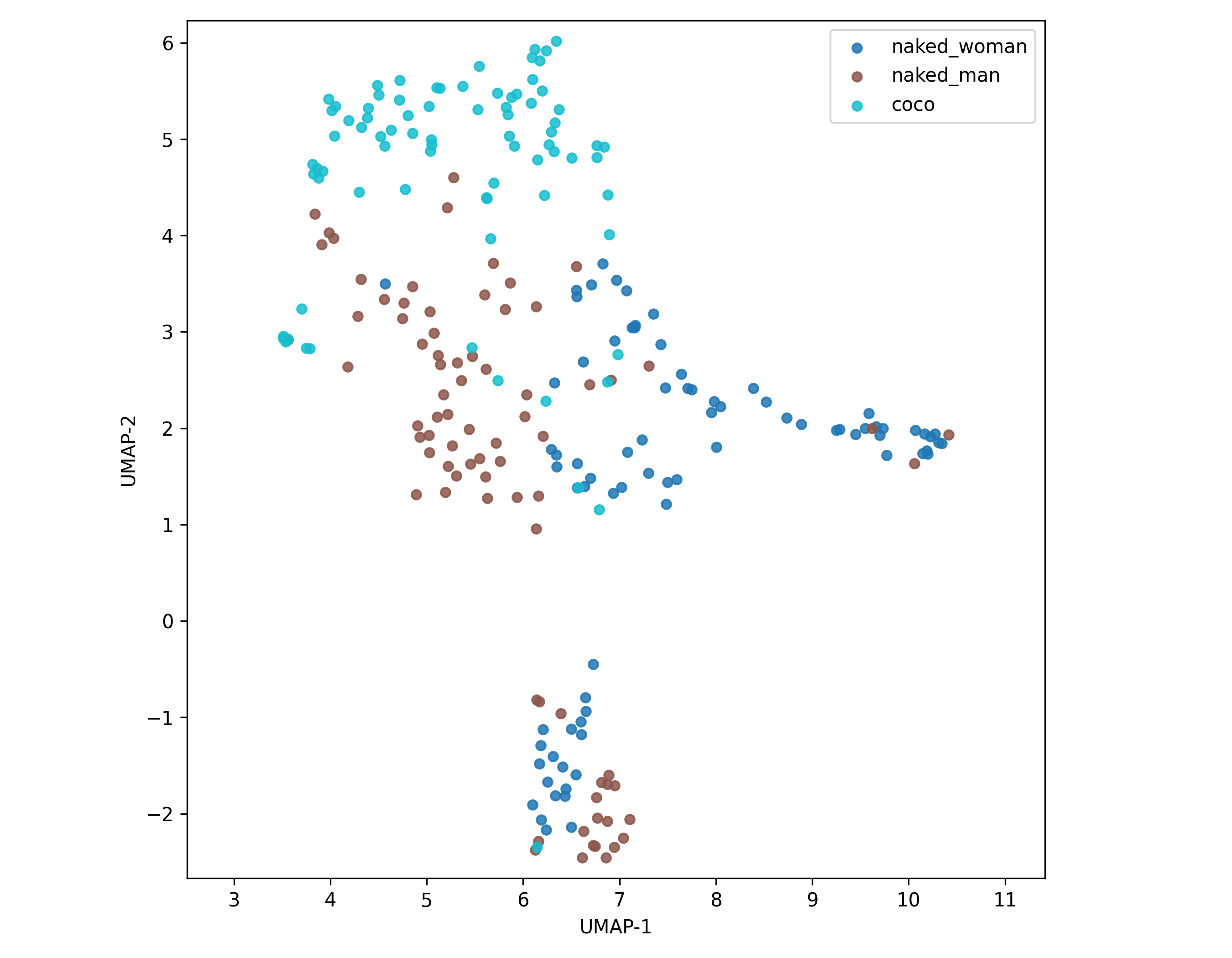}
        \vspace{1pt}
        {\small Vanilla SAE (SAeUron)} \\
        {\scriptsize Cluster Rate ($\uparrow$): 0.96, Centroid Margin ($\downarrow$): 0.60}
    \end{minipage}
    \hfill
    \begin{minipage}[c]{0.49\linewidth}
        \centering
        \includegraphics[width=\linewidth]{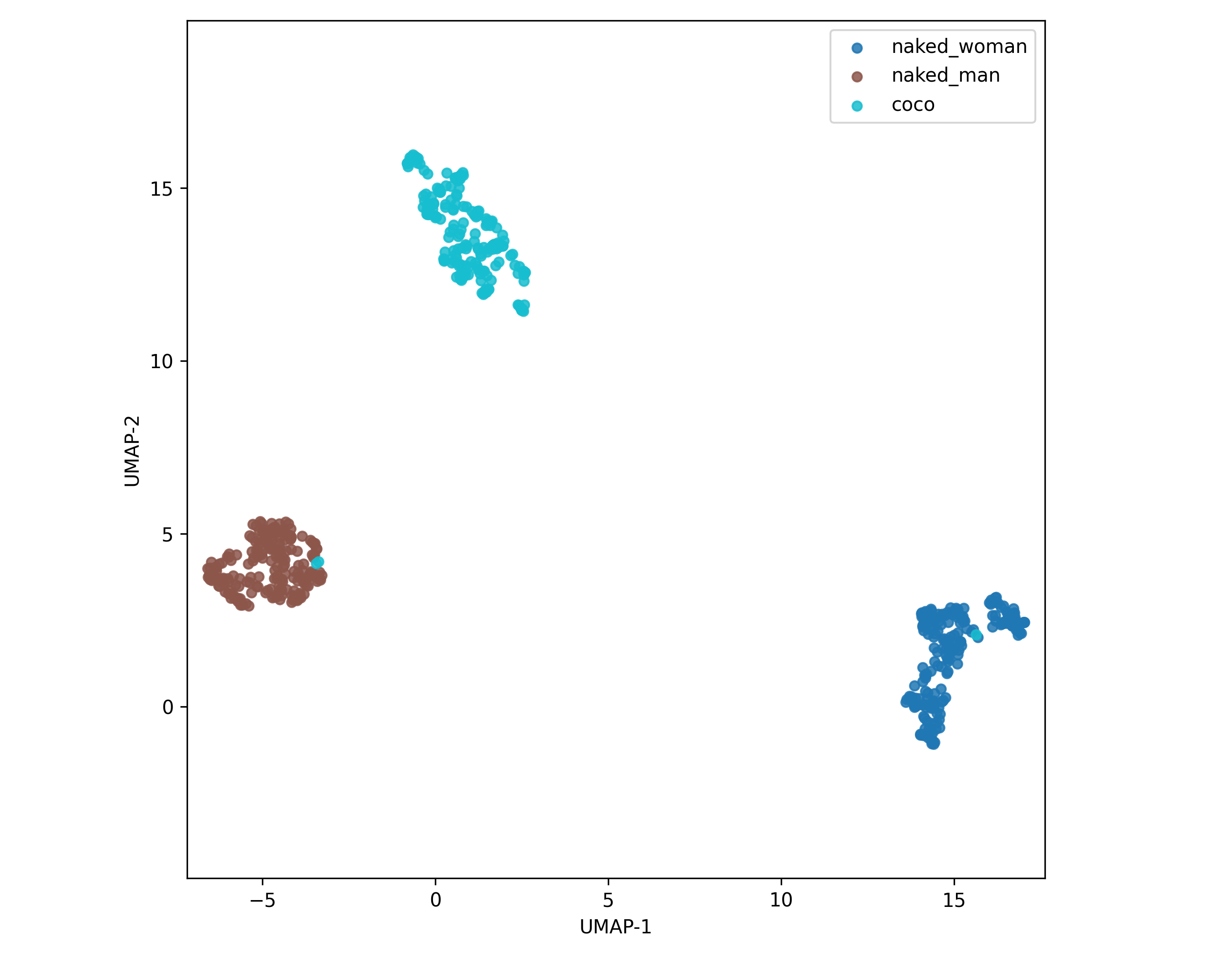}
        \vspace{1pt}
        {\small \textbf{SAEParate (Ours)}} \\
        {\scriptsize Cluster Rate ($\uparrow$): \textbf{1.00}, Centroid Margin ($\downarrow$): \textbf{0.27}}
    \end{minipage}
    \caption{UMAP visualization of sparse SAE latent representations for nudity unlearning: Vanilla SAE (SAeUron) (left) and SAEParate (right).}
    \label{fig:sae_latent_clusters_nudity}
\end{figure}

As shown in Table~\ref{tab:nudenet_gender_class_counts}, SAEParate achieves high unlearning accuracy for each target concept while maintaining or improving retention of the non-target concept,
e.g., preserving \textit{``naked woman''} when unlearning
\textit{``naked man''}, compared to SAeUron. Interestingly, Figure~\ref{fig:sae_latent_clusters_nudity} shows that SAEParate learns a concept-discriminative SAE latent space that separates fine-grained nudity concepts from each other as well as from normal COCO captions. These results demonstrate that our method can robustly separate desired concepts across diverse scenarios, suggesting a promising extension to label-available downstream unlearning tasks.

\clearpage

\subsection{Fine-Grained Unlearning with Binary Forget/Retain Supervision}
\label{app:golden-retriever}

SAEParate requires supervision to define positive and negative relationships during training, but it does not require a fine-grained taxonomy with a separate label for every retained subclass. To evaluate this reduced-label setting directly, we consider a binary forget/retain formulation for fine-grained dog-breed unlearning. Golden Retriever samples form the \emph{forget} group, while 13 non-target dog breeds are jointly assigned to a single \emph{retain} group, without individual breed labels. This setting is particularly challenging because the target is a subordinate category of the broader \emph{dog} concept, requiring breed-specific evidence to be suppressed while shared canine features and other dog breeds are preserved.

In the standard multi-class setting, positive pairs are defined as
\[
\mathcal{P}(i)
=
\{j \neq i \mid y_j = y_i\}.
\]
Under binary forget/retain supervision, $y_i$ instead indicates whether a sample belongs to the forget or retain group. Positive pairs are constructed within each binary group, while negative pairs are formed across the two groups. Therefore, the same contrastive formulation can be applied without requiring a separate label for every retained subclass.

\begin{table}[ht]
\centering
\caption{
Fine-grained Golden Retriever unlearning using only binary
forget/retain supervision. All 13 non-target dog breeds are assigned
to a single retain group.
}
\label{tab:golden_retriever_binary}
\small
\setlength{\tabcolsep}{5pt}
\begin{tabular}{lcccc}
\toprule
Method
& UA ($\uparrow$)
& RA ($\uparrow$)
& Dog FID ($\downarrow$)
& COCO CLIPScore ($\uparrow$) \\
\midrule
Vanilla SAE (SAeUron)
& 75.0\%
& 33.1\%
& 214.15
& \textbf{28.56} \\
\textbf{SAEParate}
& \textbf{100.0\%}
& \textbf{77.2\%}
& \textbf{112.34}
& 28.39 \\
\bottomrule
\end{tabular}
\end{table}

As shown in Table~\ref{tab:golden_retriever_binary}, SAEParate achieves substantially stronger Golden Retriever unlearning while preserving the 13 non-target dog breeds more effectively than SAeUron. It also reduces Dog FID from 214.15 to 112.34, while maintaining a comparable COCO CLIPScore, indicating that the improvement does not come from a substantial degradation in general generation quality.

\paragraph{Comparison with TRUST.}
We additionally compare SAEParate with TRUST~\cite{kori2026selective}, which also aims to reduce interference with semantically related retained concepts but does so through selective fine-tuning of concept-relevant cross-attention parameters. We evaluate both methods on the same Golden Retriever forget/retain split.

\begin{table}[ht]
\centering
\caption{
Comparison with TRUST on fine-grained Golden Retriever unlearning.
Both methods are evaluated using the same forget/retain split.
}
\label{tab:golden_retriever_trust}
\small
\setlength{\tabcolsep}{8pt}
\begin{tabular}{lcc}
\toprule
Method
& UA ($\uparrow$)
& RA ($\uparrow$) \\
\midrule
TRUST
& \textbf{100.0\%}
& 68.8\% \\
\textbf{SAEParate}
& \textbf{100.0\%}
& \textbf{77.2\%} \\
\bottomrule
\end{tabular}
\end{table}

As shown in Table~\ref{tab:golden_retriever_trust}, both methods achieve 100.0\% UA, while SAEParate achieves higher retention accuracy on the 13 non-target dog breeds (77.2\% vs.\ 68.8\%).

Together, these results show that SAEParate does not require a fine-grained taxonomy over all retained concepts. With only binary forget/retain supervision, it achieves a strong unlearning--retention trade-off against SAeUron and preserves closely related non-target breeds more effectively than TRUST at the same unlearning accuracy. Thus, coarse group-level supervision can be sufficient when the objective is to distinguish a target concept from a broader retained group.

\clearpage

\subsection{Generalization to Unseen Concepts}
\label{app:unseen_evaluations}

For measuring the generalization ability of our method, we train the SAE using the first 25 style themes from the list (out of 50 total themes), following the setup of SAeUron Appendix F. All training configurations are kept identical to the style setup described in Appendix~\ref{app:sae_training_details}. For unlearning, we set $\gamma_c=-1.0$ for all concepts, while keeping the remaining unlearning procedure identical to our main method. Table~\ref{tab:half_data_results} compares our method with SAeUron under three evaluation setups. \textit{In-distribution} denotes evaluation restricted to the 25 style themes used for SAE training, while \textit{out-of-distribution} (OOD) denotes evaluation restricted to the remaining 25 unseen style themes. \textit{All data} reports performance on the combined set of seen and unseen style themes. We report UA, IRA, and CRA for each setup to assess whether the learned SAE features generalize beyond the concepts observed during training.

\begin{table}[h]
\centering
\caption{Generalization performance when the SAE is trained on half of the style themes.}
\label{tab:half_data_results}
\vspace{2pt}
\small
\setlength{\tabcolsep}{5pt}
\renewcommand{\arraystretch}{1.05}
\begin{tabular}{ll|cccc}
\toprule
\textbf{Method} & \textbf{Setup} 
& \textbf{UA ($\uparrow$)} 
& \textbf{IRA ($\uparrow$)} 
& \textbf{CRA ($\uparrow$)} 
& \textbf{Avg. ($\uparrow$)} \\
\midrule
\multirow{3}{*}{Vanilla SAE (SAeUron)}
& All data            & 75.00\% & 90.18\% & 80.74\% & 81.97\% \\
& In-distribution     & 99.76\% & 99.23\% & 98.48\% & 99.16\% \\
& Out of distribution & 51.00\% & 67.38\% & 98.36\% & 72.25\% \\
\midrule
\multirow{3}{*}{\textbf{SAEParate (Ours)}}
& All data            & 84.80\% & 83.42\% & 97.40\% & 88.54\% \\
& In-distribution     & 97.00\% & 98.35\% & 96.40\% & 97.25\% \\
& Out of distribution & 72.60\% & 68.49\% & 98.40\% & 79.83\% \\
\bottomrule
\end{tabular}
\end{table}

Notably, SAEParate substantially improves OOD unlearning accuracy
(72.60\% vs.\ 51.00\%). On the full set, SAEParate also achieves much
higher CRA (97.40\% vs.\ 80.74\%), indicating that unlearning a style
causes less collateral damage to images of other styles, including
those unseen during SAE training. This comes at the cost of lower
in-domain retention on the full set (83.42\% vs.\ 90.18\%) and slightly
lower in-distribution performance. We attribute the OOD and CRA gains
to the concept-separated latent space, which reduces feature sharing
across styles.

\clearpage

\subsection{Object Sequential Unlearning}
\label{app:object_sequential_unlearning}

We conduct an object sequential unlearning task, which assumes multiple concept-level unlearning requests and better reflects real-world deployment scenarios. Since UnlearnCanvas provides only a style sequential unlearning setting, we extend the evaluation to object concepts, where Vanilla SAE (SAeUron) tends to struggle more with balancing unlearning and retention.

We sequentially unlearn the following object concepts:
\begin{itemize}
    \item $\mathcal{T}_1$: Architectures
    \item $\mathcal{T}_2$: Bears
    \item $\mathcal{T}_3$: Birds
    \item $\mathcal{T}_4$: Butterfly
    \item $\mathcal{T}_5$: Cats
\end{itemize}

At each step, we add a new target concept to the set of unlearning requests and evaluate performance using UA, IRA, and CRA. We define RA as the average of IRA and CRA, i.e., $\mathrm{RA} = (\mathrm{IRA} + \mathrm{CRA}) / 2$.

As shown in Table~\ref{tab:object_sequential_unlearning}, SAEParate achieves a better balance between sequential unlearning and retention compared to SAeUron. In particular, SAEParate avoids the severe retention collapse observed in SAeUron when the number of unlearned object concepts increases.

\begin{table}[h]
\centering
\caption{
Object sequential unlearning results. 
Each column denotes a new unlearning request, where $\mathcal{T}_1$ is the oldest request.
Each UA row reports the unlearning accuracy for a specific target object.
RA is computed as the average of IRA and CRA, averaged over the active requests in each column.
Results indicating catastrophic retaining failure are highlighted in red.
}
\label{tab:object_sequential_unlearning}
\small
\setlength{\tabcolsep}{6pt}
\renewcommand{\arraystretch}{1.08}

\begin{tabular}{cc|ccccc}
\toprule
\multicolumn{7}{c}{\textbf{Method: Vanilla SAE (SAeUron)}} \\
\midrule
\textbf{Metrics}
&
&
$\mathcal{T}_1$
&
$\mathcal{T}_1 \sim \mathcal{T}_2$
&
$\mathcal{T}_1 \sim \mathcal{T}_3$
&
$\mathcal{T}_1 \sim \mathcal{T}_4$
&
$\mathcal{T}_1 \sim \mathcal{T}_5$
\\
\cmidrule(lr){3-7}
&
&
\multicolumn{5}{c}{\textbf{Unlearning Request}}
\\
\midrule

\multirow{5}{*}{\textbf{UA}}
& $\mathcal{T}_1$
& 33.33\%
& 35.29\%
& 84.31\%
& 100.00\%
& 100.00\%
\\
& $\mathcal{T}_2$
& -
& 65.69\%
& 92.16\%
& 100.00\%
& 100.00\%
\\
& $\mathcal{T}_3$
& -
& -
& 91.50\%
& 100.00\%
& 100.00\%
\\
& $\mathcal{T}_4$
& -
& -
& -
& 93.63\%
& 97.55\%
\\
& $\mathcal{T}_5$
& -
& -
& -
& -
& 98.04\%
\\

\midrule
\textbf{RA}
&
& 97.91\%
& 93.27\%
& 87.32\%
& \cellcolor[HTML]{EFC6C6}18.40\%
& \cellcolor[HTML]{EFC6C6}4.96\%
\\

\bottomrule
\end{tabular}

\vspace{8pt}

\begin{tabular}{cc|ccccc}
\toprule
\multicolumn{7}{c}{\textbf{Method: SAEParate}} \\
\midrule
\textbf{Metrics}
&
&
$\mathcal{T}_1$
&
$\mathcal{T}_1 \sim \mathcal{T}_2$
&
$\mathcal{T}_1 \sim \mathcal{T}_3$
&
$\mathcal{T}_1 \sim \mathcal{T}_4$
&
$\mathcal{T}_1 \sim \mathcal{T}_5$
\\
\cmidrule(lr){3-7}
&
&
\multicolumn{5}{c}{\textbf{Unlearning Request}}
\\
\midrule

\multirow{5}{*}{\textbf{UA}}
& $\mathcal{T}_1$
& 96.08\%
& 70.59\%
& 92.16\%
& 68.63\%
& 98.04\%
\\
& $\mathcal{T}_2$
& -
& 85.29\%
& 92.16\%
& 84.31\%
& 89.22\%
\\
& $\mathcal{T}_3$
& -
& -
& 88.24\%
& 80.39\%
& 86.93\%
\\
& $\mathcal{T}_4$
& -
& -
& -
& 70.59\%
& 66.67\%
\\
& $\mathcal{T}_5$
& -
& -
& -
& -
& 73.33\%
\\

\midrule
\textbf{RA}
&
& 97.11\%
& 93.94\%
& 92.51\%
& 88.83\%
& 87.67\%
\\

\bottomrule
\end{tabular}
\end{table}

\begin{figure}[h]
    \centering
    \includegraphics[width=1\linewidth]{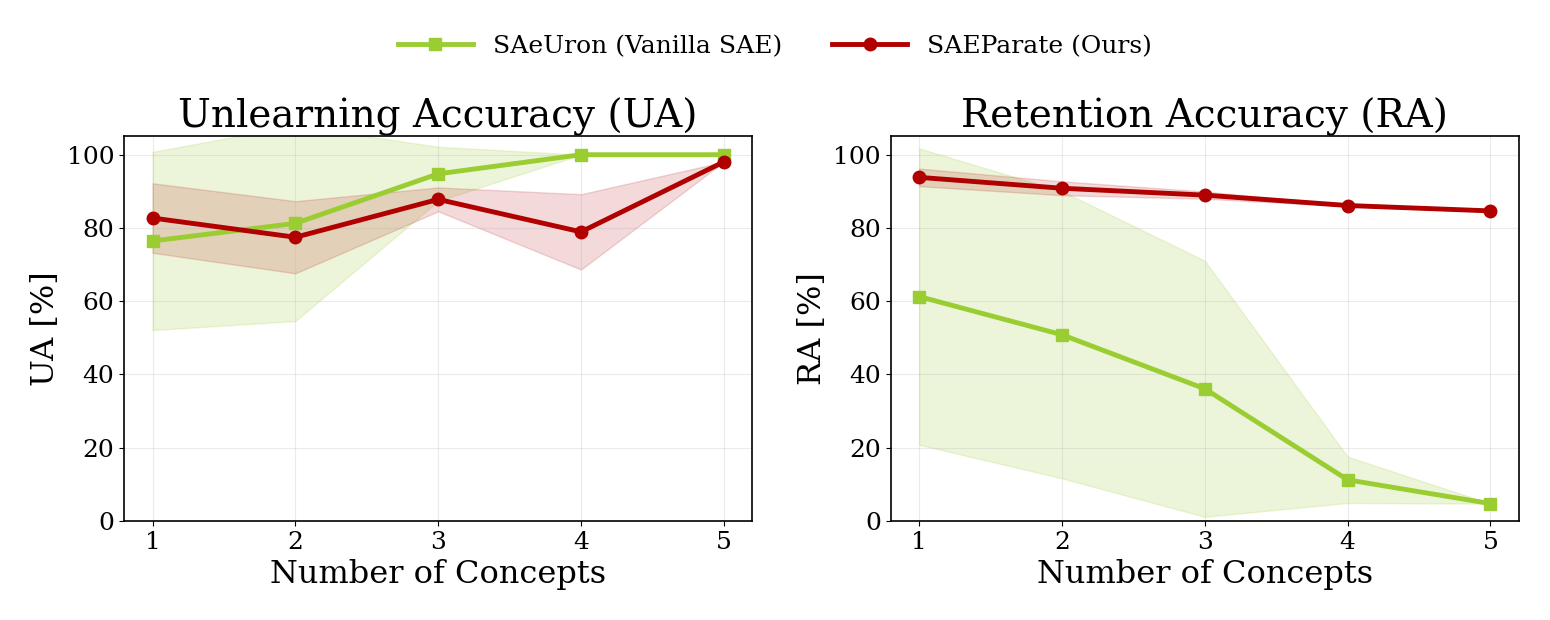}

    \caption{Sequential object unlearning results across increasing numbers of target concepts. SAEParate preserves retention accuracy substantially better than Vanilla SAE (SAeUron) under multiple unlearning requests.}
    \label{fig:seq_plot}
\end{figure}

\begin{figure}[h]
    \centering
    \includegraphics[width=1\linewidth]{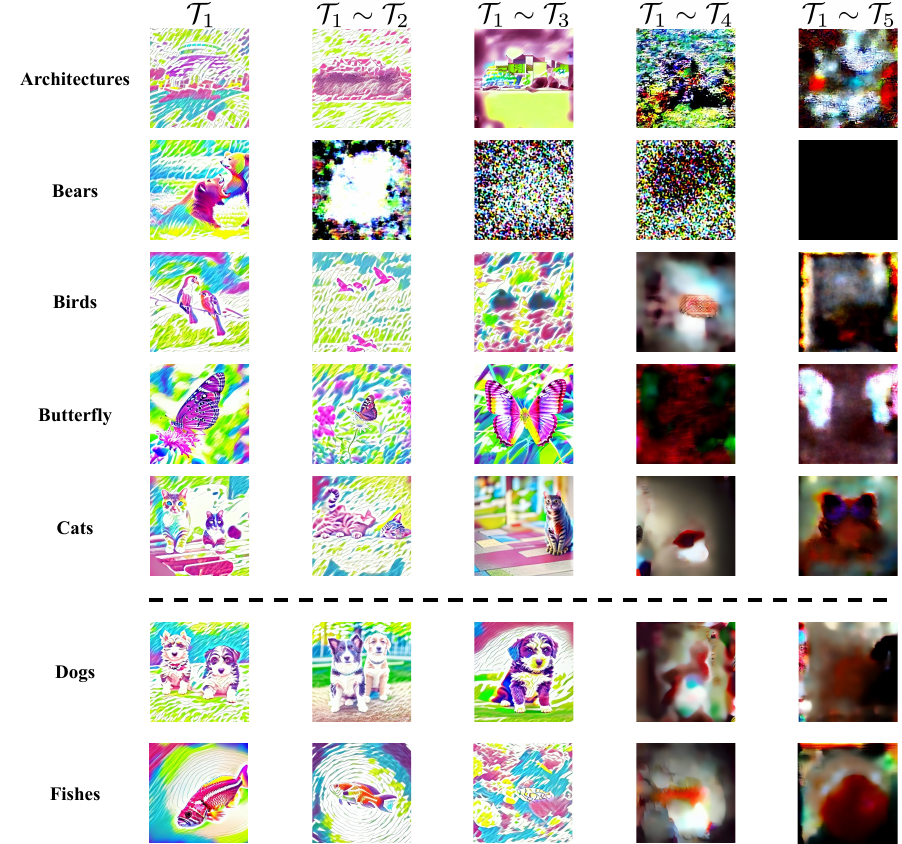}
    
    \vspace{10pt}
    {\large Vanilla SAE (SAeUron)}
    \caption{
    Qualitative visualization of object sequential unlearning for Vanilla SAE (SAeUron). 
    The first five rows correspond to the target object classes
$\mathcal{T}_1$ through $\mathcal{T}_5$, while the columns show
cumulative unlearning requests. 
    The ``Dogs'' and ``Fishes'' rows are non-target object classes and are used to evaluate retention.
    }
    \label{fig:seq_viz_saeuron}
\end{figure}

\begin{figure}[h]
    \centering
    \includegraphics[width=1\linewidth]{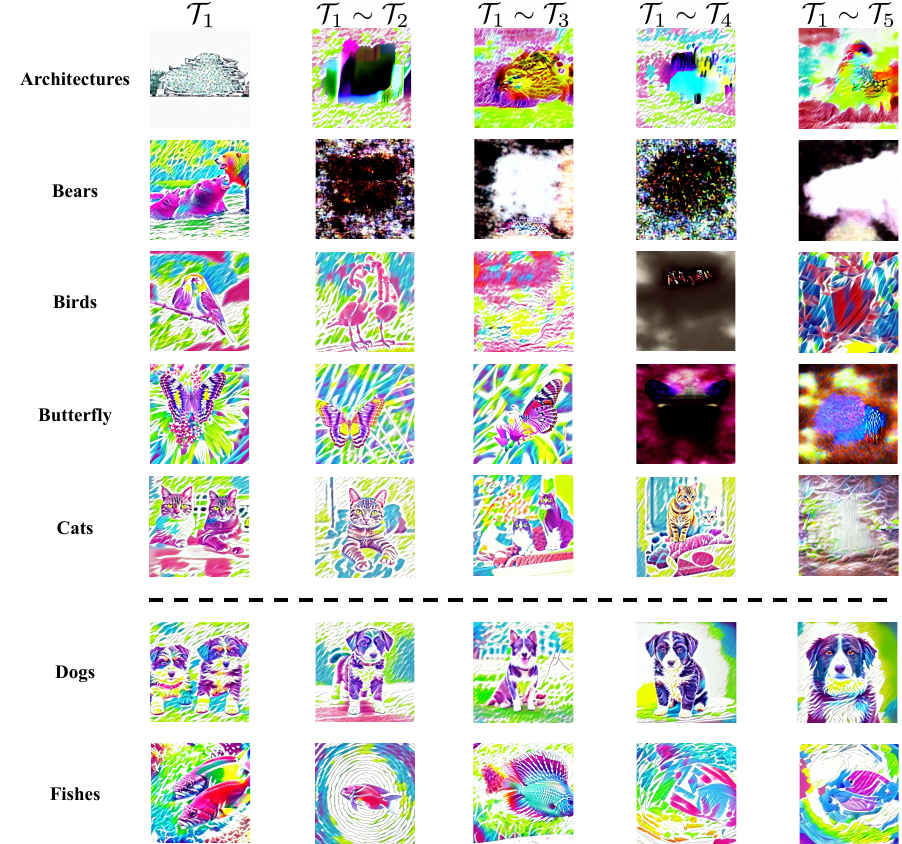}
    
    \vspace{10pt}
    {\large \textbf{SAEParate (Ours)}}
    \caption{
    Qualitative visualization of object sequential unlearning for SAEParate (Ours). 
    The first five rows correspond to the target object classes
$\mathcal{T}_1$ through $\mathcal{T}_5$, while the columns show
cumulative unlearning requests.
    The ``Dogs'' and ``Fishes'' rows are non-target object classes and are used to evaluate retention.
    }
    \label{fig:seq_viz_saeparate}
\end{figure}

We also provide qualitative results in Figures~\ref{fig:seq_viz_saeuron} and~\ref{fig:seq_viz_saeparate}. SAEParate robustly suppresses multiple target concepts while largely preserving non-target concepts, whereas Vanilla SAE fails to maintain retention after several sequential unlearning requests. These results further suggest that SAEParate reduces redundant feature selection by directly separating latent representations, allowing target concepts to be isolated in more concept-specific features.

\clearpage

\subsection{Compatibility with Alternative SAE Training Objectives}
\label{app:matryoshka_compatibility}

We further examine whether the concept-separation objective of SAEParate
can be combined with alternative SAE training schemes. Specifically, we
incorporate the fixed-prefix, multi-scale training objective of Matryoshka
SAE~\cite{bussmann2025learning} into the SAEParate pipeline and evaluate
the resulting model under the subset setting (Appendix~\ref{app:subset_setting}) in the object-unlearning setting.

\begin{table}[ht]
\centering
\caption{
Compatibility of SAEParate with the Matryoshka SAE training objective.
All results are obtained under the subset setting (Appendix~\ref{app:subset_setting}).
}
\label{tab:matryoshka_compatibility}
\footnotesize
\setlength{\tabcolsep}{3pt}
\renewcommand{\arraystretch}{1.05}
\begin{tabular}{lcccccc}
\toprule
Method
& UA ($\uparrow$)
& IRA ($\uparrow$)
& CRA ($\uparrow$)
& Avg. ($\uparrow$)
& \shortstack{Cluster\\Rate ($\uparrow$)}
& \shortstack{Centroid\\Margin ($\downarrow$)} \\
\midrule
Vanilla SAE (SAeUron)
& 58.14\%
& 96.74\%
& \textbf{97.45\%}
& 84.11\%
& 0.93
& 0.73 \\

Matryoshka SAE
& 81.18\%
& 95.61\%
& 91.06\%
& 89.28\%
& 0.94
& 0.30 \\

\textbf{SAEParate + Matryoshka}
& \textbf{93.14\%}
& \textbf{97.43\%}
& 94.39\%
& \textbf{94.99\%}
& \textbf{0.99}
& \textbf{0.02} \\
\bottomrule
\end{tabular}
\end{table}

Combining SAEParate with Matryoshka SAE substantially improves both
unlearning performance and latent-space separation over Matryoshka SAE
alone, increasing the average accuracy from 89.28\% to 94.99\% while
improving the Cluster Rate from 0.94 to 0.99 and reducing the Centroid
Margin. These results suggest that the concept-separation objective of
SAEParate is complementary to alternative SAE training objectives rather
than being tied to a specific SAE formulation. More broadly, this
compatibility indicates that SAEParate can benefit from advances in SAE
design, providing a potential avenue for alleviating SAE-specific
limitations such as feature splitting and absorption without changing the
core concept-separation framework.

\clearpage

\subsection{Preliminary Extension to an MM-DiT Backbone}
\label{app:mmdit_extension}

Extending SAEParate beyond the U-Net architecture used in our main experiments to MM-DiT-based diffusion models requires several architecture-specific considerations. In particular, MM-DiT jointly updates image and text tokens across transformer blocks, so target-concept information retained in the text-token stream may be reintroduced into the image stream after an intervention. Moreover, standalone cross-attention is replaced by joint attention over image and text tokens, requiring the cross-modal relevance signal used by SAEParate to be reformulated.

One possible adaptation is to intervene on the post-block image-token residual stream, which preserves image-token structure while incorporating both attention and feed-forward updates. Likewise, the image-query-to-text-key portion of joint attention can serve as an analogue of the cross-attention signal used in our U-Net implementation for estimating concept relevance. The remaining components of SAEParate, including contrastive representation learning, feature calibration and suppression, and the nonlinearity-enhanced encoder, are not inherently tied to U-Net-specific operators. A systematic treatment of these architecture-specific design choices is beyond the scope of this work and remains an important direction for future study. Nevertheless, we conduct a preliminary experiment to examine whether a simplified adaptation can retain the benefits of SAEParate on an MM-DiT backbone.

\paragraph{Preliminary Experiments on SD3.5-Medium-Turbo.}
We evaluate this simplified adaptation on Stable Diffusion 3.5 Medium Turbo, released by TensorArt Studios. We insert the SAE into the image-token stream of MM-DiT block 8 out of 24. The insertion depth is selected using a 20-way class-separability probe across blocks; separability remains relatively stable across early and middle blocks and begins to degrade beyond approximately block 14.

\begin{table}[ht]
\centering
\caption{
Preliminary object-unlearning results on the MM-DiT-based Stable Diffusion 3.5 Medium Turbo backbone. RA denotes object-level retention accuracy, and Cluster Rate measures class-wise separation in the learned SAE latent space.
}
\label{tab:mmdit_results}
\small
\setlength{\tabcolsep}{8pt}
\begin{tabular}{lcccc}
\toprule
Method
& UA ($\uparrow$)
& RA ($\uparrow$)
& Avg. ($\uparrow$)
& Cluster Rate ($\uparrow$) \\
\midrule
Vanilla SAE (SAeUron) 
& 43.3\%
& \textbf{70.8\%}
& 57.1\%
& 0.435 \\
\textbf{SAEParate}
& \textbf{71.7\%}
& 60.3\%
& \textbf{66.0\%}
& \textbf{0.498} \\
\bottomrule
\end{tabular}
\end{table}

Unlike UnlearnCanvas, this backbone does not provide a model specifically fine-tuned for the predefined object and style concepts. We therefore restrict retention evaluation to object-level RA. The absolute UA and RA values should also be interpreted with caution
because the UnlearnCanvas ViT classifiers were fine-tuned on
UnlearnCanvas images, which may differ in distribution from
SD3.5 Medium Turbo outputs. Nevertheless, the relative comparison remains informative because both methods are evaluated using the same backbone, prompts, generation settings, and classifiers.

As shown in Table~\ref{tab:mmdit_results}, the simplified SAEParate adaptation improves UA from 43.3\% to 71.7\% and the overall average from 57.1\% to 66.0\%, while also increasing the Cluster Rate from 0.435 to 0.498. These results demonstrate that the concept-separation objective can also improve unlearning performance on an MM-DiT-based backbone, despite differences in architecture and evaluation settings.
\clearpage

\section{Additional Visualizations}

\subsection{Additional Latent Space Visualizations}
\label{app:cluster_viz_more}

To further analyze the structure of the learned latent space, we visualize sparse latent representations using UMAP~\cite{mcinnes2018umap} and measure their concept-wise clustering behavior. The analysis is conducted on the validation set used for computing feature importance scores, which contains 20 prompts per target style, 80 prompts per target object and per target style--object combination for the joint setting. In each UMAP visualization, different colors denote different concepts, following the class and theme categories defined in UnlearnCanvas.

\begin{figure}[h]
    \centering
    \begin{minipage}[c]{0.49\linewidth}
        \centering
        \includegraphics[width=\linewidth]{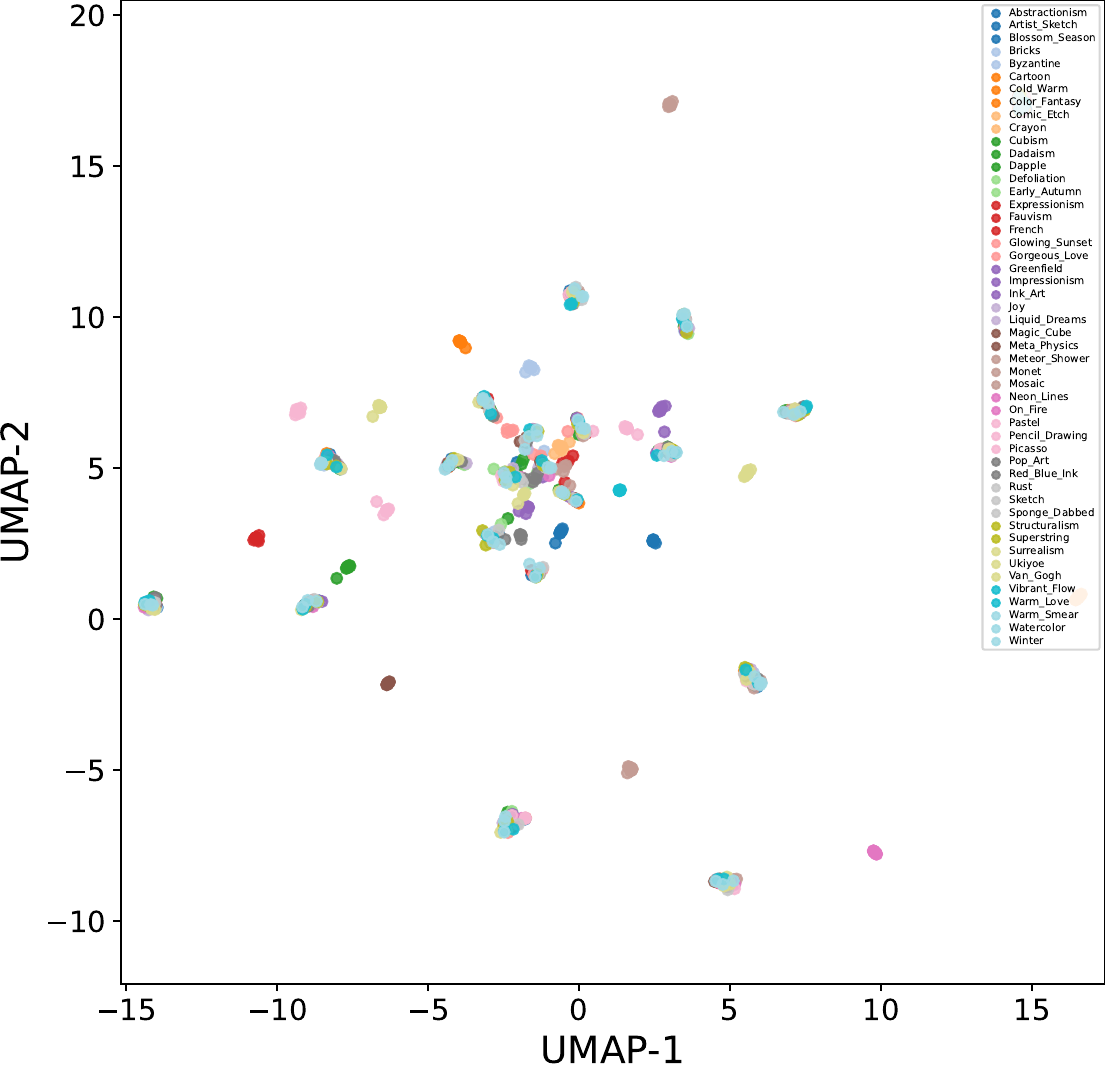}
        \vspace{1pt}
        {\small Vanilla SAE (SAeUron)} \\
        {\scriptsize Cluster Rate ($\uparrow$): 0.98, Centroid Margin ($\downarrow$): 0.76}
    \end{minipage}
    \hfill
    \begin{minipage}[c]{0.49\linewidth}
        \centering
        \includegraphics[width=\linewidth]{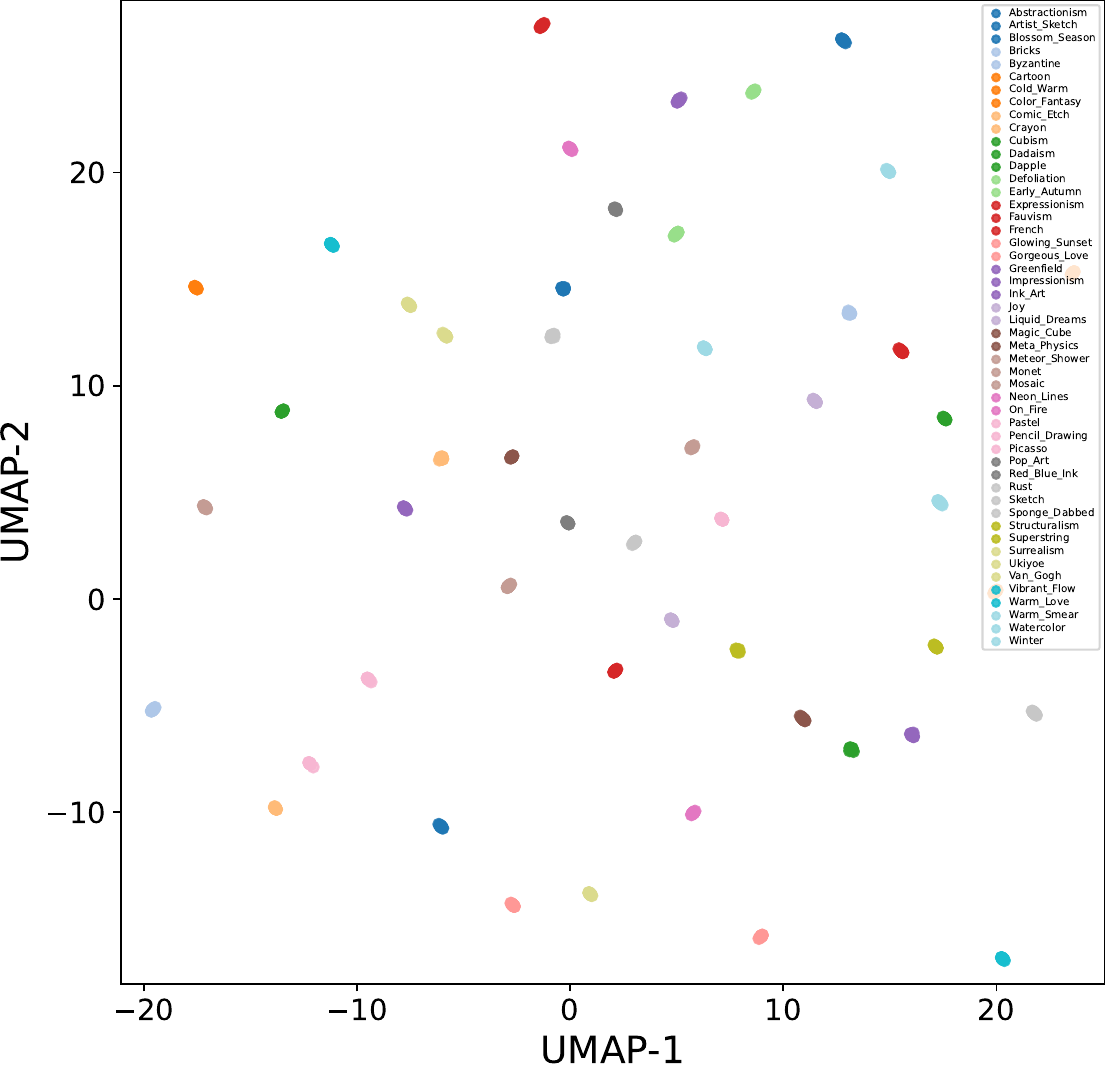}
        \vspace{1pt}
        {\small \textbf{SAEParate (Ours)}} \\
        {\scriptsize Cluster Rate ($\uparrow$): \textbf{1.00}, Centroid Margin ($\downarrow$): \textbf{0.08}}
    \end{minipage}
    \caption{UMAP visualization of sparse SAE latent representations for style unlearning: Vanilla SAE (SAeUron) (left) and SAEParate (right). Each color denotes a different style concept from UnlearnCanvas.}
    \label{fig:sae_latent_clusters_style}
\end{figure}

\begin{figure}[h]
    \centering

    \begin{minipage}[c]{0.49\linewidth}
        \centering
        \includegraphics[width=\linewidth]{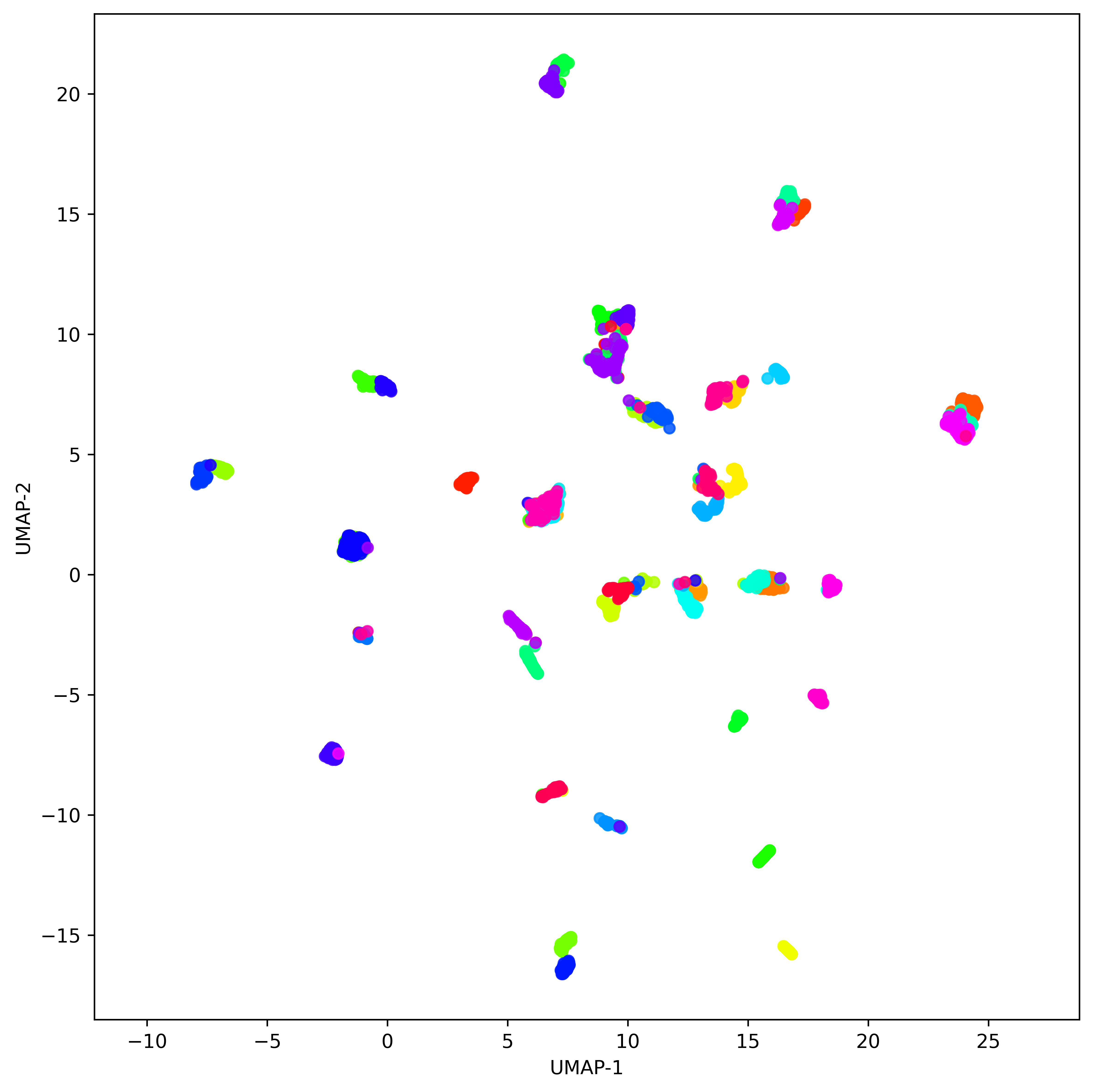}
        \vspace{1pt}
        {\small Vanilla SAE (SAeUron)} \\
        {\scriptsize Cluster Rate ($\uparrow$): 0.98, Centroid Margin ($\downarrow$): 0.93}
    \end{minipage}
    \hfill
    \begin{minipage}[c]{0.49\linewidth}
        \centering
        \includegraphics[width=\linewidth]{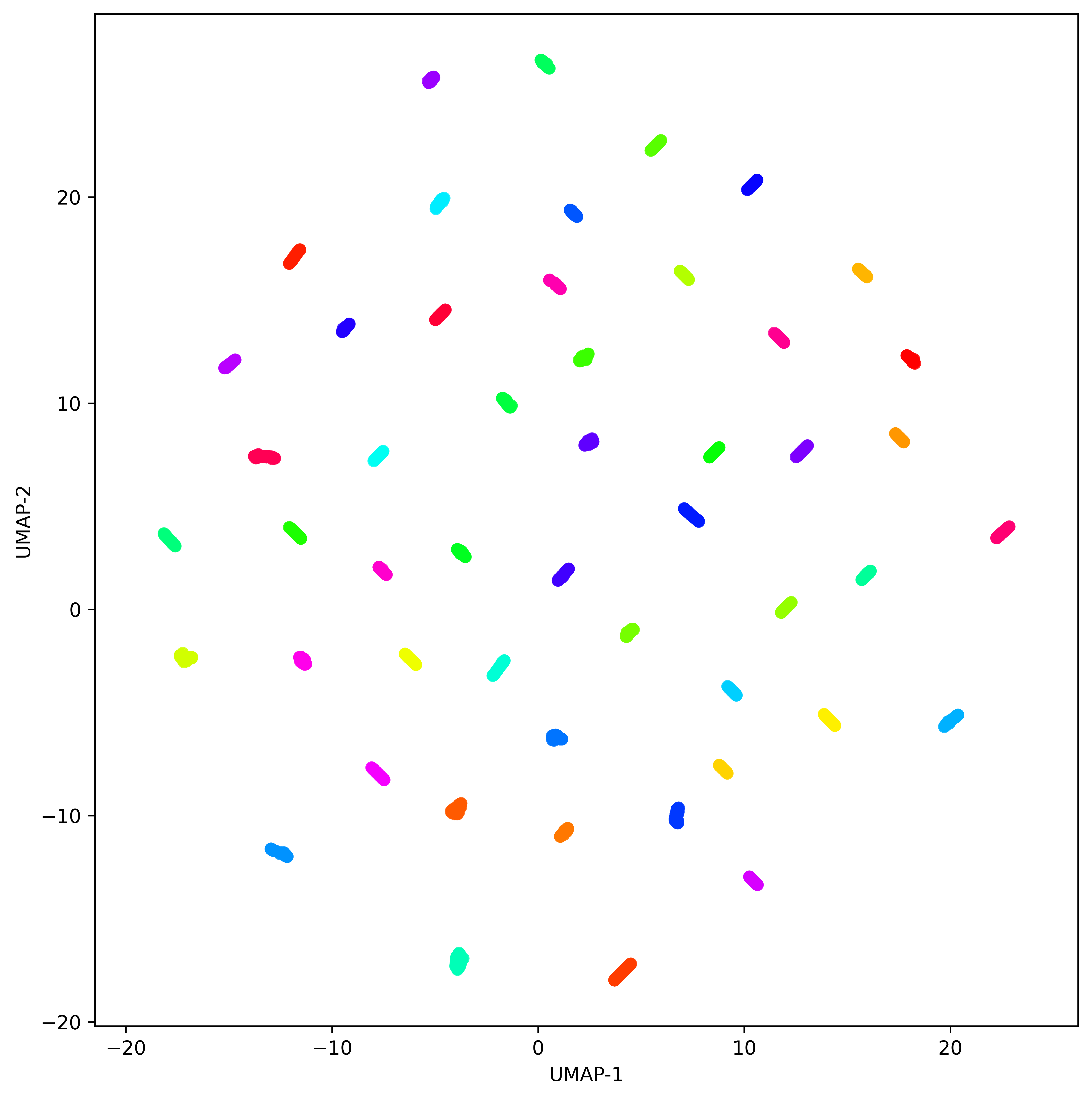}
        \vspace{1pt}
        {\small \textbf{SAEParate (Ours)}} \\
        {\scriptsize Cluster Rate ($\uparrow$): \textbf{1.00}, Centroid Margin ($\downarrow$): \textbf{0.04}}
    \end{minipage}
    \vspace{5mm}
    \includegraphics[width=0.75\linewidth]
    {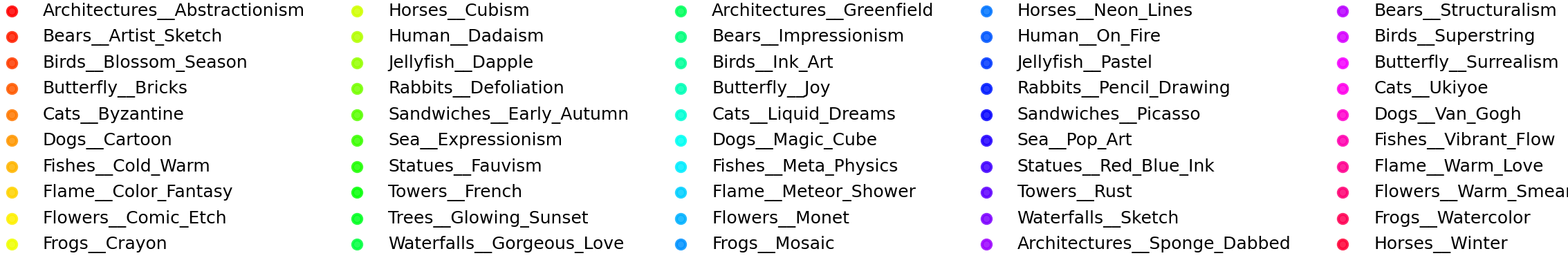}

    \caption{UMAP visualization of sparse SAE latent representations for style--object joint unlearning: Vanilla SAE (SAeUron) (left), SAEParate (right). Each color denotes a different style--object concept from UnlearnCanvas.}
    \label{fig:sae_latent_clusters_joint}
\end{figure}

\begin{figure}[h]
    \centering
    \begin{minipage}[c]{0.49\linewidth}
        \centering
        \includegraphics[width=\linewidth]{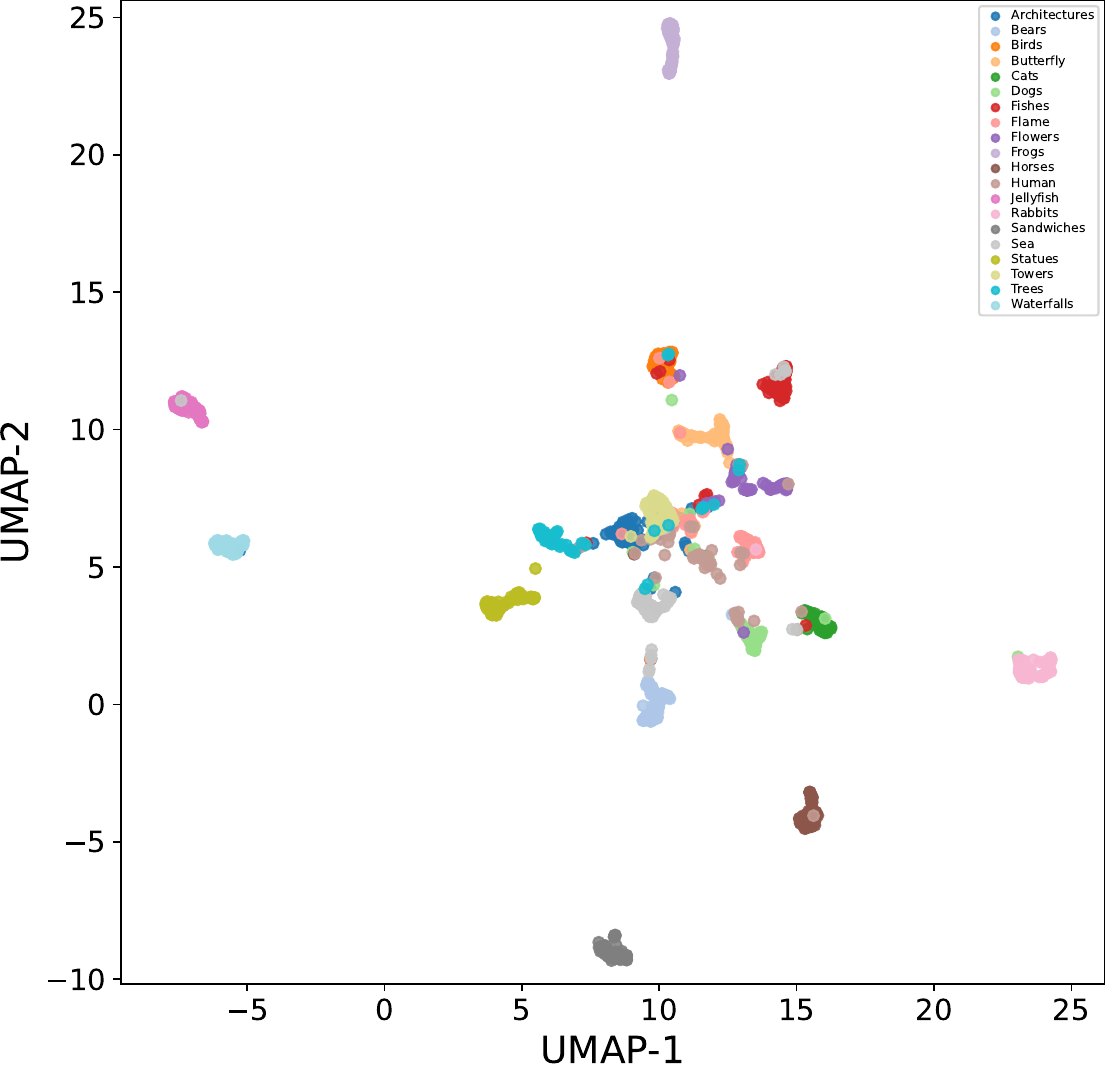}
        
        \vspace{1pt}
        {\small Vanilla SAE (SAeUron)} \\
        {\scriptsize Cluster Rate ($\uparrow$): 0.93, Centroid Margin ($\downarrow$): 0.73}
    \end{minipage}
    \hfill
    \begin{minipage}[c]{0.49\linewidth}
        \centering
        \includegraphics[width=\linewidth]{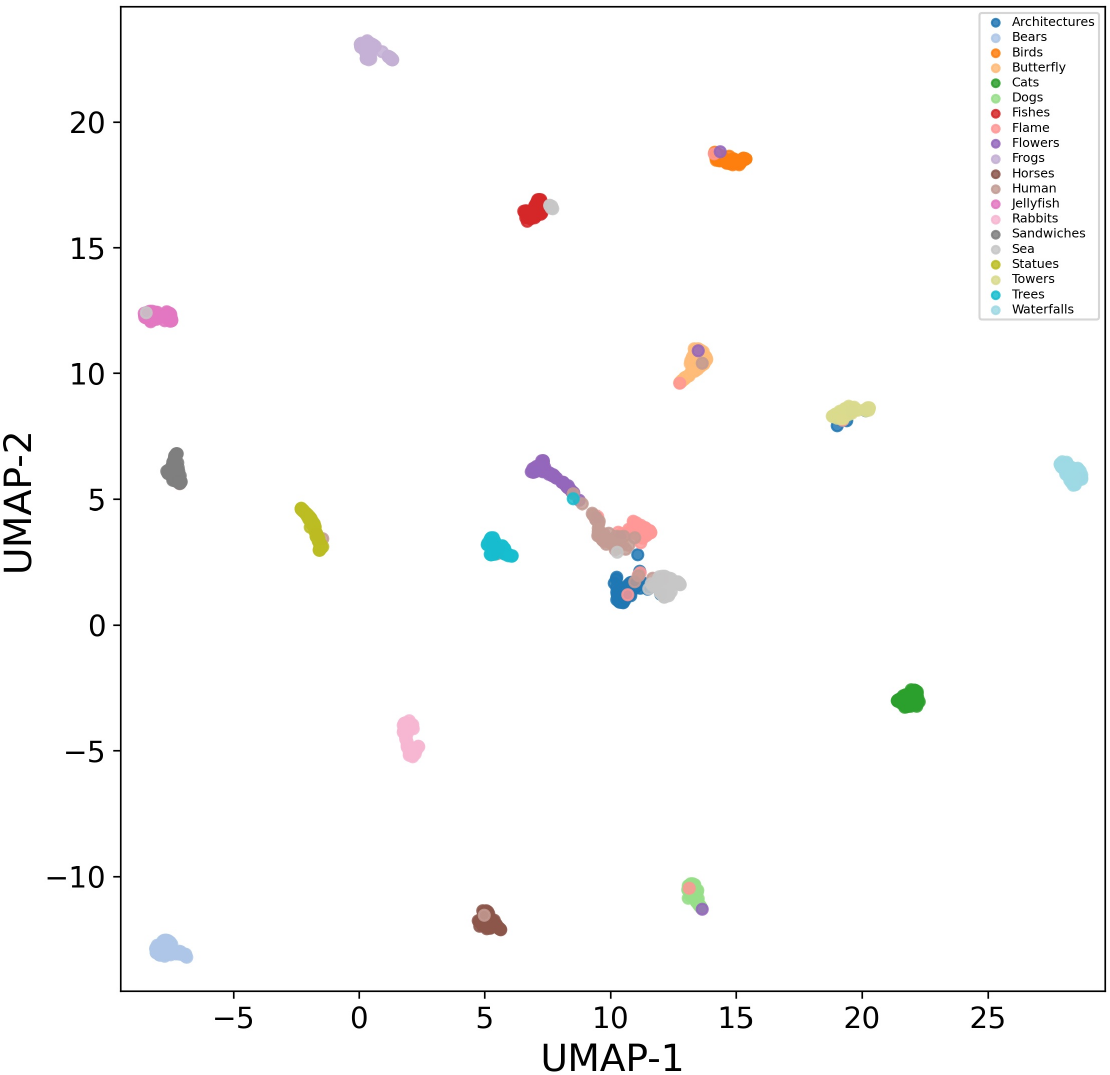}
        \vspace{1pt}
        {\small \textit{only}-$\mathrm{LSC}$} \\
        {\scriptsize Cluster Rate ($\uparrow$): 0.98, Centroid Margin ($\downarrow$): 0.71}
    \end{minipage}

    \vspace{0.5em}

    \begin{minipage}[c]{0.49\linewidth}
        \centering
        \includegraphics[width=\linewidth]{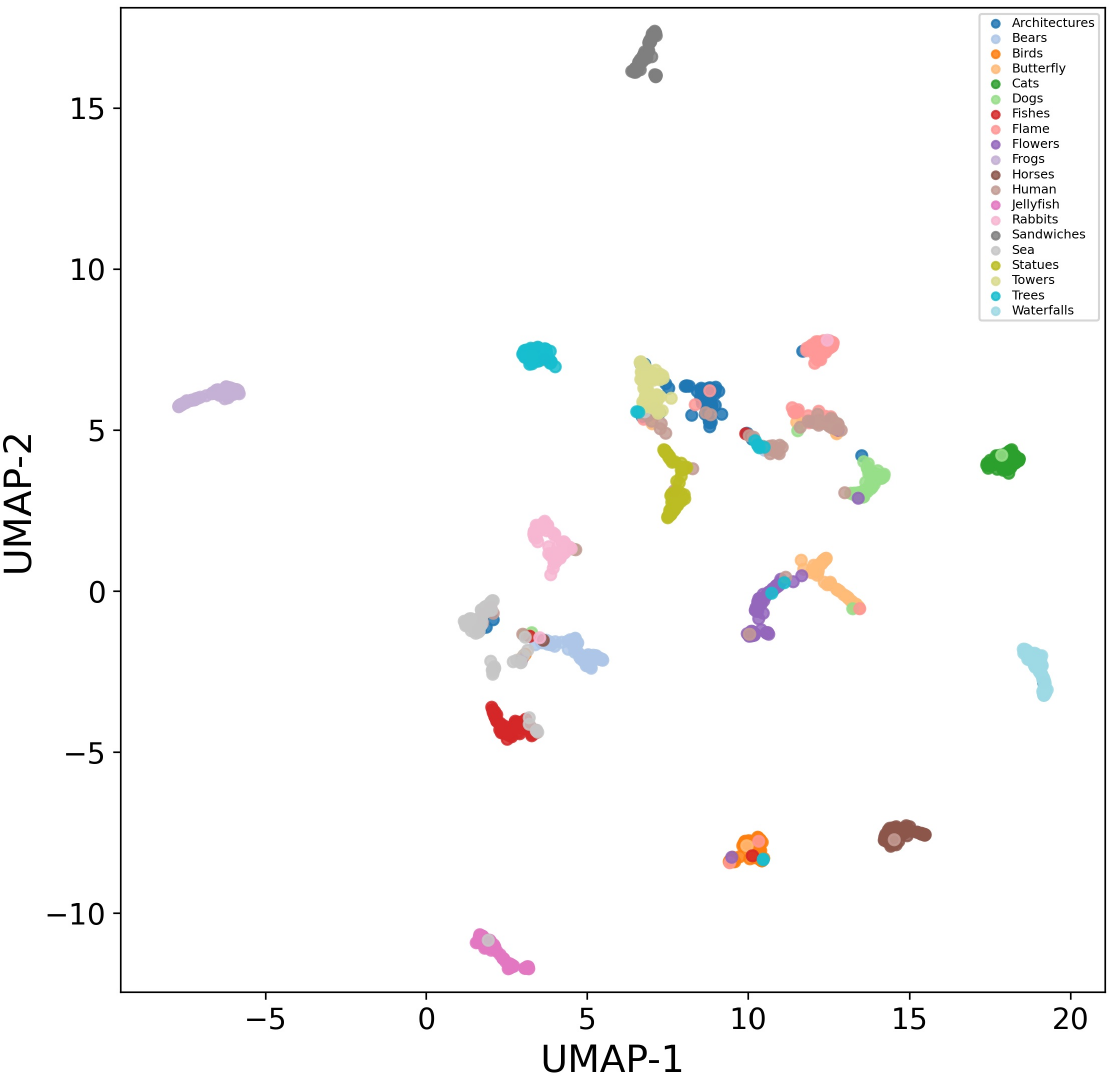}
        \vspace{1pt}
        {\small \textit{only}-GeLU} \\
        {\scriptsize Cluster Rate ($\uparrow$): 0.94, Centroid Margin ($\downarrow$): 0.64}
    \end{minipage}
    \hfill
    \begin{minipage}[c]{0.49\linewidth}
        \centering
        \includegraphics[width=\linewidth]{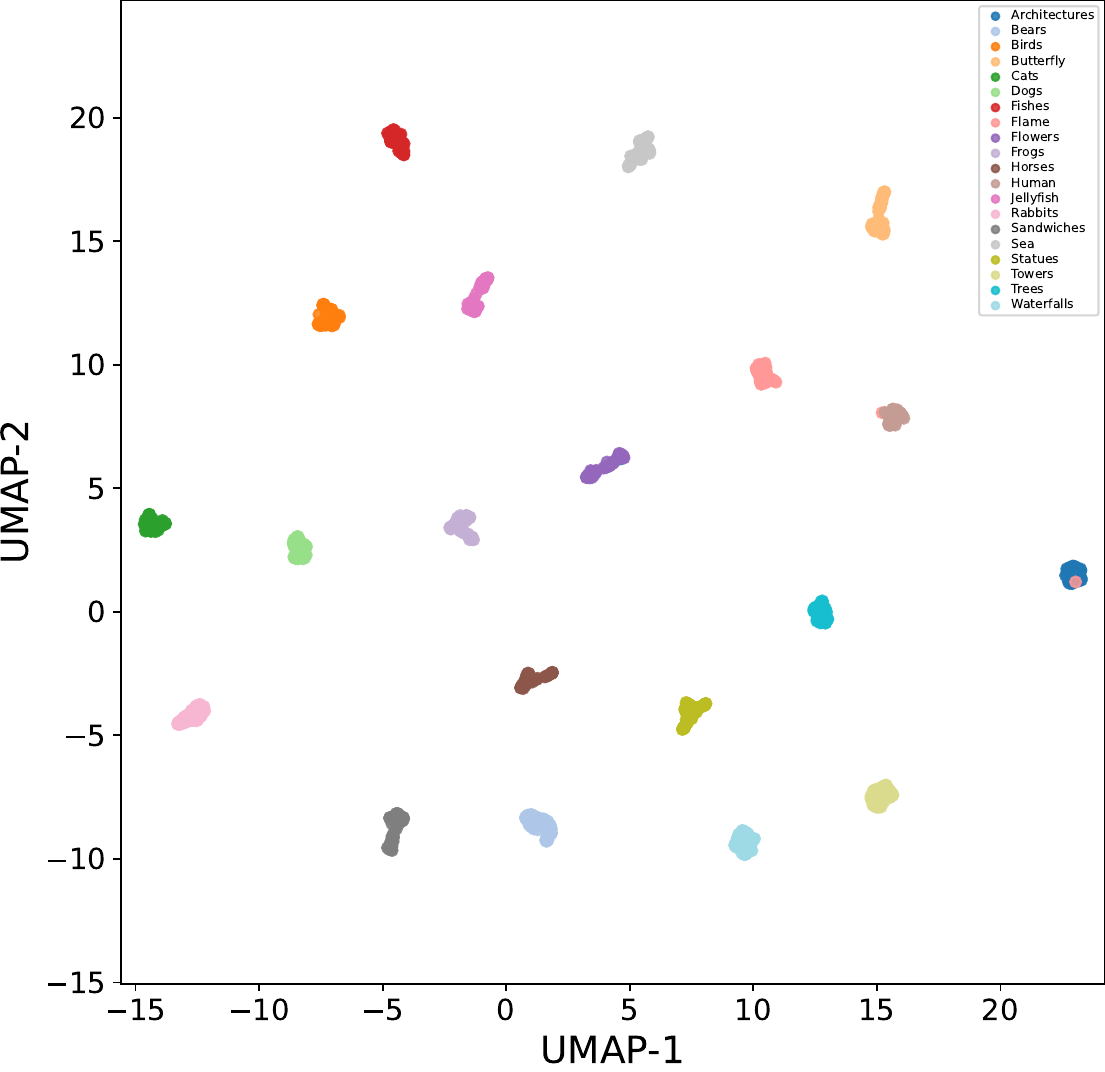}
        \vspace{1pt}
        {\small \textbf{SAEParate (Ours)}} \\
        {\scriptsize Cluster Rate ($\uparrow$): \textbf{1.00}, Centroid Margin ($\downarrow$): \textbf{0.06}}
    \end{minipage}

    \caption{
    UMAP visualization of sparse SAE latent representations for Vanilla SAE (SAeUron) (top-left), \textit{only}-$\mathrm{LSC}$ (top-right), \textit{only}-GeLU (bottom-left), and SAEParate (Ours) (bottom-right). Each color denotes a different concept class from UnlearnCanvas.
    }
    \label{fig:cluster_viz_ablation}
\end{figure}

Using the same visualization protocol as the object-level analysis in Figure~\ref{fig:sae_latent_clusters}, we first provide the style-level latent visualization in Figure~\ref{fig:sae_latent_clusters_style}. We then visualize the latent space of SAEs trained for joint style--object concepts in Figure~\ref{fig:sae_latent_clusters_joint}. Finally, we additionally report the latent space visualizations of the \textit{only}-$\mathrm{LSC}$ and \textit{only}-GeLU variants discussed in the ablation study in Section~\ref{sec:further_analyses} with Figure~\ref{fig:cluster_viz_ablation}.

Across these visualizations, Vanilla SAE exhibits noticeably entangled sparse representations, whereas SAEParate learns a more concept-discriminative latent space with clearer class-wise separation. This suggests that SAEParate reduces co-activation of shared latent features across concepts, thereby mitigating interference during the unlearning process. 

\clearpage

\subsection{Detailed Performance Heatmaps}

\begin{figure}[h]
    \centering
    \includegraphics[width=1.0\linewidth]{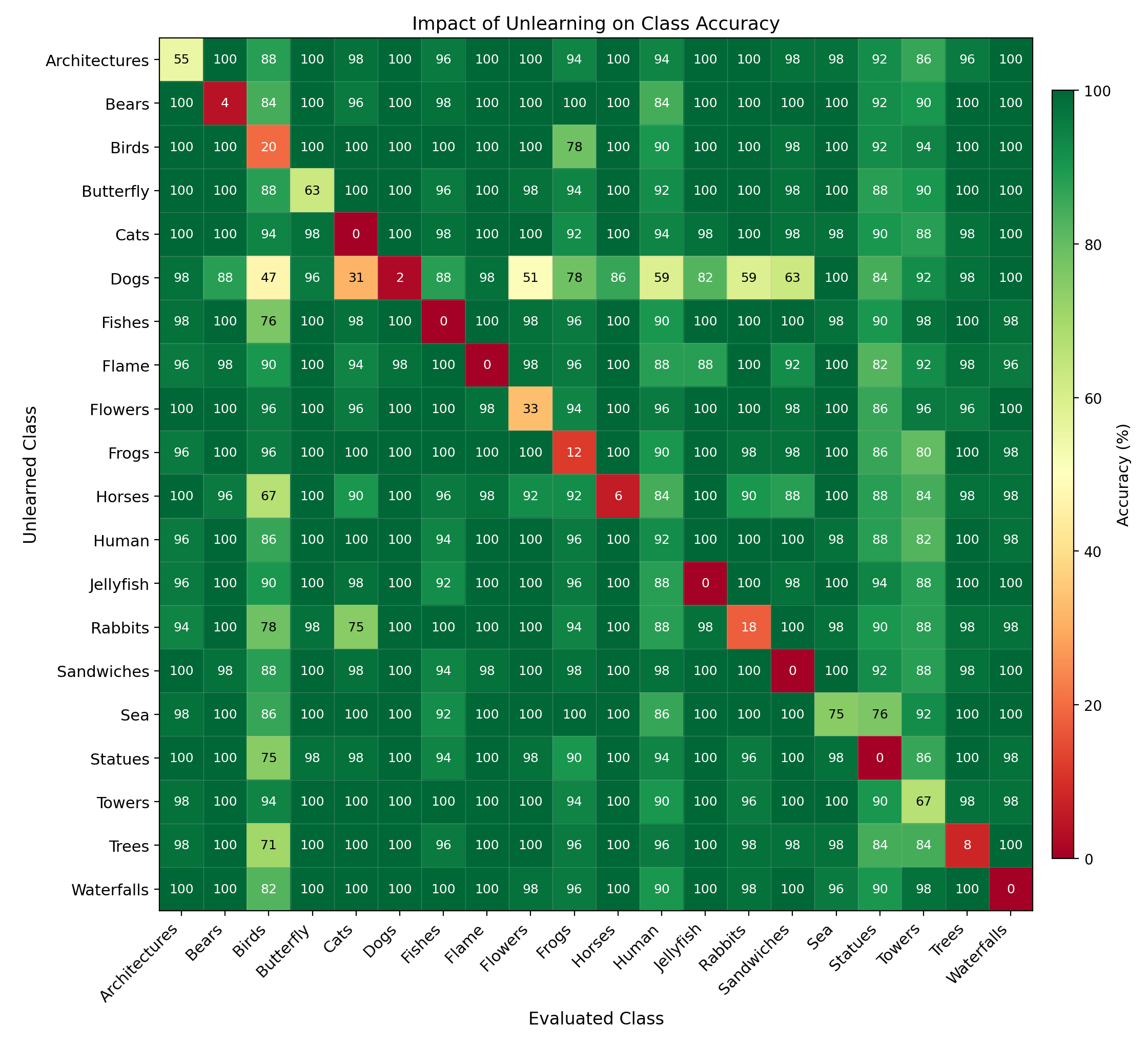}
    \caption{Detailed performance heatmap of SAeUron in object unlearning.}
    \label{fig:saeuron_object_performance_heatmap}
\end{figure}

\begin{figure}[h]
    \centering
    \includegraphics[width=1.0\linewidth]{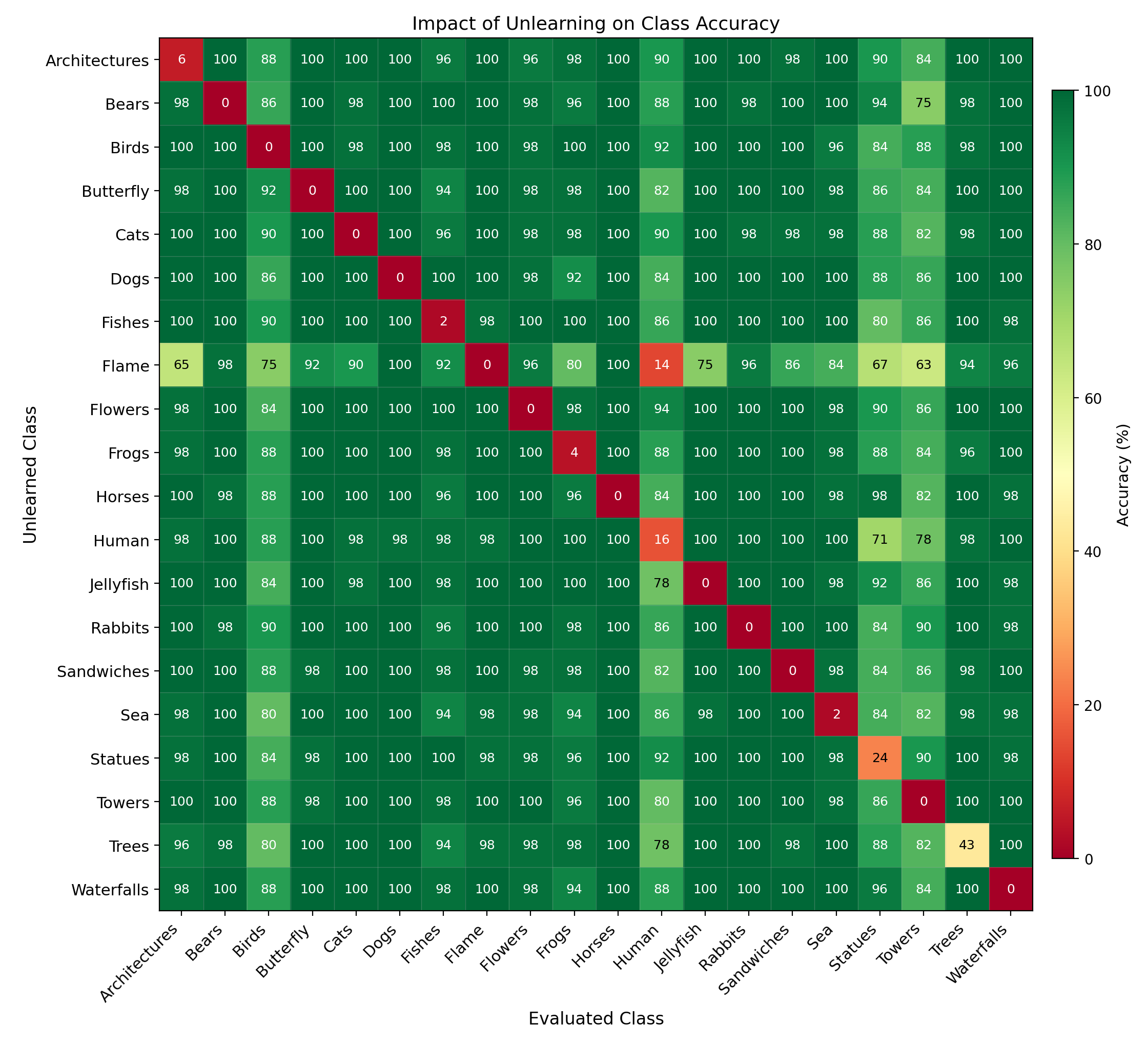}
    \caption{Detailed performance heatmap of SAEParate in object unlearning.}
    \label{fig:saeparate_object_performance_heatmap}
\end{figure}

\begin{figure}[h]
    \centering
    \includegraphics[width=1.0\linewidth]{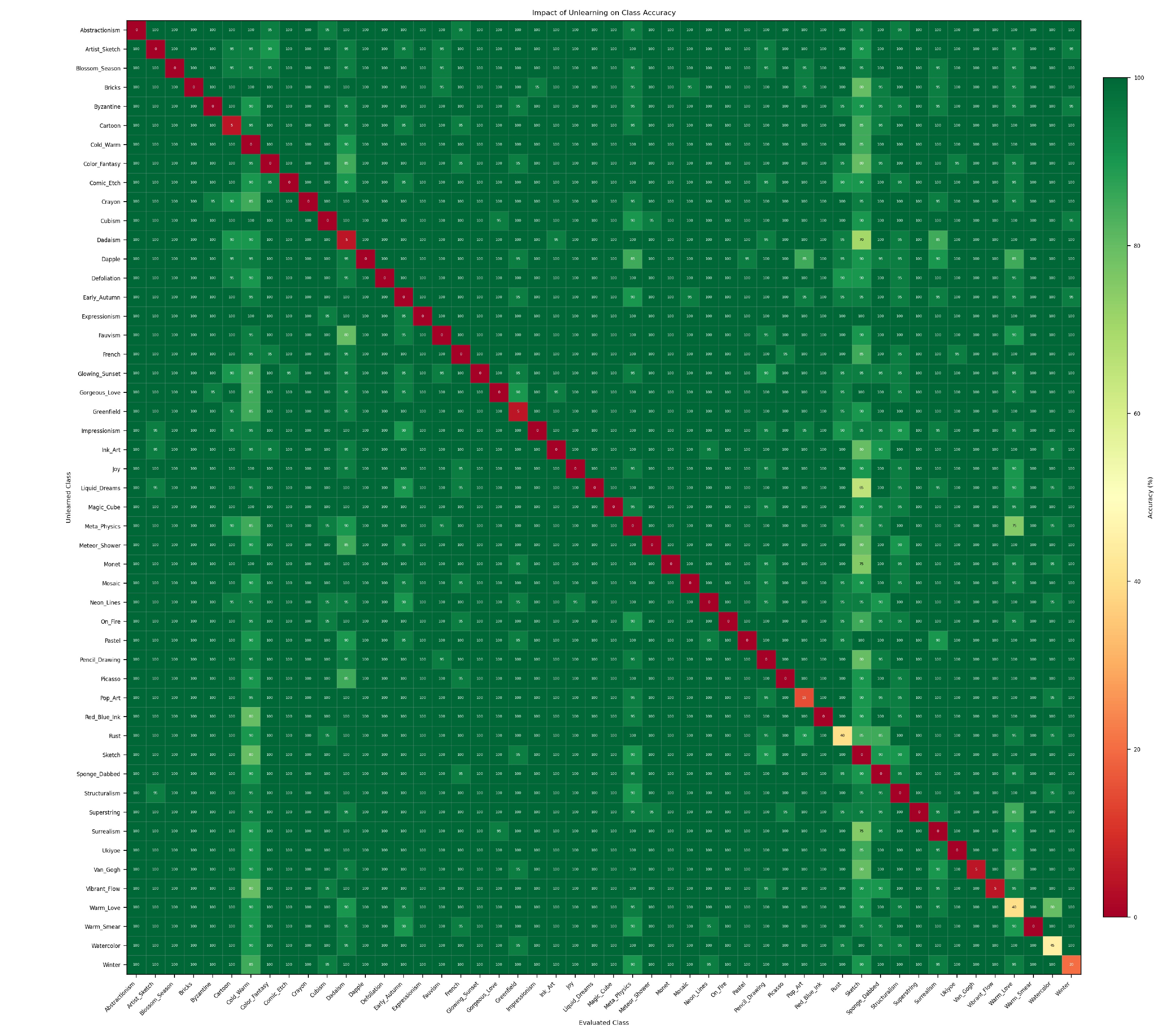}
    \caption{Detailed performance heatmap of SAeUron in style unlearning.}
    \label{fig:saeuron_style_performance_heatmap}
\end{figure}

\begin{figure}[h]
    \centering
    \includegraphics[width=1.0\linewidth]{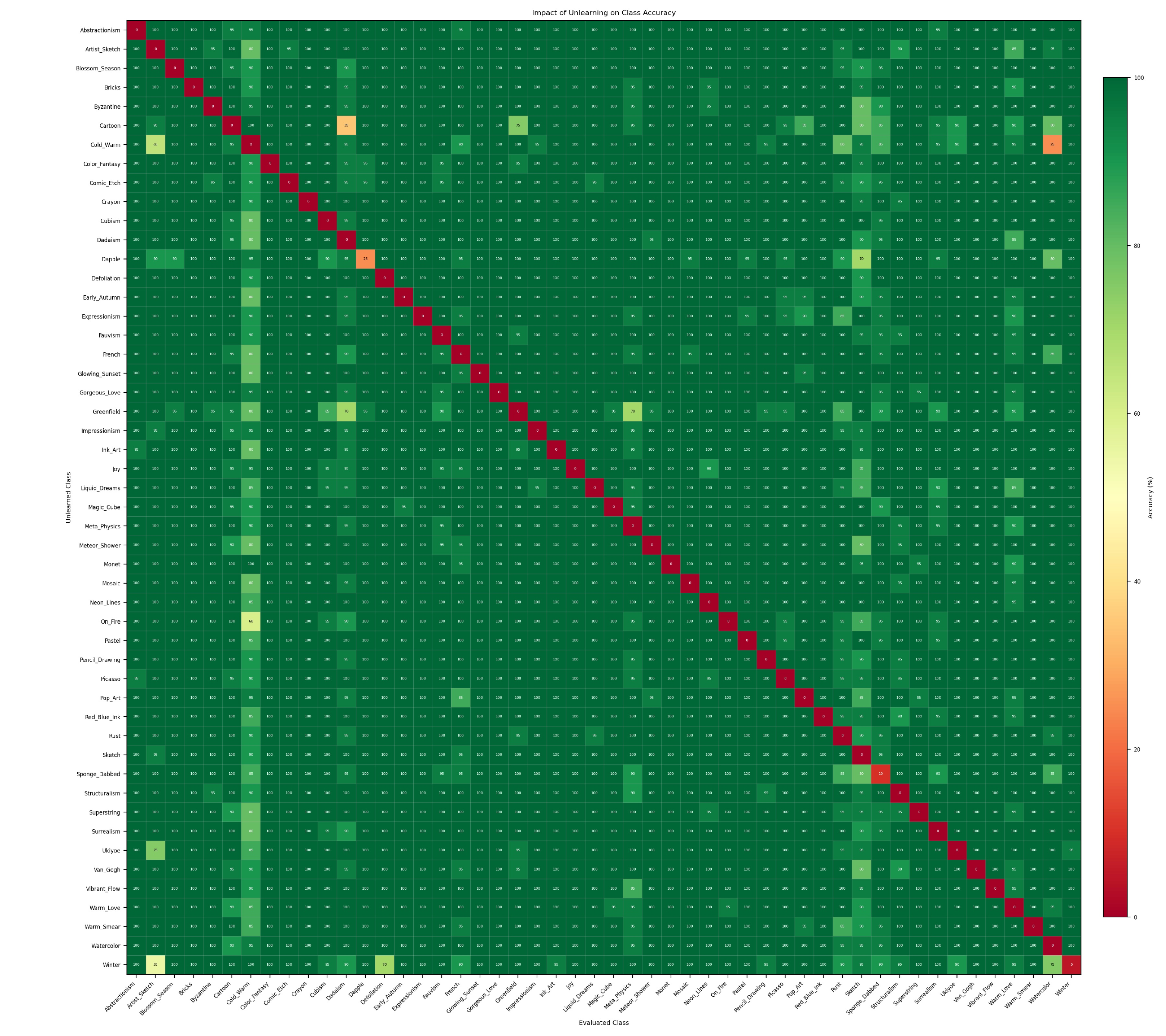}
    \caption{Detailed performance heatmap of SAEParate in style unlearning.}
    \label{fig:saeparate_style_performance_heatmap}
\end{figure}

For measuring the class-wise impact of unlearning, we report classification-accuracy heatmaps over all UnlearnCanvas concepts, including the target concept itself. We provide object-unlearning heatmaps for SAeUron and SAEParate in Figures~\ref{fig:saeuron_object_performance_heatmap} and~\ref{fig:saeparate_object_performance_heatmap}, respectively, and style-unlearning heatmaps in Figures~\ref{fig:saeuron_style_performance_heatmap} and~\ref{fig:saeparate_style_performance_heatmap}. As shown in these heatmaps, SAEParate achieves near-perfect unlearning accuracy along the target-concept diagonal while preserving high accuracy for non-target concepts.

\clearpage

\subsection{Qualitative Comparisons}

\label{app:quality_eval_results}

\begin{figure}[h]
    \centering
    \includegraphics[width=1.0\linewidth]{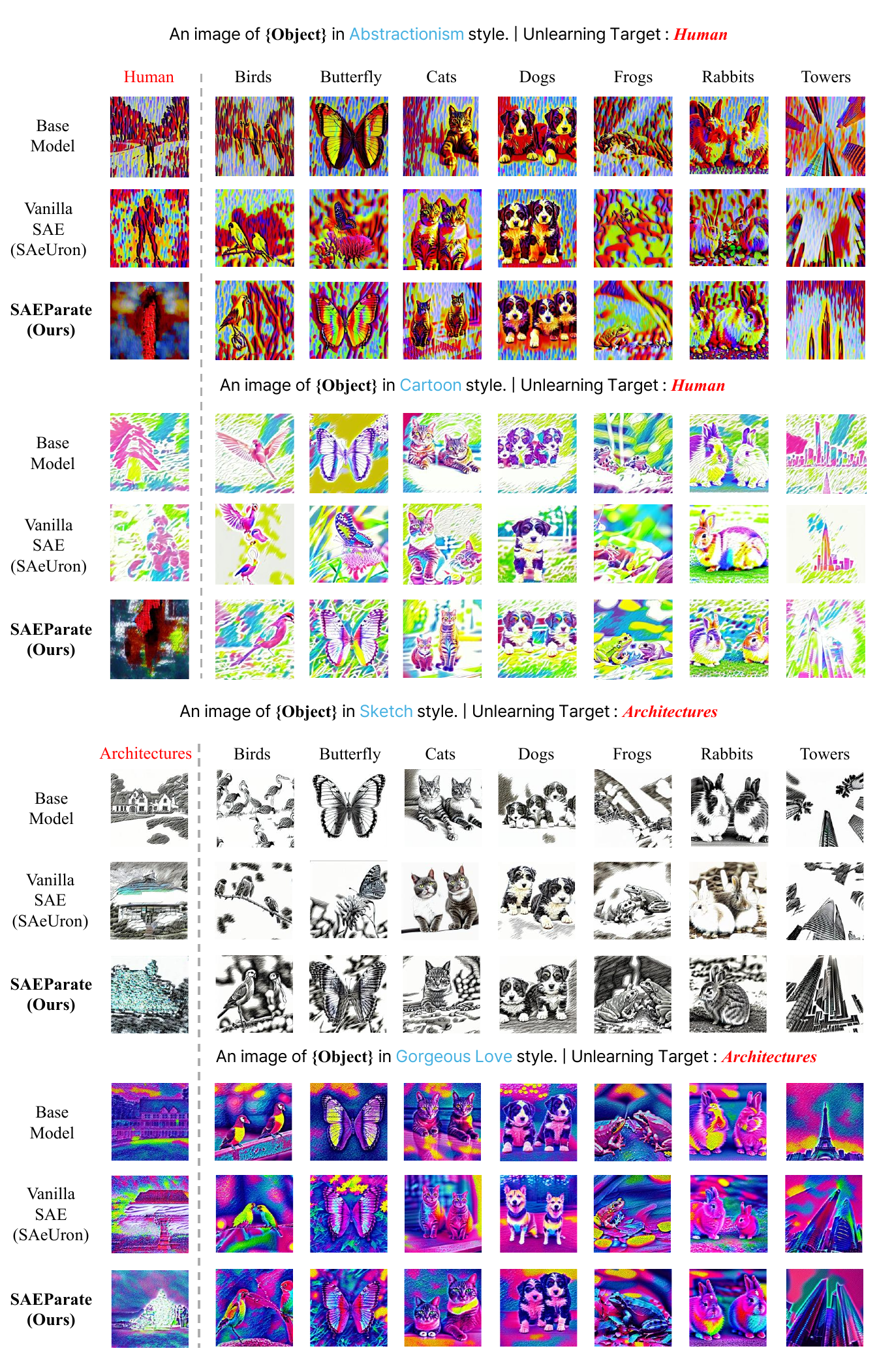}
    \caption{Qualitative comparison with Vanilla SAE (SAeUron)  on object unlearning.}
    \label{fig:image_quality_object}
\end{figure}

\begin{figure}[h]
    \centering
    \includegraphics[width=1.0\linewidth]{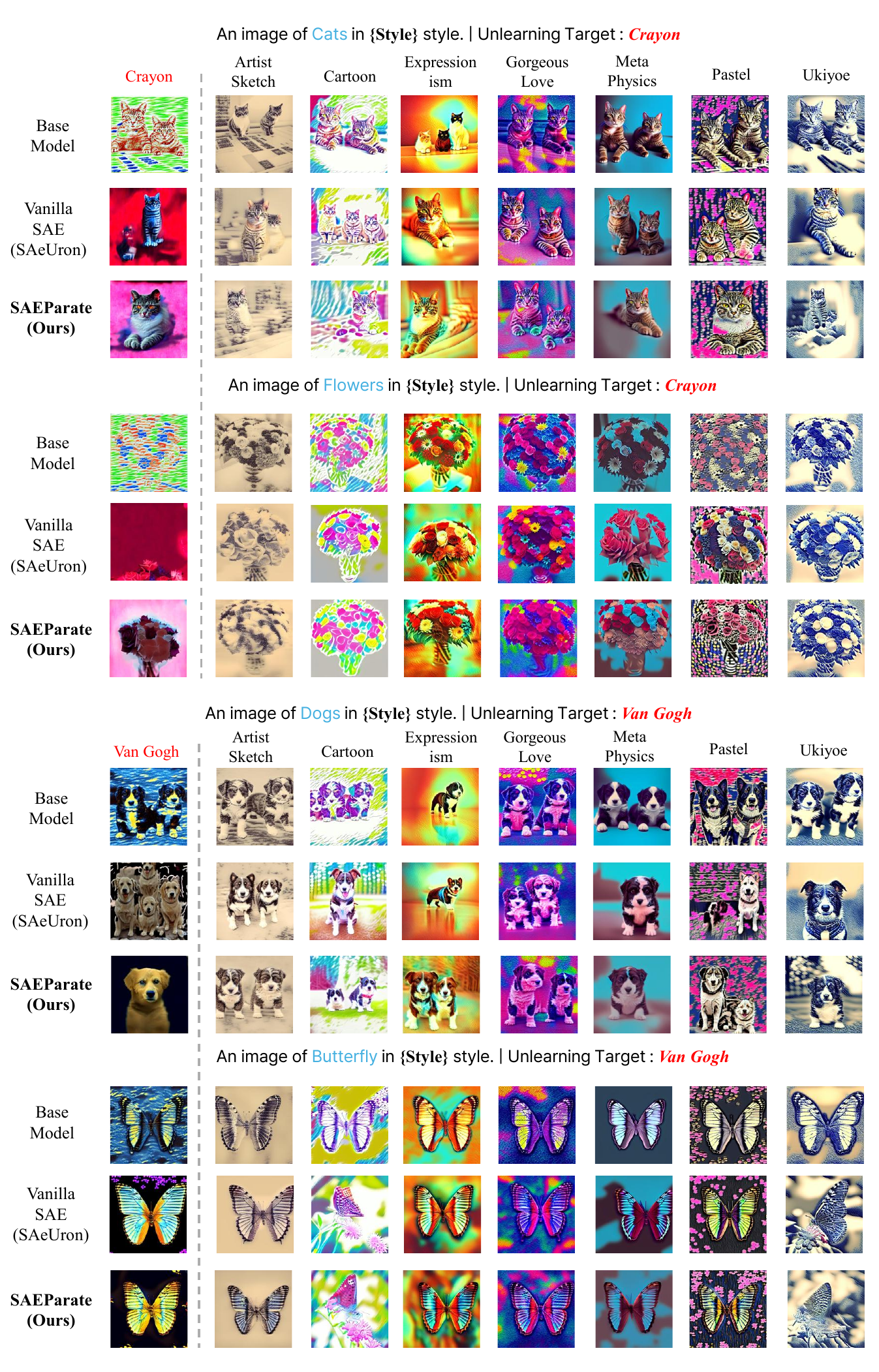}
    \caption{Qualitative comparison with Vanilla SAE (SAeUron) on style unlearning.}
    \label{fig:image_quality_style}
\end{figure}

\begin{figure}[h]
    \centering
    \includegraphics[
        width=\textwidth,
        height=1.0\textheight,
        keepaspectratio
        ]{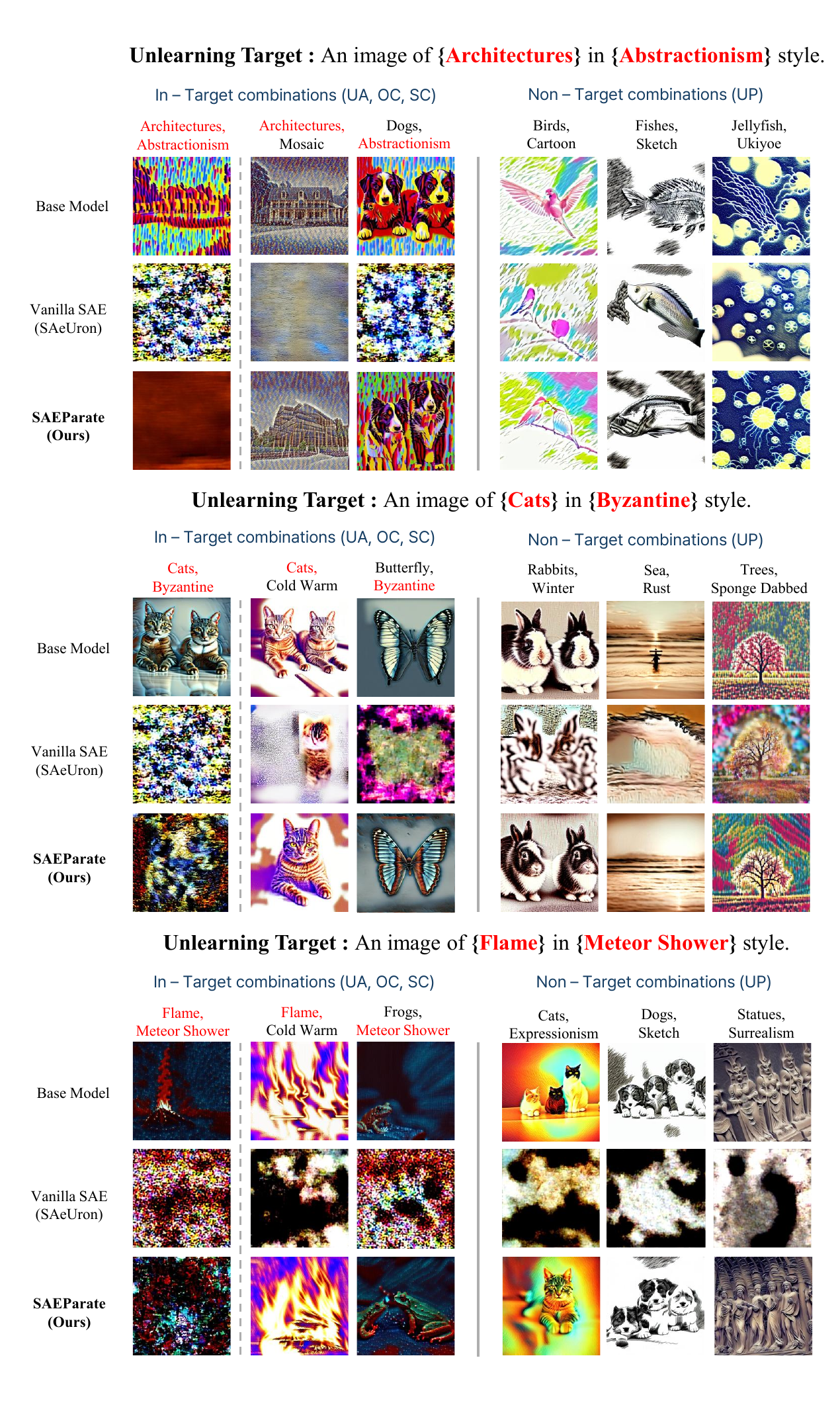}
    \caption{Qualitative comparison with Vanilla SAE (SAeUron) on style--object joint unlearning.}
    \label{fig:image_quality_joint}
\end{figure}

\begin{figure}[h]
    \centering
    \includegraphics[
        width=\textwidth,
        height=1.0\textheight,
        keepaspectratio
        ]{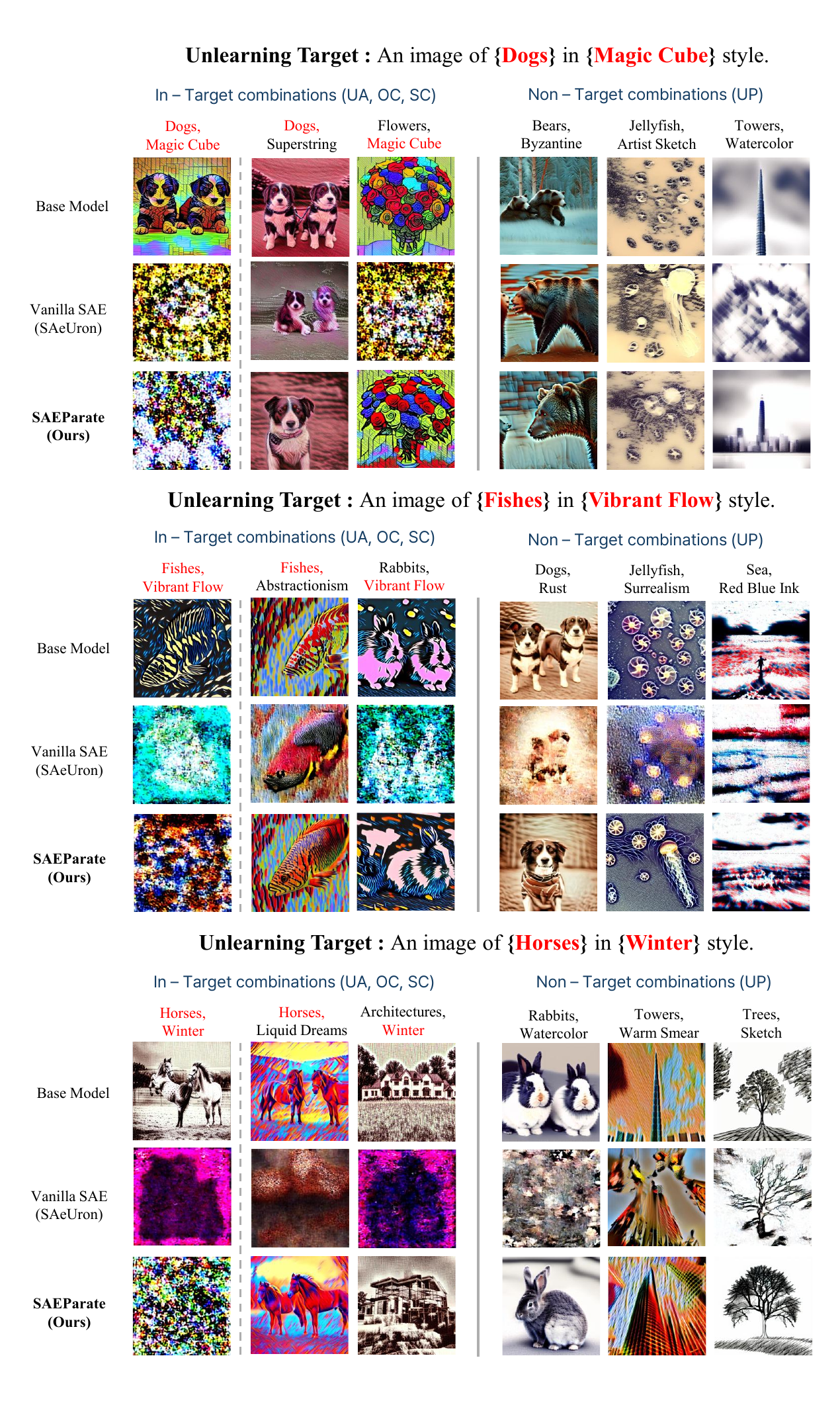}
    \caption{Qualitative comparison with Vanilla SAE (SAeUron) on style--object joint unlearning.}
    \label{fig:image_quality_joint_2}
\end{figure}

We provide qualitative comparisons for object, style, and joint unlearning scenarios, including images generated by the base model, SAeUron, and SAEParate. The results are shown in Figures~\ref{fig:image_quality_object}, \ref{fig:image_quality_style}, \ref{fig:image_quality_joint}, and \ref{fig:image_quality_joint_2}. Notably, SAEParate preserves non-target concepts more faithfully than SAeUron, with particularly strong retention in the joint unlearning examples shown in Figures~\ref{fig:image_quality_joint} and \ref{fig:image_quality_joint_2}.

\section{Broader Impacts}
\label{app:broader_impact}

This work aims to advance machine learning by improving selective concept unlearning in text-to-image diffusion models. While our method is designed to remove unwanted, biased, or harmful content while preserving non-target concepts, it is inherently dual-use and could be misused for censorship, unfair content suppression, or undesirable manipulation of model behavior. Therefore, real-world deployment should require transparent target definitions, careful evaluation, and appropriate human oversight.

\clearpage

\newpage

\end{document}